\documentclass[11pt]{article}
\usepackage[utf8]{inputenc}
\usepackage{fullpage}
\usepackage{xcolor}
\usepackage{amsmath, amsfonts}
\usepackage{mathrsfs}
\usepackage{graphicx,psfrag,epsf}
\usepackage{enumerate}
\usepackage[
backend=biber,
giveninits=true,
uniquename=false,
date=year,
style=authoryear-comp,
uniquelist=false,
sorting=nty,
url=false,
doi=false,
eprint=false,
maxbibnames=10,
minbibnames=3,
maxcitenames=2,
mincitenames=1
]{biblatex}
\DeclareNameAlias{sortname}{family-given}

\AtEveryBibitem{%
  \clearfield{doi}%
  \clearfield{isbn}%
  \clearfield{issn}%
  \clearfield{eprint}%
}

\usepackage{authblk} 

\usepackage{amsthm}
\usepackage{bbm}
\usepackage{stackengine}
\usepackage{hyperref}
\usepackage{float}

\AtEveryBibitem{\clearfield{month}}  % Hide month
\AtEveryBibitem{\clearlist{language}}

\newtheorem{proposition}{Proposition}[subsection] 

\newtheorem{theorem}{Theorem}
\newtheorem{corollary}{Corollary}
\newtheorem{lemma}{Lemma}
\newtheorem{remark}{Remark}
\newtheorem{assumption}{Assumption}

\newtheorem{innercustomgeneric}{\customgenericname}
\providecommand{\customgenericname}{}
\newcommand{\newcustomtheorem}[2]{%
  \newenvironment{#1}[1]
  {%
   \renewcommand\customgenericname{#2}%
   \renewcommand\theinnercustomgeneric{##1}%
   \innercustomgeneric
  }
  {\endinnercustomgeneric}
}
\newcustomtheorem{customlemma}{Lemma}
\newcustomtheorem{customcor}{Corollary}

\def\spacingset#1{\renewcommand{\baselinestretch}%
{#1}\small\normalsize} \spacingset{1}

\providecommand{\keywords}[1]
{
  \small	
  \textbf{\textit{Keywords---}} #1
}

\title{Dimension-reduced outcome-weighted learning for estimating individualized treatment regimes in observational studies}
\author[]{Sungtaek Son}
\author[]{Eardi Lila}
\author[]{Kwun Chuen Gary Chan}
\affil[]{Department of Biostatistics, University of Washington, Seattle}
\date{\vspace{-5ex}}
\spacingset{1.45} % DON'T change the spacing!

\begin{document}

\maketitle

\begin{abstract}
\noindent Individualized treatment regimes (ITRs) aim to improve clinical outcomes by assigning treatment based on patient-specific characteristics. However, existing methods often struggle with high-dimensional covariates, limiting accuracy, interpretability, and real-world applicability. We propose a novel sufficient dimension reduction approach that directly targets the contrast between potential outcomes and identifies a low-dimensional subspace of the covariates capturing treatment effect heterogeneity. This reduced representation enables more accurate estimation of optimal ITRs through outcome-weighted learning. To accommodate observational data, our method incorporates kernel-based covariate balancing, allowing treatment assignment to depend on the full covariate set and avoiding the restrictive assumption that the subspace sufficient for modeling heterogeneous treatment effects is also sufficient for confounding adjustment.  We show that the proposed method achieves universal consistency, i.e., its risk converges to the Bayes risk, under mild regularity conditions. We demonstrate its finite sample performance through simulations and an analysis of intensive care unit sepsis patient data to determine who should receive transthoracic echocardiography.
\end{abstract}


\section{Introduction}
An individualized treatment regime (ITR) is a mapping from individual-level characteristics to a treatment rule, which aims to maximize the expected outcome and accommodate treatment effect heterogeneity across the population \parencite{murphy_marginal_2001, bickel_optimal_2004, zhao_efficient_2019}. Statistical methods for identifying optimal treatments have been applied across a range of clinical areas, including inflammatory bowel disease, cancer, depression, substance abuse, and beyond \parencite{zhao_efficient_2019, zhao_reinforcement_2009, xie_multiple_2022, rosthoj_estimation_2006}.

Broadly speaking, there are two main approaches to learning ITRs: \textit{indirect} and \textit{direct} methods. Indirect methods model the conditional mean of the potential outcomes under each treatment, $\{Y(+1),Y(-1)\}$, given covariates $X$, that is, \( \mathbb{E}[Y(+1) \mid X] \) and \( \mathbb{E}[Y(-1) \mid X] \) \parencite{murphy_marginal_2001, zhao_reinforcement_2009, watkins_q-learning_1992, schulte_q-_2014}. Alternatively, they may model the conditional average treatment effect, \( \mathbb{E}[Y(+1) - Y(-1) \mid X] \) \parencite{murphy_optimal_2003, bickel_optimal_2004, schulte_q-_2014}. In these approaches, the optimal treatment at a given covariate value $X = x$ is defined as the one yielding the larger expected outcome (under the convention that larger outcomes are preferable). Therefore, the problem effectively reduces to accurate estimation of treatment effect heterogeneity, and methods such as R-learning \parencite{nie_quasi-oracle_2021} can then be used to derive optimal treatment regimes. Direct methods, on the other hand, aim to learn the treatment rule that maximizes the expected potential outcome by formulating an optimization problem over a prespecified class of decision functions. For example, outcome-weighted learning (OWL) and its extensions formulate this problem as a weighted support vector machine (SVM) \parencite{zhao_estimating_2012, zhou_residual_2017, liu_augmented_2018, zhou_augmented_2017}. While these methods can outperform indirect approaches, they remain sensitive to high-dimensional covariates \parencite{dasgupta_feature_2019}. 

Sufficient dimension reduction (SDR) techniques provide a powerful tool to mitigate the curse of dimensionality and improve performance across a variety of statistical modeling tasks \parencite{cook_dimension_2002, cook_principal_2018}. %\parencite{cook_dimension_2002, shin_probabilityenhanced_2014, fukumizu_gradient-based_2014, shin_principal_2017, cook_principal_2018, kang_forward_2022}. 
In this work, we propose an SDR framework for estimating individualized treatment regimes. The proposed SDR aims to project high-dimensional covariates onto a lower-dimensional subspace that preserves the conditional average treatment effect, while allowing treatment assignment $A$ to depend on the full covariate set $X$. We show that by targeting the subspace that retains treatment effect heterogeneity, our approach enables a more accurate estimation of optimal treatment regimes via direct methods.

The idea of reducing the dimensionality of the covariate space to facilitate learning ITRs has been explored in prior work. For example, \textcite{park_constrained_2021} proposes a single-index model that linearly transforms the covariates into a one-dimensional variable. This reduced representation is then used to estimate the treatment–covariate interaction effect, from which an optimal treatment rule can be derived. However, this approach relies on the assumption that a single linear combination of covariates adequately captures treatment effect heterogeneity, which may be overly restrictive.  SDR approaches for learning ITRs that relax this assumption have been proposed by \textcite{park_sufficient_2020} and \textcite{zhou_parsimonious_2021}. Specifically, \textcite{park_sufficient_2020} formulate the problem of estimating a reduced subspace of the covariates as a constrained least squares problem, modeling the interaction between treatment and the reduced covariate representation within a Reproducing Kernel Hilbert Space (RKHS). However, this approach is limited to randomized clinical trial settings, where treatment assignment is independent of the covariates. \textcite{zhou_parsimonious_2021} focus on the setting of continuous treatments, proposing a direct and pseudo-direct learning approach. Both approaches use the Nadaraya–Watson estimator of the conditional expectation function. However, the direct approach involves a computationally intensive alternating algorithm that iteratively updates the basis for the reduced subspace and the decision rule. The pseudo-direct method, on the other hand, relies on the additional assumption that the outcome depends only on the dimension-reduced covariates. Both approaches require solving non-convex objective functions and depend on aggressive dimension reduction to enable the use of the Nadaraya–Watson estimator.

Some SDR approaches designed for average treatment effect (ATE) estimation can also be used to learn ITRs. For example, joint SDR \parencite{huang_joint_2017, cheng_sufficient_2022} identifies a transformation \( B^\top X \) such that \( \{Y(+1), Y(-1)\} \perp\!\!\!\perp X \mid B^\top X \) and \( A \perp\!\!\!\perp X \mid B^\top X \). Based on this reduced representation, an ITR can be learned by estimating the conditional ATE given \( B^\top X \). However, this approach relies on the strong assumption that there exists a subspace \( \mathrm{span}(B) \) such that the conditional independence holds for the potential outcomes under both regimes, i.e., \( Y(+1) \) and \( Y(-1) \). This assumption is relaxed in \textcite{huang_robust_2023}; however, it is still assumed that $A \perp\!\!\!\perp X \mid B^\top X$, which can also be restrictive as it assumes that the subspace sufficient for modeling treatment effect heterogeneity is also sufficient for adjusting for confounding.

Our proposed SDR approach estimates a reduced subspace for learning ITRs. Specifically, we build upon gradient kernel dimension reduction (gKDR) \parencite{fukumizu_gradient-based_2014} to identify a subspace that captures the relationship between the covariates and the contrast between the two potential outcomes. This nonparametric method avoids the elliptical distribution assumption required by classical techniques such as sliced inverse regression \parencite{li_sliced_1991}. In addition, gKDR offers a computationally efficient procedure that scales well to large datasets and high-dimensional covariates, which are often challenging to deal with using other methods that also circumvent the ellipticity assumption: e.g., MAVE \parencite{xia_adaptive_2002} and KDR \parencite{fukumizu_kernel_2009}. 

Our approach accommodates observational settings by allowing treatment assignment to depend on the full set of covariates, rather than restricting it to the same reduced subspace that captures treatment effect heterogeneity. This is achieved by incorporating covariate balancing weights into the proposed SDR framework to account for differences in covariate distributions across treatment groups. Several strategies exist for estimating such weights. The most common is inverse propensity weighting (IPW), although this can be sensitive to model misspecification. As an alternative, a growing literature proposes methods that directly estimate balancing weights in the context of ATE estimation \parencite{imai_covariate_2014, zubizarreta_stable_2015, chan_globally_2016, wong_kernel-based_2018, hirshberg_minimax_2019, kallus_optimal_2022}. \textcite{chen_robust_2024} introduce a covariate balancing technique for learning ITRs with OWL methods. Here, we integrate the kernel-based covariate functional balancing (KCB) approach proposed by \textcite{wong_kernel-based_2018} in our SDR framework and develop asymptotic theory demonstrating its suitability for learning ITRs. Finally, the estimated subspace and the balancing weights are used to learn optimal treatment regimes via augmented outcome-weighted learning (AOL) \parencite{zhou_augmented_2017, liu_augmented_2018}. We establish theoretical guarantees for the overall procedure under mild regularity conditions.
%hainmueller_entropy_2012,

\begin{figure}
\centering
\includegraphics[width = 0.8\linewidth]{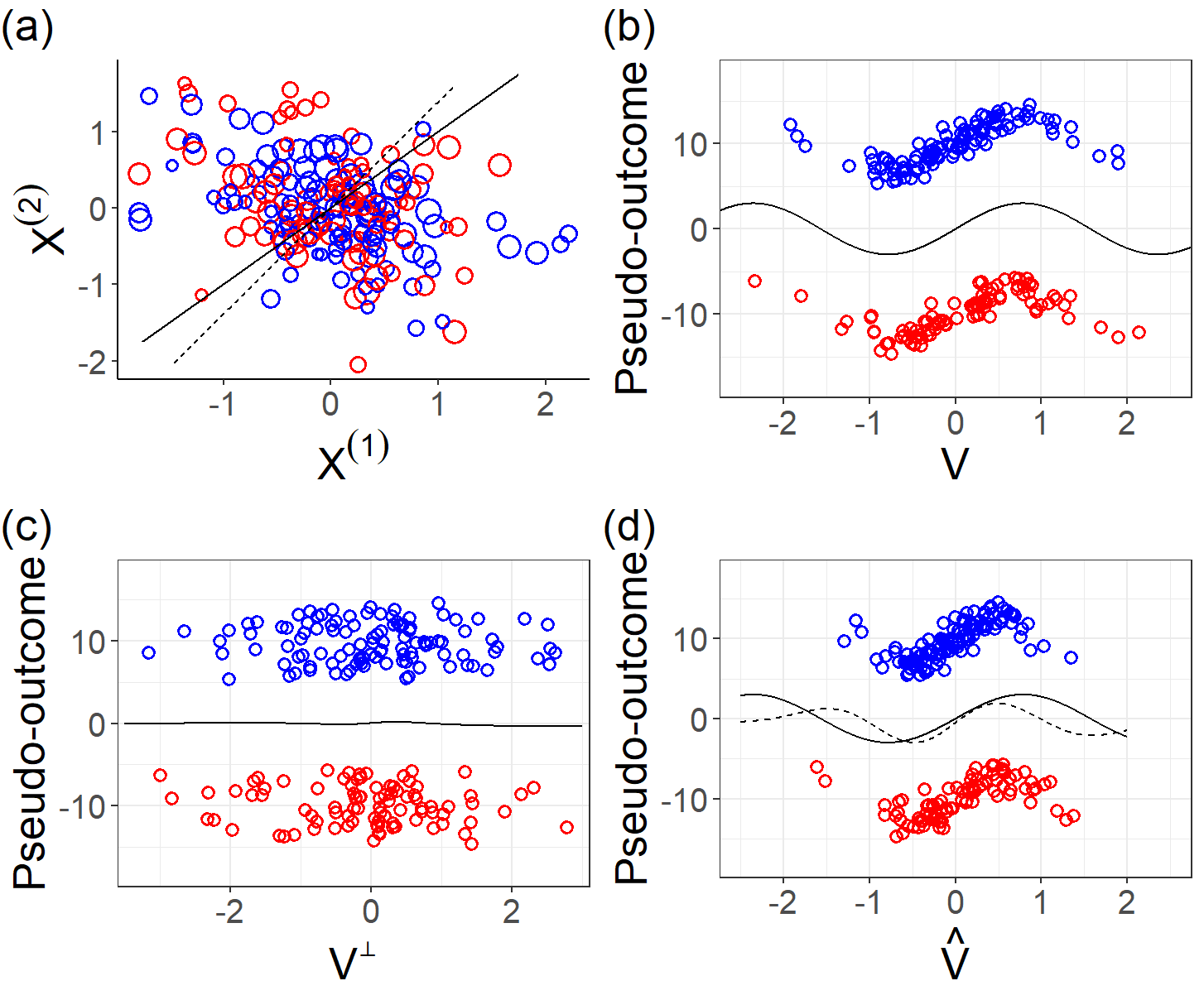}
\caption{Blue points correspond to the treatment group ``+1'' and red points correspond to the treatment group ``-1''. The sample sizes for the +1 and -1 treatment groups are 110 and 90, respectively. (a) shows the distribution of the data with respect to the covariates $\left(X^{(1)}, X^{(2)}\right)$. The magnitudes of the outcomes are represented by varying point sizes. The solid line represents the true projection line that best captures treatment effect heterogeneity, while the dashed line denotes the estimated projection line obtained from our proposed SDR approach. (b) displays the true pseudo-outcomes ($Y/\mathrm{Pr}(+1|X)$ and $-Y/\mathrm{Pr}(-1|X)$ for treatment groups $A=+1$ and $A=-1$) in the reduced subspace, where $\left(X^{(1)}, X^{(2)}\right)$ is projected onto $V = B_0^\top X$. The curve represents the heterogeneous treatment effect as a function of $V$. (c) shows the pseudo-outcomes projected onto $V^{\perp}$, the space perpendicular to that spanned by $B_0$. The solid line shows the treatment effect as a function of $V^\perp$. (d) shows the pseudo-outcomes as a function of $\hat{V} = \hat{B}^\top X$, where $\hat{B}$ is an estimate of $B_0$ obtained using the proposed SDR framework. The solid line shows the true heterogeneous treatment effect along the different levels of the dimension-reduced covariate. The dashed line shows the estimated decision rule along the different levels of the dimension-reduced covariate. The signs of these functions, which determine the true and estimated treatment assignments, are equal in the range $\hat{V} \in (-1, 1)$, where approximately 95\% of the data points lie.
}
\label{fig:illust}
\end{figure}

Figure~\ref{fig:illust} illustrates the proposed framework. In Figure~\ref{fig:illust}(a), we show the data, which consist of triplets of outcome $Y$, covariates $\left(X^{(1)}, X^{(2)}\right)$, and treatment group $A$. The solid line represents the linear transformation $B^\top X$ that best captures heterogeneity in treatment effect. Figure~\ref{fig:illust}(b) shows the pseudo-outcomes, defined as $Y/\mathrm{Pr}(+1|X)$ and $-Y/\mathrm{Pr}(-1|X)$ for treatment groups $A=+1$ and $A=-1$, respectively, as a function of $V = B^\top X$. Pseudo-outcomes allow us to account for differences in covariate distributions across treatment groups. Figure~\ref{fig:illust}(c) shows the pseudo-outcomes as a function of $V^\perp$, the subspace perpendicular to $\mathrm{span}(B)$. Note that there is no notable heterogeneous treatment effect when the covariates are projected onto this direction. Using the proposed SDR framework, we estimate the direction that best captures treatment effect, shown as the dashed line in Figure~\ref{fig:illust}(a). Figure~\ref{fig:illust}(d) depicts the pseudo-outcomes as a function of the estimated subspace and the optimal decision rule fitted using AOL. The sign of the true heterogeneous treatment effects (solid line) and the fitted decision function (dashed line) are equal in the range $\hat{V} \in (-1, 1)$, in which approximately 95\% of the data points lie. Moreover, using an unsupervised dimension reduction method such as principal components analysis would provide a poor reduction for the purpose of identifying optimal ITRs, since the first principal direction is more closely aligned with $V^\perp$.

The remainder of the paper is organized as follows. Section~\ref{sec:preliminaries} introduces the notation and outlines the assumptions underlying the proposed framework. In Section~\ref{sec:methods}, we present our methodology for SDR and describe the subsequent steps for optimal ITR estimation. Section~\ref{sec:theory} establishes the theoretical guarantees of the proposed approach. We present simulation results in Section~\ref{sec:simulations}. In Section~\ref{sec:application}, we demonstrate the utility of our method using intensive care unit sepsis patient data.

\section{Preliminaries}\label{sec:preliminaries}
We first introduce notation and the problem setup. Let $A \in \{-1, +1\}$ denote a binary treatment assignment, $X \in \mathcal{X} \subseteq \mathbb{R}^p$ the vector of covariates, and $Y \in \mathbb{R}$ a continuous outcome. We denote potential outcomes \parencite{rubin_estimating_1974} by $\{Y(A): A \in \{-1, +1\} \}$ and the observed outcome by $Y = \mathbbm{1}(A = +1) Y(+1) + \mathbbm{1}(A = -1) Y(-1)$, where $\mathbbm{1}(\cdot)$ is the indicator function. We use standard asymptotic notation throughout: if a sequence of random variables $X_n$ is such that $X_n = O_p(a_n)$, then $X_n/a_n$ is bounded in probability. If $X_n = o_p(a_n)$, then $X_n/a_n$ converges to zero in probability as $n \to \infty$. We use $X_n \to_p X$ when $X_n = X+ o_p(1)$.

In practice, the observed data consist of triplets $\{(A_i, Y_i, X_i) : i \in [n]\}$, where $n$ denotes the sample size and $[n]$ denotes the index set $\{1, 2, \dots, n\}$. We denote the $j$-th component of a random vector $X$ by $X^{(j)}$. The propensity score is defined as $\pi(A, X) = \Pr(A \mid X)$. An RKHS defined on a space $\mathcal{S}$ is denoted by $\mathcal{H}_{\mathcal{S}}$, and its associated norm and inner product are $\|\cdot\|_{\mathcal{H}_{\mathcal{S}}}$ and $\langle \cdot , \cdot \rangle_{\mathcal{H}_{\mathcal{S}}}$, respectively.

Our proposed methodology is based on the following assumptions:

\begin{assumption}[No unmeasured confounding]
    The potential outcomes are independent of treatment assignment given covariates: $\{Y(+1), Y(-1)\} \perp\!\!\!\perp A \mid X$.
\end{assumption}

\begin{assumption}[Positivity]
    The propensity score is strictly positive: $\pi(a, x) > 0$ for all $a \in \{-1, +1\}$ and all $x \in \mathcal{X}$.
\end{assumption}

\begin{assumption}[Central mean subspace]
\label{asm:SDR}
    There exists a central mean subspace, $\mathrm{span}(B_0)$, such that $\mathbb{E}[Y(+1) - Y(-1) \mid X] = \mathbb{E}[Y(+1) - Y(-1) \mid B_0^\top X],$ where $B_0 \in \mathbb{R}^{p \times u}$ with $u < p$.  Moreover, $\mathrm{span}(B_0)$ is the minimal subspace for which this equality holds \parencite{cook_dimension_2002}.
\end{assumption}

\textbf{Assumptions 1} and \textbf{2} are standard in the causal inference literature and are required to ensure identifiability. \textbf{Assumption 3} is central to the proposed methodology, which estimates a central mean subspace such that $B_0^\top X$ captures treatment effect heterogeneity. This enables a more efficient estimation of an ITR based on the reduced covariate representation. A key feature of our approach is that we do not assume $\{Y(+1), Y(-1)\} \perp\!\!\!\perp X \mid B_0^\top X$, thereby allowing for complex relationships between the potential outcomes and the covariates, as long as their contrast depends on a lower-dimensional subspace. Additionally, we do not impose the assumption $A \perp\!\!\!\perp X \mid B^\top X$, thus accommodating settings where the subspace sufficient for capturing treatment effect heterogeneity may not be sufficient for adjusting for confounding.

Let $d: \mathcal{X} \to \{-1, +1\}$ denote a decision rule that defines an ITR. This can be expressed as the sign of a decision function $f: \mathcal{X} \to \mathbb{R}$ as $d(\cdot) = \mathrm{sign}\, \circ f(\cdot)$. The value function, representing the expected reward under the decision rule $d$, is defined as the expected potential outcome under the treatment assigned by $d$, and is given by
\begin{equation}
\label{eq:value}
    \mathbb{E}\left[ Y\{ d(X) \} \right] = \mathbb{E}\left[ \frac{Y}{\pi(A, X)} \mathbbm{1}\{d(X) = A\}  \right] = \mathbb{E}\left[ \frac{Y}{\pi(A, X)} \mathbbm{1}\{\mathrm{sign} \circ f(X) = A\}  \right].
\end{equation}
Without loss of generality, we assume that larger values of the outcome are more desirable.

Given the central mean subspace $\mathrm{span}(B_0)$, we define the dimension-reduced covariates as $V_0 = B_0^\top X \in \mathcal{V}$, where $\mathcal{V} = \{v : v = B_0^\top x,\, x \in \mathcal{X}\} \subseteq \mathbb{R}^u$ denotes the corresponding reduced subspace. Given an estimate $\hat{B}$ of the subspace basis with rank $\hat{u}$ selected based on an appropriate tuning procedure, the estimated low-dimensional covariates are $\hat{V} = \hat{B}^\top X \in \hat{\mathcal{V}}$, where  $\hat{\mathcal{V}} = \{v: v=\hat{B}^\top x, \, x \in \mathcal{X} \} \subseteq \mathbb{R}^{\hat{u}}$. The decision functions in these reduced spaces are denoted by $\tilde{f}^{V}: \mathcal{V} \to \mathbb{R}$ and $\tilde{f}^{\hat{V}}: \hat{\mathcal{V}} \to \mathbb{R}$, respectively. With a slight abuse of notation, we use $d(\cdot)$ to denote the decision rule defined on either $\mathcal{V}$ or $\hat{\mathcal{V}}$, depending on the context.

\section{Method}\label{sec:methods}
In this section, we present our proposed methodology. Section~\ref{sec:gKDR} introduces a gradient kernel dimension reduction for identifying a low-dimensional subspace that captures the dependence of the contrast between potential outcomes on the covariates. We begin by constructing a suitable pseudo-outcome and then develop the gKDR framework for this variable. Section~\ref{sec:KCB_FB} describes how the estimates obtained via KCB can be integrated into our SDR procedure to account for differences in covariate distributions across treatment groups. Finally, Section~\ref{sec:AOL} outlines how AOL can be applied within the identified sufficient subspace of the covariates.

\subsection{Gradient kernel dimension reduction for the difference between potential outcomes}\label{sec:gKDR}

\subsubsection{Construction of the pseudo-outcome}
\label{subsub:pseudo_outcome}

The value function (\ref{eq:value}) is maximized when the decision function has the same sign as $\mathbb{E}[Y(+1) - Y(-1)|X=x]$. Hence the conditional ATE characterizes the Bayes-optimal decision boundary \parencite{zhao_estimating_2012}. \textbf{Assumption 3} ensures that SDR with respect to $\mathbb{E}[Y(+1) - Y(-1)|X=x]$ enables the identification of a low-dimensional subspace of the domain of the optimal decision rule. Under \textbf{Assumptions 1–3}, we can identify $\mathrm{span}(B_0)$ by noting that $\mathbb{E}[ AY/\pi(A,X) \mid B^\top _0X ]= \mathbb{E}[ \mathbb{E}\{ AY/\pi(A,X) \mid X \} \mid B^\top _0 X ]= \mathbb{E}[Y(+1) - Y(-1) \mid B^\top _0 X] = \mathbb{E}[Y(+1) - Y(-1) \mid X]$. Therefore, the problem reduces to finding a matrix $B_0$ such that $\mathbb{E}[AY / \pi(A,X) \mid X] = \mathbb{E}[AY / \pi(A,X)\mid B^\top_0 X]$. To facilitate estimation, we introduce a function $g$ defined as the projection of $\{1/\pi(A, X) - 1\}Y$ onto the linear span of $X$, evaluated at $x$:
\begin{equation}
    g(x) = \mathbb{E}\left[ x^\top (\mathbb{E}[X X^\top])^{-1}X \{1/\pi(A, X) - 1\}Y \right].
    \label{eq:g_def}
\end{equation}
Throughout the paper, we assume that $\mathbb{E}[XX^\top]$ is positive definite and $\mathbb{E}[X\{1/\pi(A,X) - 1\}Y]$ exists, so that $g(x)$ is well-defined. As motivated in Section \ref{sec:AOL}, we then define the pseudo-outcome $Z = A\{Y - g(X)\}/\pi(A,X)$, which can be interpreted as a contrast between weighted residuals. Since $\mathbb{E}\left[ A g(X) /\pi(A,X) \mid X\right] = g(X) \mathbb{E}\left[ A/\pi(A,X) \mid X\right] = 0$, \textbf{Assumption 3} implies $\mathbb{E} \left[ Z \mid B_0^\top X \right] = \mathbb{E} \left[ \mathbb{E}[Z \mid X] \mid B_0^\top X \right] = \mathbb{E}\left[ Y(+1) - Y(-1) \mid X \right]$. Next, we describe our approach for estimating $\mathrm{span}(B_0)$, and in Section~\ref{sec:gx_Estimation} we outline our procedure for estimating $g(x)$.

\subsubsection{Gradient kernel dimension reduction}
Gradient-based approaches have been widely used for developing SDR methods, including IADE \parencite{hristache_direct_2001} and wOPG \parencite{kang_forward_2022}. Here, we extend gKDR \parencite{fukumizu_gradient-based_2014} to our setting, targeting estimation of the central mean subspace using pseudo-outcomes. Let $\mathcal{H}_{\mathcal{X}}$ denote the RKHS of real functions $f: \mathcal{X} \to \mathbb{R}$, equipped with a reproducing kernel $K_X: \mathbb{R}^p \times \mathbb{R}^p \to \mathbb{R}$ inducing an inner product $\langle \cdot,\cdot \rangle_{\mathcal{H}_{\mathcal{X}}}$. Let $P_{X,Z}$ be the joint distribution of $(X,Z)$. Then, for any $x \in \mathcal{X}$, $\mathbb{E}[K_X(x,X) Z] = \int K_X(x, x') \int z' dP_{Z|X}(z' | x') dP_{X}(x') = \int K_X(x, x') \mathbb{E}[Z|X = x']  dP_{X}(x')$,
where $P_{Z|X}$ and $P_X$ are the distributions of $Z|X$ and $X$, respectively. Under the additional assumption that $\mathbb{E}[Z|X = x] \in \mathcal{H}_{\mathcal{X}}$, we have $\left( C^{-1}_{XX} \mathbb{E}[K_X(\cdot, X) Z] \right)(x) = \mathbb{E}[Z|X = x]$,
where $C^{-1}_{XX}$ is the inverse of the covariance operator $C_{XX}: \mathcal{H}_{\mathcal{X}} \to \mathcal{H}_{\mathcal{X}}$, which admits the integral operator representation $(C_{XX} f)(x) = \int K_{X}(x,x') f(x') dP_X(x').$ Then, 
\begin{equation}
\label{eq:AlternRKHS}
  \mathbb{E}\left[Z|X=x \right] = \left\langle \mathbb{E}[Z|X=\cdot], K_X(\cdot, x) \right\rangle_{\mathcal{H}_{\mathcal{X}}} = \left\langle C^{-1}_{XX} \mathbb{E}[K_X(\cdot, X) Z], K_X(\cdot, x) \right\rangle_{\mathcal{H}_{\mathcal{X}}} 
\end{equation}
by the reproducing property. We use the gradient of $\mathbb{E}[Z|X=\cdot] $ to estimate $B_0$ by noting 
\begin{equation}\label{eq:grad1}
    \frac{\partial}{\partial x^{(m)}} \mathbb{E}\left[Z | X=x\right] = \sum_{a \in [u]} B_{0,ma} \frac{\partial}{\partial v^{(a)}} \mathbb{E}[Z | B_0^\top X = v], 
\end{equation}
where $B_{0,ma}$ is the $(m,a)$-th entry of $B_0$. Moreover, in the Supplementary Materials, we show that from Equation \eqref{eq:AlternRKHS} and Assumption (A6) (also defined in the Supplementary Materials), we have that
\begin{equation}
    \frac{\partial}{\partial x^{(m)}} \mathbb{E}\left[Z | X=x\right] = \left\langle C^{-1}_{XX} \mathbb{E}[K_X(\cdot, X) Z], \frac{\partial K_X(\cdot, x)}{\partial x^{(m)}} \right\rangle_{\mathcal{H}_{\mathcal{X}}} = \mathbb{E}\left[ C^{-1}_{XX} \frac{\partial K_X(X, x)}{\partial x^{(m)}} Z \right].
\label{eq:gradient}
\end{equation}
Equating \eqref{eq:grad1} and \eqref{eq:gradient} yields $\sum_{a \in [u]} B_{0,ma} \partial \mathbb{E}[Z | B_0^\top x = v]/ \partial v^{(a)}  = \mathbb{E}\left[ C^{-1}_{XX} \partial K_X(X, x)/\partial x^{(m)} Z \right]$. Next, define the matrix $M(x) \in \mathbb{R}^{p \times p}$ with entries
\begin{align}
    M_{mj}(x) &= \left\{ \sum_{a \in [u]} B_{0,ma} \frac{\partial}{\partial v^{(a)}} \mathbb{E}[Z | B_0^\top X = v] \right\} \left\{ \sum_{a \in [u]} B_{0,ja} \frac{\partial}{\partial v^{(a)}} \mathbb{E}[Z | B_0^\top X = v] \right\} \nonumber\\
        &=\mathbb{E}\left[ C^{-1}_{XX} \frac{\partial K_X(X, x)}{\partial x^{(m)}} Z \right] \cdot \mathbb{E}\left[ C^{-1}_{XX} \frac{\partial K_X(X,x)}{\partial x^{(j)}} Z \right]. \nonumber
\end{align}
Then, $M(x) = B_0 D(x) B^\top_0$, where $D(x) \in \mathbb{R}^{u \times u}$ is a symmetric matrix with entries $D_{ab}(x) = \partial \mathbb{E}[Z|B_0^\top x = v] / \partial v^{(a)} \cdot \partial \mathbb{E}[Z|B_0^\top x = v]/\partial v^{(b)}. $ Since, $\mathbb{E}[M(X)] = B_0 \mathbb{E}[D(X)] B_0^\top$, a basis for $\mathrm{span}(B_0)$ can be obtained from the leading $u$ eigenvectors of $\mathbb{E}[M(X)]$.

\subsubsection{Gradient kernel dimension reduction with balancing weights}
\label{sec:gKDR_balancing}
In practice, constructing the empirical counterpart of the pseudo-outcome requires estimating the inverse propensity score. While this is trivial in randomized settings, the estimates of the propensity score in observational studies can substantially affect the results of the analysis. Although the proposed SDR framework is compatible with any plug-in estimator of the propensity score, in this work, we adopt a kernel-based covariate balancing approach and integrate the resulting balancing weights $\hat{w}_i$ in our SDR and ITR frameworks. This yields the empirical pseudo-outcome $\tilde{Z}_i = \hat{w}_i A_i \{Y_i - g_{\hat{w}}(X_i)\}$, where $\hat{w}_i$ and $g_{\hat{w}}(X_i)$ are defined in Section \ref{sec:KCB_FB}. Once $\{\tilde{Z}_i;~ i \in [n]\}$ are obtained, for any $x \in \mathcal{X}$, we estimate $M(x)$ using $\hat{W}(x)$, whose $(m,j)$-entry is given by:
\[
\hat{W}_{mj}(x) = \left\{ \frac{1}{n} \sum_{i \in [n]} (\hat{C}_{XX} + \epsilon_n I)^{-1} \frac{\partial K_X(X_i, x)}{\partial x^{(m)}} \tilde{Z}_i \right\} \left\{ \frac{1}{n} \sum_{i \in [n]} (\hat{C}_{XX} + \epsilon_n I)^{-1} \frac{\partial K_X(X_i, x)}{\partial x^{(j)}} \tilde{Z}_i \right\},
\]
where \( I \) is the identity operator, \( \epsilon_n I \) is a Tikhonov regularization \parencite{nashed_theory_1986}, and $\hat{C}_{XX}$ is the empirical covariance operator such that $(\hat{C}_{XX} f)(x) = n^{-1} \sum_{i \in [n]} K_X(x, X_i) f(X_i)$.

Finally, we estimate \( \mathbb{E}[M(X)] \) by the empirical average $\tilde{W}_n = n^{-1} \sum_{i \in [n]} \hat{W}(X_i)$.
In Theorem~\ref{thm:MatrixConv}, we show that $ \tilde{W}_n$ converges in probability to $\mathbb{E}[M(X)]$, for every $x \in\mathcal{X}$. The top \( \hat{u} \) eigenvectors of \( \tilde{W}_n \) are used to obtain \( \hat{B} \). From Corollary~\ref{cor:eigenvectorConv}, and under appropriate identifiability conditions to resolve invariances, it follows that \( \hat{B} \to_p B_0 \) in Frobenius norm.

\begin{remark}
A limitation of gKDR is that the rank of the estimated matrix $\tilde{W}_n$, and therefore $ \hat B$, is bounded by that of the outer-product $\tilde{Z} \tilde{Z}^\top$, where $\tilde{Z} = (\tilde{Z}_1, \cdots, \tilde{Z}_n)^\top$. The latter may be rank-deficient \parencite{fukumizu_gradient-based_2014}, for example, in randomized studies with binary outcomes. In such settings, the gKDR-v variant can be used \parencite{fukumizu_gradient-based_2014}, which partitions the data, estimates $M(x)$ within each subset, and then aggregates the results to obtain $\tilde{W}_n$. In observational studies, however, the pseudo-outcomes $\tilde{Z}_i$ typically take many distinct values, so the standard gKDR procedure is usually sufficient.
\end{remark}

\subsection{Kernel-based covariate balancing}\label{sec:KCB_FB}
We propose using the KCB method of \textcite{wong_kernel-based_2018} to compute the balancing weights, $\{ \hat{w}_i : i \in [n] \}$. These appear both in the construction of the pseudo-outcome $\{\tilde{Z}_i\}$ and the objective function for learning the optimal ITR. We illustrate KCB for the treatment group $A = +1$, but equivalent results hold for $A = -1$. KCB is motivated by the moment condition $\mathbb{E}\left\{ m(X)\mathbbm{1}(A=+1)/\pi(+1,X) \right\} = \mathbb{E}\{m(X)\}$, for all measurable $m:\mathcal{X} \to \mathbb{R},$ which ensures that the mean of any \( m(X) \) matches the marginal mean of the potential outcome under \( A = +1 \). Introduce $\mathcal{H}_n = \{m \in \mathcal{H}_{\mathcal{X}}; \|m\|^2_n= n^{-1}\sum_{i \in [n]} m^2(X_i)=1 \}$. The empirical and penalized counterpart of this moment condition, used to estimate $\{\hat{w}_i: A_i = +1\}$, is then
\begin{equation}
    \min_{w \succeq 1} \left[ \sup_{m \in \mathcal{H}_n} \left\{ \left(\frac{1}{n} \sum_{i: A_i = +1} w_i m(X_i) -\frac{1}{n} \sum_{i \in [n]} m(X_i)\right)^2 - \lambda_1 \| m \|^2_{\mathcal{H}_{\mathcal{X}}} \right\} + \lambda_2 \frac{1}{n} \sum_{i: A_i = +1} w^2_i \right].
    \label{eq:objective}
\end{equation}
This enforces the moment condition above for all $ m:\mathcal{X} \to \mathbb{R}$ in a rich function space, with appropriate regularization. The weights $\{\hat{w}_i: A_i = -1\}$ can be obtained similarly. A key contribution of this work is Theorem~\ref{thm:WeightConv}, which establishes that the resulting weights satisfy $n^{-1} \sum_{i: A_i = +1} \{ \hat{w}_i - 1/\pi(A_i, X_i) \}^2 = o_p(1),$ under mild regularity conditions. Therefore, these weights provide a suitable notion of covariate balancing within our SDR framework and, as shown in Section~\ref{sec:Theory}, ensure universal consistency of the estimated decision rule based on the dimension-reduced covariate space. We also emphasize that the RKHS $\mathcal{H}_{\mathcal{X}}$ used for defining the covariate balancing weights need not be the same as the one used for gKDR.

\subsubsection{Estimation of $g(x)$}
\label{sec:gx_Estimation}
When $\pi(A,X)$ is known, we can estimate $g(x)$ in \eqref{eq:g_def} using $g_n(x) = n^{-1} \sum_{i \in [n]} x^\top (n^{-1}\mathbb{X}^\top \mathbb{X})^{-1}X_i (1/\pi(A_i, X_i) - 1) Y_i$, where $x^\top (n^{-1}\mathbb{X}^\top \mathbb{X})^{-1}X_i$ is the sample analogue of $x^\top (\mathbb{E}[X X^\top])^{-1}X$, and $\mathbb{X} \in \mathbb{R}^{n \times p}$ is the design matrix. We show $g_n(x)$ uniformly converges to $g(x)$ in probability in Section S2.1 of the Supplementary Materials. When $\pi(A,X)$ is not known, we propose using $g_{\hat{w}}(x) = n^{-1} \sum_{i \in [n]} x^\top (n^{-1}\mathbb{X}^\top \mathbb{X})^{-1}X_i (\hat{w}_i - 1) Y_i$ in place of $g_n$. As with $g_n(x)$, we can show that $g_{\hat{w}}(x)$ converges uniformly in probability to $g(x)$ (see Corollary \ref{cor:gx_converge}).

\subsection{Estimation of optimal ITRs} \label{sec:AOL}
The optimal decision function that maximizes the value function (\ref{eq:value}) can be characterized as the minimizer of the risk $\mathcal{R}(f) = \mathbb{E}\left[ Y/\pi(A, X) \mathbbm{1}\{\mathrm{sign} \circ f(X) \neq A\} \right]$. As shown in Section S8 of the Supplementary Materials, this can be equivalently written as
\begin{align}
    \mathcal{R}(f) &= \mathbb{E}\left[ |Y-g(X)|/\pi(A,X) \mathbbm{1}\{A \cdot \mathrm{sign}(Y-g(X)) \neq \mathrm{sign} \circ f(X)\} \right] \nonumber\\
        &~~ + \mathbb{E}\left[X^\top \right] \left\{\mathbb{E}[XX^\top] \right\}^{-1} \mathbb{E}[\pi(-A,X) XY / \pi(A,X)] + \mathbb{E}[\max\{-Y + g(X),0\} / \pi(A,X)] .\nonumber
\end{align}
Since the last two terms do not depend on the decision function $f(X)$, minimizing $\mathcal{R}(f)$ is equivalent to minimizing $\mathbb{E}[ |Y - g(X)|/\pi(A,X) \mathbbm{1}\{A \cdot \mathrm{sign}(Y - g(X)) \neq \mathrm{sign} \circ f(X)\} ]$. Moreover, given that the Bayes optimal decision rule is determined by the sign of $\mathbb{E}[Y(+1)-Y(-1)|X] = \mathbb{E}[Y(+1)-Y(-1)|B^\top _0 X]$, this loss can be equivalently minimized in the reduced space using a function $\tilde{f}^{V}:\mathcal{V} \to \mathbb{R}$. Replacing the discontinuous and nonconvex 0–1 loss with a convex surrogate loss $\phi(\cdot)$ and using the decision rule $\tilde{f}^V$ leads to the definition of the $(\phi, g)$-risk, $\mathcal{R}_{\phi,g}(\tilde{f}^V) = \mathbb{E}\left[ |Y - g(X)|/\pi(A,X) \phi\Big( A \cdot \mathrm{sign}\{Y - g(X)\} \cdot \tilde{f}^V(B_0^\top X)\Big) \right]$. Minimizing $\mathcal{R}_{\phi,g}(\tilde{f}^V)$ is then a convex optimization problem, therefore addressing the potential non-convexity that arises when negative values are observed for either $Y/\pi(A,X)$ or $\{Y-g(X)\} / \pi(A,X)$. %Note that the $(\phi, g)$-risk can take any function as an input. For example, when learning a decision rule on $\hat{\mathcal{V}}$, it takes $\tilde{f}^{\hat{V}}:\hat{\mathcal{V}} \to \mathbb{R}$ as input. 
Fisher consistency of $\phi(\cdot)$, meaning that any minimizer of the $(\phi, g)$-risk is also a minimizer of the original 0–1 loss, can be shown by using classical results \parencite{bartlett_convexity_2006, bach_learning_2024}.

Let $K_u: \mathbb{R}^{u} \times \mathbb{R}^{u} \to \mathbb{R}$ be a kernel function. Using the observed sample $(X_i, A_i, Y_i)$, balancing weights $\hat{w}_i$ and the estimator $g_{\hat{w}}(x)$, we define for any $\tilde{f}^{\hat{V}}: \hat{\mathcal{V}} \to \mathbb{R}$ in the RKHS induced by $K_u$,
\[
Q\left( \tilde{f}^{\hat{V}} \right) = \frac{1}{n} \sum_{i \in [n]} \hat{w}_i \left| Y_i - g_{\hat{w}}(X_i) \right| \, \phi \Big( A_i \cdot \mathrm{sign} \left\{ Y_i - g_{\hat{w}}(X_i) \right\} \tilde{f}^{\hat{V}}(\hat{B}^\top X_i) \Big) + \lambda_n \alpha^\top \hat{G} \alpha, %\|\tilde{f}^{\hat{V}} \|^2
\]
where $\tilde{f}^{\hat{V}}(\hat{B}^\top x) = \sum_{i \in [n]} \alpha_i K_u(\hat{B}^\top X_i, \hat{B}^\top x) + \alpha_0$, $\hat{G} \in \mathbb{R}^{n\times n}$ is the Gram matrix with $(i,j)$-th entry $K_u(\hat{B}^\top X_i, \hat{B}^\top X_j)$, $\alpha_0 \in \mathbb{R}$, and $\alpha = (\alpha_1, \alpha_2, \cdots, \alpha_n)^\top \in \mathbb{R}^n$. Finally, we estimate the decision function using $\tilde{f}^{\hat{V}}_n = \arg\min_{\tilde{f}^{\hat{V}}} Q\left(\tilde{f}^{\hat{V}}\right)$, and the corresponding decision rule is given by $\mathrm{sign} \circ \tilde{f}^{\hat{V}}_n (\hat{B}^\top x)$. 

In Lemma~\ref{lem:linTransConv}, we show that $\mathcal{R}_{\phi,g} (\tilde{f}^{\hat{V}}_n) - \inf_{\tilde{f}^{\hat{V}}} \mathcal{R}_{\phi,g} (\tilde{f}^{\hat{V}}) \to_p 0$. Let $f^*:\mathcal{X} \to \mathbb{R}$ and $\tilde{f}^*: \mathcal{V} \to \mathbb{R}$ be the Bayes optimal decision functions from which the Bayes rule is given by $d^*(x) = \mathrm{sign} \circ f^*(x)$ or $d^*(B_0^\top x) = \mathrm{sign} \circ \tilde{f}^*(B_0^\top x)$. Let $\mathcal{R}^* = \mathcal{R}(f^*)$ be the corresponding optimal risk. In Proposition~\ref{prop:UnivConsist}, we establish universal consistency of our estimator of the optimal decision rule, $\mathrm{sign} \circ \tilde{f}^{\hat{V}}_n (\cdot)$, that is, $\mathcal{R}(\tilde{f}^{\hat{V}}_n) \to_p \mathcal{R}^* $.

Note that indirect methods can also be applied once the covariates have been dimension-reduced. One may use Nadaraya-Watson regression to estimate the treatment effect heterogeneity and identify an optimal treatment. However, such methods often suffer in practice when the reduced subspace is not extremely low-dimensional (e.g., dimension lower than four). Dimension-reduced covariates can also be used in conjunction with more flexible learners, such as the R-learner framework \parencite{nie_quasi-oracle_2021}.

\section{Theoretical guarantees}\label{sec:theory}
\label{sec:Theory}
Here, we provide theoretical guarantees for the proposed approach to estimating optimal ITRs. The proofs are provided in the Supplementary Material (Sections S1-S7). We first show that the balancing weights $\{\hat{w}_i\}$, converge to the true IPWs in mean squared error, as stated in the following theorem.

\begin{theorem}
\label{thm:WeightConv}
Assume $\lambda_1 \asymp n^{-1}$, and let the KCB weights $\{\hat{w}_i: A_i = +1\}$ be the solution to the optimization problem in \eqref{eq:objective}. Then, under the regularity conditions $\mathrm{(A1)}- \mathrm{(A5)}$ in Section S1 of the Supplementary Material,
\[
\frac{1}{n} \sum\limits_{i: A_i=+1} \left\{\hat{w}_i - \frac{1}{\pi(+1,X_i)}\right\}^2 = o_p(1).
\]   
\end{theorem}
By a symmetric argument, $n^{-1} \sum_{i: A_i=-1} \left\{\hat{w}_i - 1/\pi(-1,X_i) \right\}^2 = o_p(1)$ for $\{\hat{w}_i: A_i = -1\}$, and therefore $n^{-1} \sum_{i \in [n]} \{\hat{w}_i - 1/\pi(A_i,X_i) \}^2 = o_p(1).$ To the best of our knowledge, no existing result establishes convergence of KCB weights to the true inverse propensity weights. In contrast, \textcite{wong_kernel-based_2018} show only that $\left| n^{-1} \sum_{i: A_i=+1} \hat{w}_i m(X_i) - n^{-1} \sum_{i \in [n]} m(X_i) \right| = O_p\left( n^{-1/2} \right)$.  

Theorem \ref{thm:WeightConv} implies that the difference between $g_{\hat{w}}(x)$ and $g(x)$ is uniformly negligible in probability over $x \in \mathcal{X}$, as formalized in the following corollary.
\begin{corollary}
Suppose $\mathcal{X}$ is compact and $\mathbb{E}[Y^2] < \infty$. Then, 
    \[ \sup_{x \in \mathcal{X}}|g_{\hat{w}}(x) - g(x)| = o_p(1). \]
\label{cor:gx_converge}
\end{corollary}

By using Theorem~\ref{thm:WeightConv} and Corollary \ref{cor:gx_converge}, we can establish the consistency of our estimator for $\mathbb{E}[Zf(X)]$, where $f:\mathcal{X} \to \mathbb{R}$ is any measurable function. This is an important building block for showing the convergence of the estimator of $\mathrm{span}(B_0)$ obtained from the proposed gKDR approach. This result is formalized in the following theorem.

\begin{theorem}
\label{thm:CrossCov}
Let $\{\hat{w}_i; i \in [n]\}$ denote the KCB weights, and recall the pseudo-outcomes $Z = A \{Y - g(X)\}/\pi(A,X)$ and $\tilde{Z}_i = \hat{w}_i A_i \{Y_i - g_{\hat{w}}(X_i)\}$. Then, for any measurable $f:\mathcal{X} \to \mathbb{R}$ on compact $\mathcal{X}$, such that $ \mathbb{E}[Z^2 f^2(X)] < \infty$ and $ \mathbb{E}[f^2(X)] < \infty$, we have
\[
\frac{1}{n} \sum\limits_{i \in [n]} \tilde{Z}_i f(X_i) \to_p \mathbb{E}[Z f(X)].
\]
%\frac{1}{n} \sum\limits_{i \in [n]} \hat{w}_i A_i Y_i f(X_i)  = \mathbb{E}[Z f(X)] + o_p(1)~~ \text{and}~~ 
\end{theorem}

The following theorem establishes the consistency of the empirical $(\phi,g)$-risk when the KCB weights are used. Intuitively, from Corollary \ref{cor:gx_converge}, we expect that the number of discrepancies between the signs of $Y_i - g(X_i)$ and $Y_i - g_{\hat{w}}(X_i)$ to be controlled.
\begin{theorem}
\label{thm:ConsistentValue}
Let $\{\hat{w}_i; i \in [n]\}$ be the KCB weights. Suppose $\mathcal{X}$ is compact and that the second moment of $|Y-g(X)|\phi\big(A \cdot \mathrm{sign}\{Y-g(X)\} f(X)\big)$ is bounded. Assume that the distribution $P$ of $(X,A,Y)$ satisfies $|Y - g(X)|/\pi(A,X) \leq M_g$ and $|f(X)| \leq M_f$ almost everywhere, where $M_g, M_f \in \mathbb{R}$ are sufficiently large constants. Assume further that $\phi(\cdot)$ is $L$-Lipschitz. Then, as long as 
$\sum_{i \in [n]} \mathbbm{1} \{ |Y_i - g(X_i)| < \epsilon \} \leq C n^{\xi}$, 
for some $\epsilon > 0$, constant $C$, and $\xi < 1$, we have
$$\frac{1}{n} \sum_{i \in [n]} \hat{w}_i |Y_i - g_{\hat{w}}(X_i)| \phi\Big( A_i \cdot \mathrm{sign}\{Y-g_{\hat{w}}(X_i)\} f(X_i) \Big) \to_p \mathcal{R}_{\phi, g}(f). $$
\end{theorem}

To establish that $\mathrm{span}(\hat{B})$ is a consistent estimator of $\mathrm{span}(B_0)$, we first show the convergence of $\hat{W}(x)$ to $M(x)$ and that of $\tilde{W}_n$ to $\mathbb{E}[M(X)]$. We additionally make assumptions (A6)–(A10), in the Supplementary Material, which are adapted from those in \textcite{fukumizu_gradient-based_2014}. The following theorem formalizes these statements.

\begin{theorem}
\label{thm:MatrixConv}
Assume $\mathbb{E}[Y^2 \{ C^{-1}_{XX} \partial K(x, X)/\partial x^{(a)} \}^2 ], ~ \mathbb{E}[g^2(X) \{ C^{-1}_{XX} \partial K(x, X)/\partial x^{(a)} \}^2 ] < \infty$ for any $x \in \mathcal{X}$ and $a \in [p]$, with $p$ is fixed. For a constant $\xi \geq 0$ suppose that $\partial K(\cdot, x) / \partial x^{(a)}$ is in the range of $C^{\xi + 1}_{XX}$. Then, under assumptions (A6)-(A10), for every $x \in \mathcal{X}$,
$$ \left\|\hat{W}(x) - M(x) \right\|_F = o_p(1) %+ O_{p}\left( n^{-\min\left\{ \frac{1}{3}, \frac{2 \xi + 1}{4 \xi + 4}\right\}} \right). 
,$$
Moreover, if $\mathbb{E}[\|M(X)\|^2_F] < \infty$ and $\partial K(\cdot,x)/\partial x^{(a)} = C^{\xi + 1}_{XX} h_{a,x}$, where $h_{a,x} \in \mathcal{H}_{\mathcal{X}}$ and $E \|h_{a,x}\|_{\mathcal{H}_{\mathcal{X}}} < \infty$, then
$$ 
\left\|\tilde{W}_n - \mathbb{E}[M(X)] \right\|_F =  o_p(1)%+ O_{p}\left( n^{-\min\left\{ \frac{1}{3}, \frac{2 \xi + 1}{4 \xi + 4}\right\}} \right)
.
$$
\end{theorem}
%\begin{remark}
When the true propensity scores are known, this result extends to settings in which the number of covariates $p$ grows with the sample size, thereby generalizing Theorem 4 of \textcite{fukumizu_gradient-based_2014} to our framework (see Remark S1 in the Supplementary Materials).

Theorem~\ref{thm:MatrixConv} implies the following corollary, which follows from Theorem 2 of \textcite{yu_useful_2015}, a variant of the Davis–Kahan $\sin \theta$ theorem \parencite{davis_rotation_1970}.

\begin{corollary}
\label{cor:eigenvectorConv}
Let $B_0$ and $\hat{B}$ be matrices whose columns are the eigenvectors corresponding to the $u$ largest eigenvalues of $\mathbb{E}[M(X)]$ and $\tilde{W}_n$, respectively. Suppose the non-zero eigenvalues of $\mathbb{E}[M(X)]$ are distinct, and that for each $j$, the $j$-th column $\hat{b}_j$ of $\hat{B}$ satisfies $\hat{b}_j^\top b_{0j} \geq 0$, where $b_{0j}$ is the $j$-th column of $B_0$. Then, $\left\| \tilde{W}_n - \mathbb{E}[M(X)] \right\|_F =  o_p(1)$ implies
$$ 
\left\|\hat{B}- B_0 \right\|_F = o_p(1).
$$
\end{corollary}
Assuming $\mathcal{X}$ is compact, then Corollary \ref{cor:eigenvectorConv} implies that $\hat{B}^\top x \to_pB_0^\top x$ for any $x \in \mathcal{X}$.

For the following analysis, we make a technical assumption similar to that in \textcite{wong_kernel-based_2018} and \textcite{athey_approximate_2018}, and assume $\hat{w}_i \leq M_w n^{1/3},~ i \in [n]$, where $M_w$ is a large constant. This upper bound constraint can be incorporated in the estimation of $\{\hat{w}_i: i \in [n]\}$ in \eqref{eq:objective}. %Motivated by this upper-bound constraint, we further assume that the distribution $P$ of $(X,A,Y)$ satisfies $|Y - g_{\hat{w}}(X)| / \pi(A,X) \leq M_g n^{1/3}$ almost everywhere, for some constant $M_g > 0$.

\begin{lemma}
\label{lem:linTransConv}
Assume that the surrogate loss function $\phi(\cdot)$ is Lipschitz continuous and the kernel satisfies $K_u(v, v) < \infty$ for all $v \in \mathbb{R}^u$. Suppose $\mathcal{X}$ is compact, and $\lambda_n > 0$ with $\lambda_n \to 0$ and $n^{1/3} \cdot \lambda_n \to \infty$ as $n \to \infty$. Let $\alpha^{\hat{V}}_{0n}$ be the estimated intercept of $\tilde{f}^{\hat{V}}_n$. Additionally, assume that and the distribution $P$ of $(X, A, Y)$ satisfies $|Y - g(X)| / \pi(A,X) \leq M_g$ and $|\sqrt{\lambda_n} \, \alpha^{\hat{V}}_{0n}| \leq M_{\alpha_0} < \infty$ almost everywhere, and the second moment of $\phi\Big( A \cdot \mathrm{sign}\{Y-g(X)\} \tilde{f}^{\hat{V}}(\hat{B}^\top X) \Big)$ is bounded. Furthermore, suppose that for a sufficiently large constant $M_w$,  $\hat{w}_i \leq M_w n^{1/3}$ for all $i \in [n]$. Then,
\[
\mathcal{R}_{\phi, g}(\tilde{f}_n^{\hat{V}}) \to_p \inf_{\tilde{f}^{\hat{V}}} \mathcal{R}_{\phi, g}(\tilde{f}^{\hat{V}})
\]
\end{lemma}

By combining this result with Lemma S3 and Lemma S4 in the Supplementary Material, we establish the universal consistency of the ITR learned by the proposed method under the hinge loss. This is formalized in the following proposition.

\begin{proposition}
\label{prop:UnivConsist}
Suppose that the RKHS of $\tilde{f}^{\hat{V}}:\hat{\mathcal{V}} \to \mathbb{R}$ is induced by a universal kernel, and $\mathcal{X}$ is a compact metric space. Let $\phi(\cdot)$ be the hinge loss function. Assume that the conditions stated in Lemma \ref{lem:linTransConv} hold. Then,
\[\mathcal{R}( \tilde{f}^{\hat{V}}_n ) \to_p \mathcal{R}^*.
\]
\end{proposition}

\section{Simulations}\label{sec:simulations}
In this section, we evaluate the finite-sample performance of the proposed dimension-reduced outcome-weighted learning, which here we refer to as DOL, across three settings. The data were generated in both randomized and non-randomized scenarios, where $\Pr(A = +1 \mid X) = \Pr(A = -1 \mid X) = 1/2$ in the randomized setting. We compare the DOL against four methods: (AOL) the augmented outcome-weighted learning by \textcite{zhou_augmented_2017}, which uses the full covariate set $X$; (SI) constrained single-index regression \parencite{park_constrained_2021}; (QL) an $\ell_1$-penalized partial least squares ($\ell_1 \text{-} PLS$) Q-learning method \parencite{qian_performance_2011}; and (RL) R-learner using Gaussian kernel ridge regression \parencite{nie_quasi-oracle_2021}.

In the non-randomized setting, we additionally compared two covariate balancing strategies for the proposed method: IPW estimated via logistic regression, and KCB weights. The same balancing weights were applied for both the gKDR and the ITR estimation.%(augmented) outcome-weighted learning approach.

Performance was evaluated using accuracy and the value function. Accuracy measures the proportion of the learned rules that agree with the Bayes optimal rule. In each setting, a test set of 10,000 observations, was generated. For each Monte Carlo iteration, a training set was generated, the decision rule was estimated, and performance was evaluated by using the test set (see Section S9.1 of Supplementary Material). The hinge loss, $\phi(x) = \max(0, 1-x)$, was used as the surrogate loss in AOL and the proposed DOL. In both methods, the penalty term $\lambda_n$ was selected from a grid of candidate values by choosing the one that yielded the largest two-fold cross-validated value function. Further details on the software implementation and estimation of ITRs are deferred to Section S9.2 of Supplementary Material.

Additional simulation results to show the decreasing trend in the projection errors along increasing sample size is presented in Section S10.2 of Supplementary Material.

\subsection{Simulation settings}
We considered $n = 500$ and $n =1{,}000$, with all settings using $p = 50$. We randomly generated a latent factor $Z \in \mathbb{R}^{50}$ from a uniform distribution over the range $[-2, 2]$. This was then used to generate the observed covariates $X \in \mathbb{R}^{50}$. With the main effect, $\mu(X)$, and the treatment-covariate interaction, $\tilde{f}^V(V)$, the outcomes were generated from $N\{\mu(X) + A \cdot \tilde{f}^V(V), 1\}$. Further details of simulation settings are in Section S9.3 of Supplementary Material.

\begin{figure}
\centering
\includegraphics[width = 5.5in]{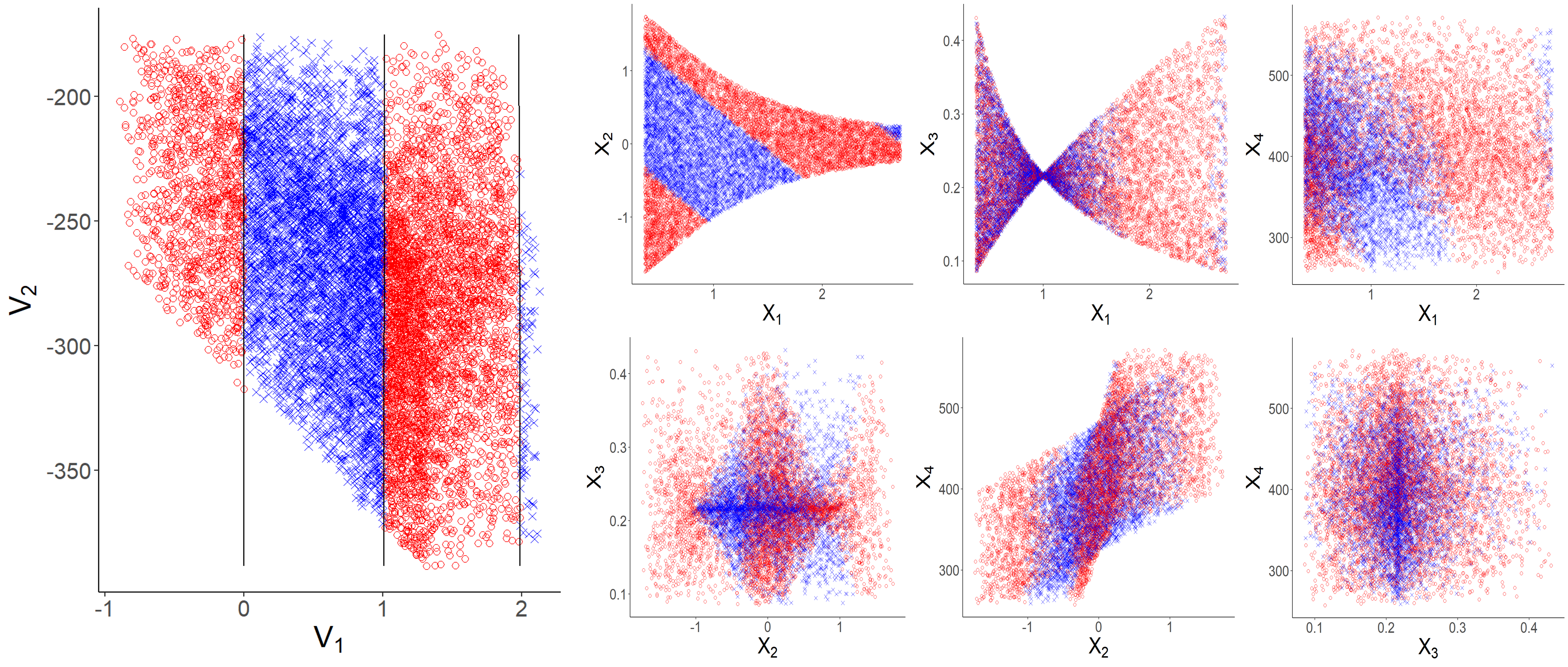}
\caption{Scatterplot of the Bayes optimal treatment regimes for different values of the covariates generated under Setting 1. The red circles and blue crosses represent data points from the test dataset, corresponding to $d^*(X) = -1$ (or $d^*(V_0) = -1$) and $d^*(X) = +1$ (or $d^*(V_0) = +1$), respectively. The solid lines in the left plot display the decision boundaries in the dimension-reduced space, defined as $\left\{v = \left(v^{(1)}, v^{(2)}\right) : \tilde{f}^*(v) = 0 \right\}$.
}
\label{fig:Setting1}
\end{figure}

Setting 1 considers $V = B^\top _0 X \in \mathbb{R}^2$. The decision boundaries over the original covariates $X$ and the reduced subspace are shown in Figure~\ref{fig:Setting1}. %The boundaries of the Bayes regimes are clearly defined in the reduced representation and when $X^{(1)}$ and $X^{(2)}$ are considered, but not for any other pairs of $X$.
Setting 2 also considers $V = B^\top _0 X \in \mathbb{R}^2$. The decision boundaries are shown in Figure S1 of the Supplementary Material. This scenario introduces additional complexity due to the non-smoothness of the decision boundary.
%In this scenario, the boundaries of the optimal treatment regimes are clearly defined in the reduced representation and for $X^{(1)}$ and $X^{(3)}$, but show additional complexity due to the non-smoothness of the decision boundary. No clear boundaries can be observed for any other pairs of covariates in $X$. 
Setting 3 considers higher-dimensional reduced subspace where $V = B^\top _0 X \in \mathbb{R}^4$. This setting introduces two pairs of highly correlated covariates in the reduced subspace.

\subsection{Results}
\begin{figure}
\centering
\includegraphics[width = 5.5in]{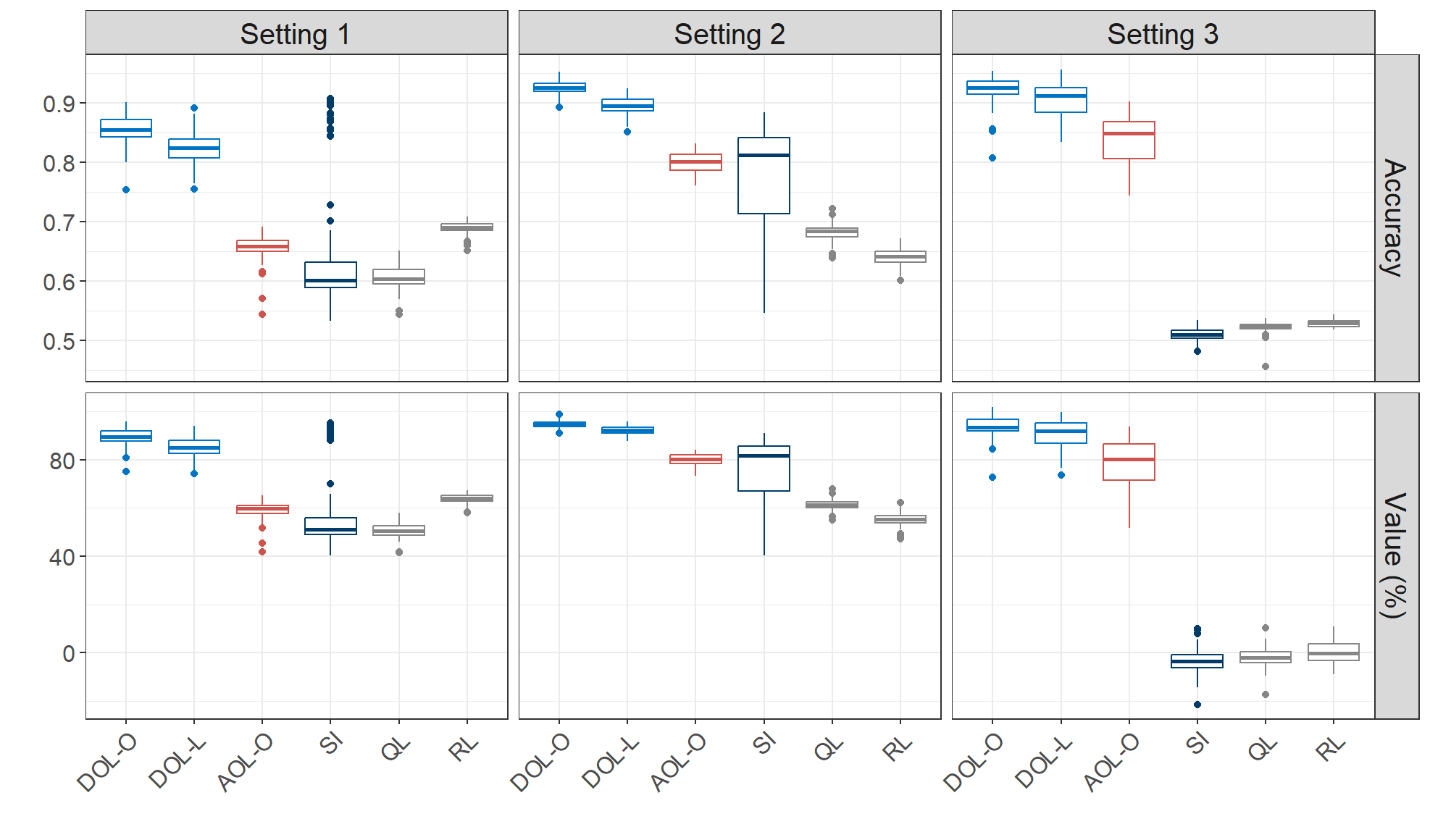}
\caption{Simulation results for $n = 1{,}000$ from the randomized scenarios across the three different settings. Accuracy represents the proportion of correctly predicted optimal treatments. Value (\%) represents the value function recovered by the estimated decision rule as a percentage of the Bayes optimal value function. The performance of the proposed method is shown under DOL-O and DOL-L: DOL-O uses the oracle $g(x)$, while DOL-L uses $g_{\hat{w}}(x)$ estimated via linear regression.
}
\label{fig:Random}
\end{figure}

We show the results for $n = 1{,}000$. The results for $n = 500$ are provided in Supplementary Material (Section S10.1), showing similar trends. Figure~\ref{fig:Random} displays the results for the randomized scenario. DOL-O refers to the proposed method using the oracle $g(x)$ defined as $\mathbb{E}[\{1/\pi(A,X)-1\}Y|X=x]$ which is an alternative formulation of $g(x)$ originally proposed to be used in the AOL by \textcite{zhou_augmented_2017}. DOL-L uses $g_{\hat{w}}(x)$ that estimates $g(x)$ with linear regression as described in Section \ref{sec:gx_Estimation}. AOL-O represents the estimator from AOL that uses the oracle $g(x)$. From Figure~\ref{fig:Random}, the DOL outperforms the existing methods. DOL-L performs close to DOL-O, indicating that the proposed method that uses $g_{\hat{w}}(x)$ performs close to using the true form of the oracle $g(x)$.

\begin{figure}
\centering
\includegraphics[width = 5.5in]{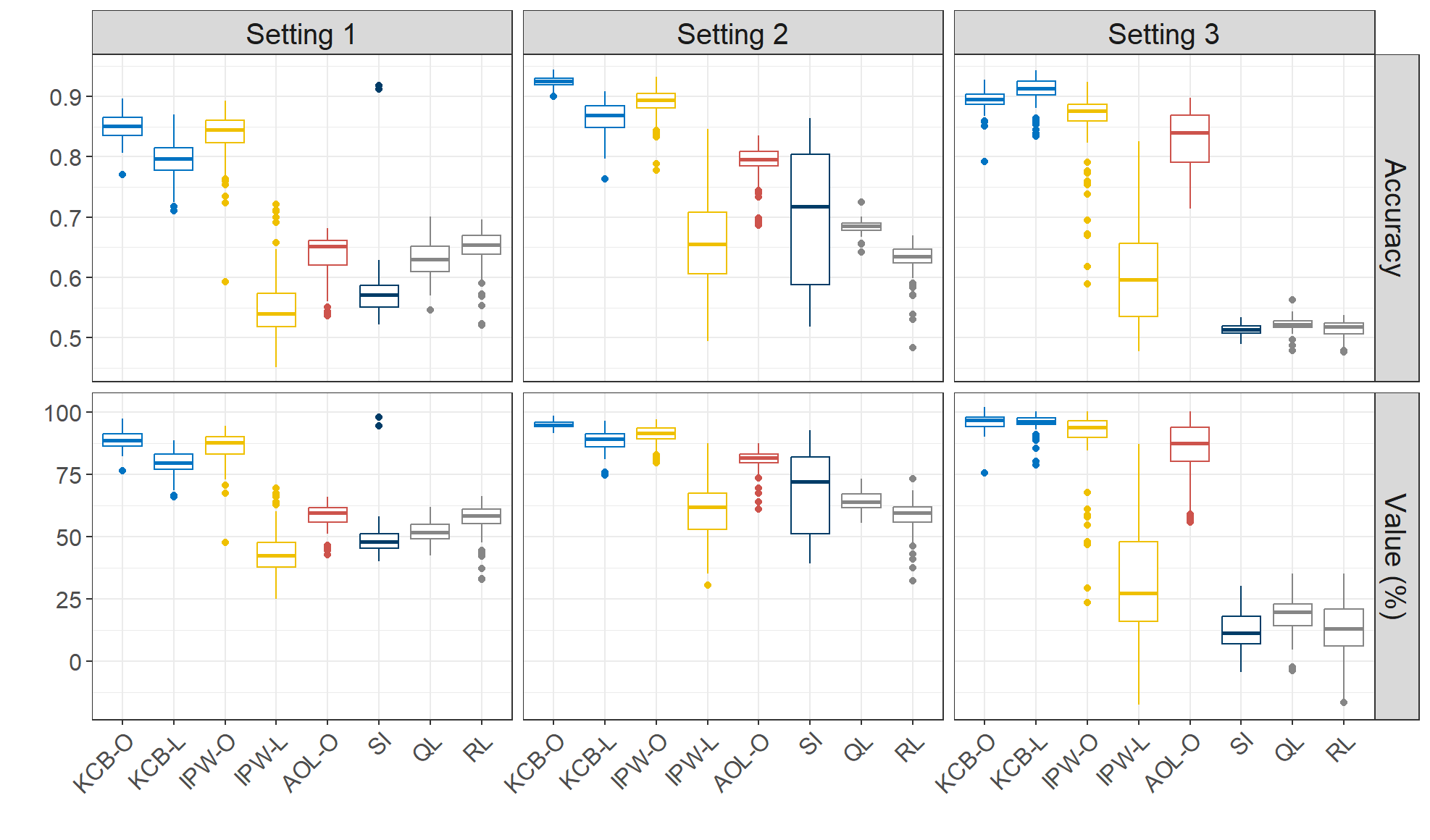}
\caption{Simulation results for $n = 1{,}000$ from the non-randomized scenarios across the three different settings. Accuracy represents the proportion of correctly predicted optimal treatments. Value (\%) represents the value function recovered by the estimated decision rule as a percentage of the Bayes optimal value function. The performance of the proposed method is shown under KCB-O and KCB-L: KCB-O uses the oracle $g(x)$, while KCB-L uses the fitted values of $g_{\hat{w}}(x)$ obtained via linear regression. IPW-O uses the oracle $g(x)$, and IPW-L uses $g_{\hat{w}}(x)$ from linear regression, with propensity scores estimated via logistic regression.
}
\label{fig:NonRandom}
\end{figure}

The results for the non-randomized scenario are in Figure~\ref{fig:NonRandom}. We considered two cases for the DOL: the estimator that uses the oracle $g(x)$ (KCB-O), and another where $g_{\hat{w}}(x)$ is obtained via linear regression (KCB-L). For the IPW, we also considered two cases: using the oracle $g(x)$ (IPW-O), and using $g_{\hat{w}}(x)$ (IPW-L). AOL was implemented using the true propensity scores and the oracle $g(x)$.
First, we can observe that using KCB weights (KCB-O) outperforms using inverse propensity scores from a misspecified logistic regression model (IPW-O). IPW-O tends to achieve higher accuracy and value function than KCB-L on occasion which is expected as the AOL is a doubly robust method, so that using the oracle $g(x)$ suffices for universal consistency. However, it shows unstable performance and is outperformed by KCB-L in Setting 3. IPW-L shows notable declines in both accuracy and value function when the $g(x)$ is estimated by using the propensity scores estimated from a misspecified logistic regression model. 

KCB-L demonstrates very good performance across all three settings. This suggests that KCB is an effective weighting strategy for use in non-randomized studies. Despite using $g_{\hat{w}}(x)$ which is an estimator of $g(x)$, KCB-L outperformed AOL-O, which used the true form of the oracle $g(x)$ and the true inverse propensity weights. In Setting 3, KCB-L achieves higher accuracy compared to KCB-O; however, this improvement does not translate into a much higher value function.

\section{Application}\label{sec:application}
We illustrate our proposed method by investigating ITR recommendations for using transthoracic echocardiography (TTE) in intensive care unit (ICU) sepsis patients. We retrieved data on 6,361 patients from the MIMIC-III database \parencite{johnson_mimic-iii_2016}, as described in \textcite{feng_transthoracic_2018}. In the dataset, TTE was performed on 3,262 patients (51.3\%) during or within 24 hours before ICU admission, and the remaining 3,099 (48.7\%) did not receive TTE. The outcome is binary, indicating 28-day survival (1 if survived, 0 if not). We used 35 baseline variables (see Section S11 of Supplementary Material). Missing values were imputed by using MissForest \parencite{stekhoven_missforestnon-parametric_2012}.  

To estimate an optimal ITR, we used the proposed DOL, the AOL, $\ell_1 \text{-} PLS$ (QL), and the decision rule that prescribes TTE for every patient (Treat all), which represents the average treatment effect of TTE. We randomly sampled 1,000 subjects without replacement to construct a training set to learn ITRs. We used the remaining 5,361 subjects as a validation set by comparing the 28-day survival rates between the treatments (TTE or non-TTE) that match the estimated decision rule and those that differ. We repeated this procedure over 100 Monte Carlo replications. At each iteration, we evaluated the modified value function \parencite{chen_robust_2024}, $\mathbb{E}[Y\{d(X)\} - Y\{-d(X)\}]$, with the ITR estimated from the training set. %Note that this represents the average effect of a decision rule. 
In our data, this indicates the improvement in the 28-day survival rate. See Supplementary Material (Section S11.1) for the details on learning ITR and the evaluation of the modified value function.

\begin{figure}
\centering
\includegraphics[width = 4in]{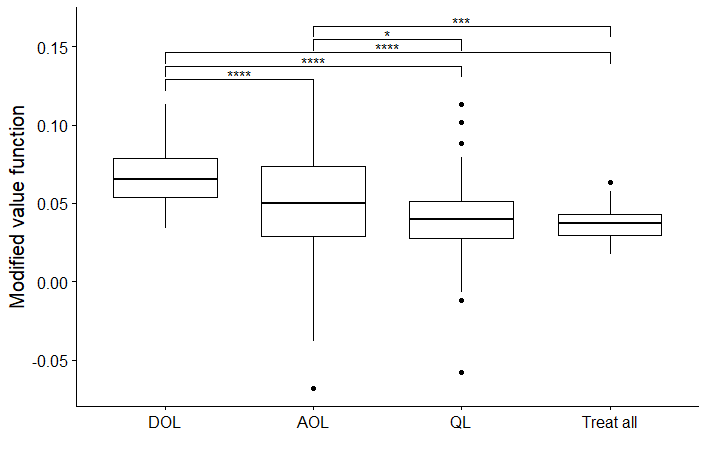}
\caption{Modified value functions over the 100 repeats. DOL represents the proposed method. AOL used the raw patient characteristics for learning ITR and fitting logistic regression to estimate IPW. QL is the $\ell_1 \text{-} PLS$. ``Treat all'' represents the average effect of TTE. P-values from paired $t$-tests are displayed using asterisks: **** $<0.0001$; *** $<0.001$; * $< 0.05$.}
\label{fig:Real}
\end{figure}

The average modified value function of treating all patients with TTE is around 3.69\%, and the modified value functions are substantially higher in the AOL and the DOL results, with averages of 4.95\% and 6.72\%, respectively (Figure \ref{fig:Real}). This suggests the existence of treatment effect heterogeneity and that treatment assignment based on a decision rule that considers the individual patient characteristics could improve the survival rate. %There was no notable difference between the $\ell_1 \text{-} PLS$ (QL) and treating all patients with TTE. This may be attributed to the underlying nonlinearity in the true association between 28-day survival and patient characteristics. 
The proposed method further improves upon AOL, which demonstrates the benefit of dimension reduction combined with KCB weighting for learning ITR in observational studies. 

\section{Conclusion}

We propose a novel gradient kernel dimension reduction approach that identifies a sufficient subspace for modeling heterogeneous treatment effects, enabling more accurate estimation of optimal ITRs using direct approaches. To account for differences in covariate distributions across treatment groups, we incorporate covariate balancing weights from the KCB method. This allows treatment assignment to depend on the full set of covariates. We then formulate the problem of learning an optimal ITR as a weighted SVM problem on the reduced representation of the covariates.

Theoretical guarantees are provided for the proposed procedure, which required, among other results, a new bound on the mean squared error of the balancing weights from the KCB method and a generalization of gKDR to the setting of a pseudo-outcome defined using estimated weights.
In simulation studies, we demonstrated that the proposed method is effective and exhibits excellent performance in both randomized and observational settings. The real data analysis with ICU sepsis patients corroborates the simulation results. 

The proposed approach assumes that the data contain no missing values. A possible extension of this work is to adapt the proposed method, and specifically gKDR, to handle missing data without relying on imputation. Other extensions include accommodating multi-valued treatments and dynamic treatment regimes.

\section*{Supporting Information}

Refer to \url{https://github.com/stson327/DOL} for software implementation of the proposed method. \vspace*{-8pt}

\printbibliography

\clearpage
\setcounter{section}{0}
\setcounter{equation}{0}
\setcounter{figure}{0}
\setcounter{table}{0}
\setcounter{theorem}{0}
\setcounter{remark}{0}

\section*{Supplementary Material}
\addcontentsline{toc}{section}{Supplementary Material}
\renewcommand{\thesection}{S\arabic{section}}
\renewcommand{\theequation}{S\arabic{equation}}
\renewcommand{\thefigure}{S\arabic{figure}}
\renewcommand{\thetable}{S\arabic{table}}

\renewcommand{\theremark}{S\arabic{remark}}

\section{Proof of Theorem 1}
For simplicity, here we use $\mathcal{H}$ in place of $\mathcal{H}_{\mathcal{X}}$ and $K: \mathcal{X} \times \mathcal{X} \mapsto \mathbb{R}$ in place of $K_X: \mathcal{X} \times \mathcal{X} \mapsto \mathbb{R}$. Furthermore, define $S_1 = \{i \in [n]; A_i = +1 \}$. We first recall the regularity conditions and the statement of the theorem below.
\begin{description}
    \item[(A1)] The covariate space $\mathcal{X}$ is compact, and the covariates have a continuous density supported on $\mathcal{X}$.
    \item[(A2)] The propensity score functions $\pi(+1, x)$ and $\pi(-1, x)$ are continuous in $x \in \mathcal{X}$.
    \item[(A3)] The kernel $K_X(x_1, x_2)$ defining $\mathcal{H}$ is continuous, non-negative, bounded, and satisfies $K(x, x) > 0$ for all $x \in \mathcal{X}$.
    \item[(A4)] The RKHS $\mathcal{H}$ is dense in the space of continuous functions on $\mathcal{X}$ with respect to the supremum norm.
    \item[(A5)] The unit ball $\mathcal{H}_1 = \{ m \in \mathcal{H} : \|m\|_{\mathcal{H}} \leq 1 \}$ is a $\mathbb{P}$-Donsker class.
\end{description}

\begin{theorem}
Assume $\lambda_1 \asymp n^{-1}$, and let the KCB weights $\{\hat{w}_i: A_i = +1\}$ be the solution to the optimization problem \eqref{eq:objective}
    $$\min_{w \succeq 1} \left[ \sup_{m \in \mathcal{H}_n} \left\{ \left(\frac{1}{n} \sum_{i \in S_1} w_i m(X_i) -\frac{1}{n} \sum_{i \in [n]} m(X_i)\right)^2 - \lambda_1 \| m \|^2_{\mathcal{H}} \right\} + \frac{\lambda_2}{n} \sum_{i \in S_1} w^2_i \right],$$
where $\mathcal{H}_n = \{m \in \mathcal{H}; \|m\|^2_n= n^{-1}\sum_{i \in [n]} m^2(X_i)=1 \}$. Then, under the regularity conditions $\mathrm{(A1)}- \mathrm{(A5)}$,
\[
\frac{1}{n} \sum\limits_{i \in S_1} \left\{\hat{w}_i - \frac{1}{\pi(+1,X_i)}\right\}^2 \longrightarrow 0,
\]
in probability, as $n \rightarrow \infty$.    
\end{theorem}

Define $m_n \in \mathcal{H}_n$ as $m_n = m / \|m\|_n$ for any $m \in \mathcal{H}$ such that $m \neq 0$. Without loss of generality, assume that the Gram matrix $A$, derived from the kernel $K: \mathcal{X} \times \mathcal{X} \mapsto \mathbb{R}$, is positive definite. By eigendecomposition, $A = \Gamma \Lambda \Gamma^\top$. From this, 
$$\|m_n\|^2_{\mathcal{H}} = \alpha^\top A \alpha = n \cdot (n^{-1/2} \Lambda \Gamma^\top \alpha)^\top \Lambda^{-1} (n^{-1/2} \Lambda \Gamma^\top \alpha) = n \cdot \beta^\top \Lambda^{-1} \beta,$$
where $\alpha$ is a vector of coefficients and $\beta = n^{-1/2} \Lambda \Gamma^\top \alpha$. Note that $\beta^\top \beta = n^{-1} \alpha^\top \Gamma \Lambda^2 \Gamma^\top \alpha = n^{-1} \alpha^\top A^2 \alpha = \|m_n\|^2_n = 1$. Therefore, $\|m_n\|^2_{\mathcal{H}} \leq n/\psi_{min}$, where $\psi_{min}$ is the minimum eigenvalue of $\Lambda$. Since $\lambda_1 \asymp n^{-1}$, it is bounded by $n^{-1}D_1 \leq \lambda_1 \leq n^{-1}D_2$ for constants $D_2 \geq D_1 > 0$. From this, $\lambda_1 \|m_n\|^2_{\mathcal{H}} \leq n^{-1}D_2 \cdot n / \psi_{min} = D_2 / \psi_{min}$. Therefore, we have the upper bound as
$$ \|m_n\|^2_{\mathcal{H}} \leq \frac{D_2}{\lambda_1 \psi_{min}}. $$
Let $c_1 = \sqrt{\lambda_1 \psi_{min} / D_2}$ and define $m_1 = c_1 \cdot m_n$. Then we have $\|m_1\|^2_{\mathcal{H}} \leq 1$ and $m_1 \in \mathcal{H}_1$. We present the proof for $m_1$, and the result can be extended to $m_n \in \mathcal{H}_n$ by scaling: $m_n= m_1/c_1$.

The proof can be written in three steps, similar to the proof of Theorem 1 of \textcite{chen_robust_2024}: characterization of the dual problem; finding the limiting form of the dual problem; and showing convergence in probability to the limiting form.

    \subsection{Step 1: Characterization of the dual problem}
Define an operator $M: \mathbb{R}^{n} \mapsto \mathcal{H}$ such that 
$$Mw = \sum_{i\in S_1} w_i K(X_i, \cdot), $$
where $S_1 = \{i; A_i = +1 \}$. Define $h = \frac{1}{n} \sum_{i \in [n]} K(X_i, \cdot)$. Since $m_1 \in \mathcal{H}_1$, 
\begin{align}
    \left\{\frac{1}{n} \sum\limits_{i=1}^{n} (T_i w_i-1) m_1(X_i)\right\}^2 - \lambda_1 \|m_1\|^2_{\mathcal{H}} &= \left\{\frac{1}{n} \sum\limits_{i=1}^{n} (T_i w_i-1) \langle m_1, K(\cdot,X_i) \rangle_{\mathcal{H}} \right\}^2 - \lambda_1 \|m_1\|^2_{\mathcal{H}} \nonumber\\
        &= \left\langle m_1, \frac{1}{n} \sum\limits_{i=1}^{n} (T_i w_i-1)K(\cdot,X_i) \right\rangle^2_{\mathcal{H}} - \lambda_1 \|m_1\|^2_{\mathcal{H}} \nonumber\\
        &\leq \|m_1\|^2_{\mathcal{H}} \cdot \left\| \frac{1}{n} \sum\limits_{i=1}^{n} (T_i w_i-1)K(\cdot,X_i) \right\|^2_{\mathcal{H}} - \lambda_1 \|m_1\|^2_{\mathcal{H}} \nonumber \\
        &= \|m_1\|^2_{\mathcal{H}} \left\| \frac{1}{n} \sum_{i\in S_1} w_iK(\cdot,X_i) - \frac{1}{n} \sum_{i\in [n]} K(\cdot,X_i) \right\|^2_{\mathcal{H}} - \lambda_1 \|m_1\|^2_{\mathcal{H}} \nonumber\\
        &= \|m_1\|^2_{\mathcal{H}} \cdot n^{-2} \left\| \sum_{i\in S_1} w_iK(\cdot,X_i) - n \cdot \frac{1}{n} \sum_{i\in [n]} K(\cdot,X_i) \right\|^2_{\mathcal{H}} - \lambda_1 \|m_1\|^2_{\mathcal{H}} \nonumber\\
        &= \|m_1\|^2_{\mathcal{H}} \cdot n^{-2} \left\| Mw - n h \right\|^2_{\mathcal{H}} - \lambda_1 \|m_1\|^2_{\mathcal{H}} \nonumber\\
        &= \|m_1\|^2_{\mathcal{H}} \cdot n^{-2} \{\left\| Mw - n h \right\|^2_{\mathcal{H}} - n^2 \lambda_1\} \nonumber%\\ 
        %&\leq n^{-2} \{\left\| Mw - n h \right\|^2_{\mathcal{H}} - \lambda_{1n}\}, \nonumber
\end{align}
where $\lambda_{1n} = n^2 \lambda_1$. The inequality follows from Cauchy-Schwarz inequality. Define $s(x)$ and $r(f)$ as follows:
$$ s(x) = \Lambda(x) + \frac{\lambda_2}{2} \sum_{i \in S_1} x^2_i,$$
where $\Lambda(x) = 0 $ when $x \succeq 1$ and $\Lambda(x) = \infty$, otherwise,
and 
$$ r(f) = \frac{1}{2}\| f - nh \|^2_{\mathcal{H}} - \frac{\lambda_{1n}}{2}. $$
Then minimization of \eqref{eq:objective} with respect to $w$ is equivalent to minimizing the primal $p:\mathbb{R}^n \mapsto \mathbb{R} \cup \{ \infty \}$ defined as
$$ p(w) = s(w) + r(Mw) = \Lambda(w) + \frac{\lambda_2}{2} \sum_{i \in S_1} w^2_i + \frac{1}{2} \left\{\sum_{i \in S_1} w_i K(\cdot, X_i)-\sum_{i \in [n]} K(\cdot, X_i)\right\}^2 - \frac{\lambda_{1n}}{2}. $$
We can obtain the Fenchel-Rockafeller dual, $d(f)$, \parencite{peypouquet_convex_2015} of the primal $p(w)$:
$$ d(f) = s^*(-M^* f) + r^*(f), $$
where $s^*$ and $r^*$ are Fenchel conjugates of $s$ and $r$, respectively. Furthermore, $M^*:\mathcal{H} \mapsto \mathbb{R}^n$ is the adjoint operator of $M$.

    \subsubsection{Obtaining Fenchel conjugate for $s(x)$}
By definition of Fenchel conjugate, 
\begin{align}
    s^*(y) &= \sup_x \{ y^\top x - s(x) \} = \sup_x \{ y^\top x - \Lambda(x) - \frac{\lambda_2}{2} x^\top x \} \nonumber\\
           &= \sup_{x: x \succeq 1} \{ y^\top x - \frac{\lambda_2}{2} x^\top x \}. \nonumber
\end{align}
The optimization problem can be written as
$$ Q(x) = y^\top x - \frac{\lambda_2}{2} x^\top x - \gamma^\top  (1-x). $$
The vanishing gradient of Lagrangian of the Karush-Kuhn-Tucker (KKT) conditions \parencite{boyd_convex_2004} requires
$$ \nabla_xQ(x) = y - \lambda_2 x^* + \gamma = 0$$
for $\gamma \succeq 0$, where $x^*$ is the optimum. By solving for this equation, 
$$ x^* = \frac{1}{\lambda_2}(y + \gamma). $$
From this,
$$s^*(y) = \frac{1}{2\lambda_2} y^\top y - \frac{1}{2\lambda_2} \gamma^\top \gamma.$$
For each $i \in S_1$, $x^*_i = \lambda^{-1}_2 (y_i + \gamma_i)$ implies $\gamma_i = \lambda_2 x^*_i - y_i $. We additionally need $\gamma_i (1 - x_i) = 0 $ for complementary slackness of the KKT condition \parencite{boyd_convex_2004}. Then
\begin{equation}
  x^*_i =
    \begin{cases}
      1 & \text{and $\gamma_i =\lambda_2 - y_i >0$}\\
      \frac{1}{\lambda_2}y_i; \frac{1}{\lambda_2}y_i>1 & \text{and $\gamma_i = 0$.}
    \end{cases}       
\end{equation}
Note that $\gamma_i = \mathbbm{1}(y_i < \lambda_2) (\lambda_2 - y_i) = \max(\lambda_2 -y_i, 0)$. By definition of adjoint operator, we have $\langle M^* f, e_i \rangle_{\mathcal{H}} = \langle f, Me_i \rangle_{\mathcal{H}} = f(X_i)$, where $e_i;~i\in[n]$ is a standard basis vector. Eventually, we have
$$ s^*(-M^*f) = \frac{1}{2 \lambda_2} \sum_{i \in S_1} f^2(X_i) - \frac{1}{2\lambda_2}\sum_{i \in S_1} \max\{\lambda_2 + f(X_i), 0\}. $$

%&= \frac{1}{2 \lambda_2} \sum_{i \in S_1} f^2(X_i) - \frac{1}{2\lambda_2}\sum_{i \in S_1}\gamma^2_i\{-f(X_i)\} \nonumber\\ 
%        &= \frac{1}{2 \lambda_2} \sum_{i \in S_1} f^2(X_i) - \frac{1}{2\lambda_2}\sum_{i \in S_1} \mathbbm{1}\{-f(X_i) < \lambda_2\} \{\lambda_2 + f(X_i)\} \nonumber\\
%        &

    \subsubsection{Obtaining Fenchel conjugate for $r(f)$}
The Fenchel conjugate of $r(f)$ can be derived similarly:
$$ r^*(f) = \sup_{g \in \mathcal{H}} \left\{ \langle f,g \rangle_{\mathcal{H}} - \frac{1}{2}\|g - nh\|^2_{\mathcal{H}}  + \frac{\lambda_{1n}}{2} \right\}. $$
By using functional differentiation, the optimum $g$ satisfies $f - g^* + nh = 0$, hence $g^* = f + nh$. By plugging this in,
$$ r^*(f) = \frac{1}{2}\| f \|^2_{\mathcal{H}} + \sum_{i \in [n]} f(X_i) + \frac{\lambda_{1n}}{2}. $$
Combining $s^*(-M^*f)$ and $r^*(f)$, we obtain the following Fenchel-Rockafeller dual as
$$ d(f) = \frac{1}{2 \lambda_2} \sum_{i \in S_1} f^2(X_i) + \frac{1}{2}\| f \|^2_{\mathcal{H}} + \sum_{i \in [n]} f(X_i) - \frac{1}{2\lambda_2} \sum_{i \in S_1} \max\left[\{\lambda_2 + f(X_i)\}^2, 0 \right] + \frac{\lambda_{1n}}{2}. $$

    \subsubsection{Obtaining $\hat{w}_i$}
Let $\hat{f} = \text{argmin}_{f \in \mathcal{H}} d(f)$. By Theorem 3.51 of \textcite{peypouquet_convex_2015}, $d(\hat{f}) = -p(\hat{w})$, and $-M^*\hat{f} \in \partial s(\hat{w}) = \{ \lambda_2 \hat{w} - \gamma; \gamma \succeq 0 \}$. Therefore, for $i \in S_1$, 
$$ -\hat{f}(X_i) = \lambda_2 \hat{w}_i - \max\{\lambda_2 + \hat{f}(X_i), 0\},$$
By rearranging,
\begin{equation}
    \hat{w}_i = \frac{-\hat{f}(X_i) + \max\{\lambda_2 + \hat{f}(X_i), 0\}}{\lambda_2} = \max\left\{1,-\hat{f}(X_i)/\lambda_2 \right\}.
    \label{eq:what}
\end{equation}

    \subsection{Step 2: Limiting form of the dual problem}
By dividing $d(f)$ by $n$, we get
$$ \frac{1}{n} d(f) = \frac{1}{2\lambda_2}\frac{1}{n} \sum_{i \in S_1} f^2(X_i) + \frac{1}{2n} \|f\|^2_{\mathcal{H}} + \frac{1}{n} \sum_{i \in [n]} f(X_i) - \frac{1}{2\lambda_2} \frac{1}{n} \sum_{i \in S_1}\max\{\lambda_2 + f(X_i), 0\} + \frac{\lambda_{1n}}{2n}. $$
As $n \rightarrow \infty$, this converges to
$$ J(f) = \frac{1}{2\lambda_2}E\left[\pi(+1,X) f^2(X)\right] + \mathbb{E}[f(X)] -\frac{1}{2 \lambda_2}E\left[\pi(+1,X) \max\{\lambda_2 + f(X), 0\}^2\right] + c_2, $$
where $c_2 \in \mathbb{R}$ is a constant from that $\lambda_{1n}/n = n \lambda_1 \asymp 1$. When $\lambda_2 + f(x) \leq 0$, the derivative of $\max\{\lambda_2 + f(x), 0\}^2$ with respect to $f$ is 0. Otherwise, the derivative of $\max\{\lambda_2 + f(x), 0\}^2$ with respect to $f$ is $2\max\{\lambda_2 + f(x), 0\}$. Thus, $J(f)$ is minimized with $f^*$ such that
$$ \frac{1}{\lambda_2} \pi(+1,x) f^*(x) + 1 -\frac{1}{\lambda_2} \pi(+1,x)\max\{\lambda_2 + f^*(x), 0\} = 0.$$
Then, 
$$ f^*(x) = -\lambda_2 \frac{1}{\pi(+1, X)} + \max\{\lambda_2 + f^*(x), 0\}. $$
By plugging this in equation (\ref{eq:what}), we obtain
\begin{align}
   {w}^*_i &= \frac{-f^*(X_i)+\max\{\lambda_2 + f^*(x), 0\}}{\lambda_2} \nonumber\\
        &= \frac{1}{\lambda_2} \left\{ -\left(-\lambda_2 \frac{1}{\pi(+1,X_i)} + \max\{\lambda_2 + f^*(x), 0\} \right) + \max\{\lambda_2 + f^*(x), 0\} \right\} \nonumber\\
        &= \frac{1}{\pi(+1,X_i)}. \nonumber
\end{align}

    \subsection{Step 3: Identification of the convergence in probability}
We show the convergence $n^{-1} \sum_{i \in S_1} (\hat{w}^{un}_i - w^*_i)^2 \rightarrow 0$ as $n \rightarrow \infty$, where $\hat{w}^{un}$ is obtained from the optimization without the constraint $w \succeq 1$. \textcite{hirshberg_augmented_2017} shows a general case of convergence of the unconstrained problem. Since $\hat{w_i} = \max{(\hat{w}^{un}_i, 1)}$ from \eqref{eq:what}, we have the following inequality:
\begin{align}
    \frac{1}{n} \sum_{i \in S_1} (\hat{w}^{un}_i - w^*_i)^2 &= \frac{1}{n} \sum_{i \in S_1: w^*_i > 1} (\hat{w}^{un}_i - w^*_i)^2 + \frac{1}{n} \sum_{i \in S_1: w^*_i = 1} (\hat{w}^{un}_i - w^*_i)^2 \nonumber\\
        &=\frac{1}{n} \sum_{i \in S_1: w^*_i > 1;\hat{w}^{un} \geq 1} (\hat{w}^{un}_i - w^*_i)^2 + \frac{1}{n} \sum_{i \in S_1: w^*_i > 1; \hat{w}^{un} < 1} (\hat{w}^{un}_i - w^*_i)^2 \nonumber\\
        &~~ + \frac{1}{n} \sum_{i \in S_1: w^*_i = 1; w^{un}_i>1} (\hat{w}^{un}_i - w^*_i)^2 + \frac{1}{n} \sum_{i \in S_1: w^*_i = 1; w^{un}_i<1} (\hat{w}^{un}_i - w^*_i)^2 \nonumber\\
        &\geq \frac{1}{n} \sum_{i \in S_1: w^*_i > 1;\hat{w}^{un} \geq 1} (\hat{w}^{un}_i - w^*_i)^2 + \frac{1}{n} \sum_{i \in S_1: w^*_i > 1; \hat{w}^{un} < 1} (\hat{w}_i - w^*_i)^2 \nonumber\\
        &~~ + \frac{1}{n} \sum_{i \in S_1: w^*_i = 1; w^{un}_i>1} (\hat{w}^{un}_i - w^*_i)^2 + \frac{1}{n} \sum_{i \in S_1: w^*_i = 1; w^{un}_i<1} (\hat{w}_i - w^*_i)^2 \nonumber\\
        &= \frac{1}{n} \sum_{i \in S_1: } (\hat{w}_i - w^*_i)^2, \nonumber
\end{align}
which follows from
$$ \sum_{i \in S_1: w^*_i > 1; \hat{w}^{un} < 1} (\hat{w}^{un}_i - w^*_i)^2 \geq \sum_{i \in S_1: w^*_i > 1; \hat{w}^{un} < 1} (\hat{w}_i - w^*_i)^2 = \sum_{i \in S_1: w^*_i > 1; \hat{w}^{un} < 1} (1 - w^*_i)^2$$
and 
$$ \sum_{i \in S_1: w^*_i = 1; w^{un}_i<1} (\hat{w}^{un}_i - w^*_i)^2 \geq  \sum_{i \in S_1: w^*_i = 1; w^{un}_i<1} (\hat{w}_i - w^*_i)^2 = 0. $$
Without the Lagrangian constraint, we have $\hat{w}^{un}_i = -\lambda^{-1}_2 \hat{f}(X_i)$ and $w^*_i = -\lambda^{-1}_2 f^*(X_i) = 1 / \pi(+1, X_i)$. Note that $w^*_i = 1 / \pi(+1, X_i)$ with and without the constraint in the optimization of the limiting term of $d(f)$. Define the penalized least squares problem
$$ \frac{1}{n} \sum_{i \in S_1} (w_i - w^*_i)^2 + \frac{\lambda_2}{n} \|w\|^2_2. $$
This can be rewritten as follows from the dual expression:
$$ \frac{1}{n} \sum_{i \in S_1} \left\{\frac{-f(X_i)}{\lambda_2} - w^*_i \right\}^2 + \frac{\lambda_2}{n} \left\|\frac{-f}{\lambda_2} \right\|^2_{\mathcal{H}} = \frac{1}{n} \sum_{i \in S_1} \left\{\frac{-f(X_i)}{\lambda_2} - w^*_i \right\}^2 + \frac{1}{n \cdot \lambda_2} \|f\|^2_{\mathcal{H}} $$
whose minimization is equivalent to that of
$$ \tilde{d}(f) = \frac{\lambda_2}{2} \sum_{i \in S_1} \left\{ \frac{-f(X_i)}{\lambda_2} - w^*_i \right\}^2 + \frac{1}{2} \| f \|^2_{\mathcal{H}}. $$
Let $\tilde{f} = \text{argmin}_{f \in \mathcal{H}} \tilde{d}(f)$ and $\delta = f - \tilde{f}$. Then
\begin{align}
    0   &= \lim_{t \rightarrow 0} \frac{\tilde{d}(\tilde{f} + t\delta) - \tilde{d}(\tilde{f})}{t} \nonumber\\ 
        &= -\sum_{i \in S_1} \frac{-\tilde{f}(X_i)}{\lambda_2}  \delta(X_i) + \sum_{i \in S_1} w^*_i \delta(X_i) + \langle \tilde{f}, \delta \rangle_{\mathcal{H}} \nonumber\\
        &= -\sum_{i \in S_1} (\tilde{w}_i - w^*_i) \delta(X_i) + \langle \tilde{f}, \delta \rangle_{\mathcal{H}}. \nonumber
\end{align}
Let the Fenchel-Rockafellar dual without the Lagrangian constraint be 
$$ d_{un}(f) = \frac{1}{2 \lambda_2} \sum_{i \in S_1} f^2(X_i) + \frac{1}{2}\| f \|^2_{\mathcal{H}} + \sum_{i \in [n]} f(X_i) + \frac{\lambda_{1n}}{2}. $$
From this, define $R(\delta)$ as follows:
\begin{align}
    R(\delta) &= \frac{1}{n} d_{un}(\tilde{f} + \delta) - \frac{1}{n} d_{un}(\tilde{f}) \nonumber\\
        &= \frac{1}{n} \frac{1}{2\lambda_2} \sum_{i \in S_1} \delta^2(X_i) - \frac{1}{n} \sum_{i \in S_1} \frac{-\tilde{f}(X_i)}{\lambda_2} \delta(X_i) + \frac{1}{n} \sum_{i \in [n]} \delta(X_i) + \frac{1}{2n} \|f\|^2_{\mathcal{H}} - \frac{1}{2n} \|\tilde{f}\|^2_{\mathcal{H}} \nonumber\\
        &= \frac{1}{n} \frac{1}{2\lambda_2} \sum_{i \in S_1} \delta^2(X_i) - \frac{1}{n} \sum_{i \in S_1} \tilde{w}_i \delta(X_i) + \frac{1}{n} \sum_{i \in [n]} \delta(X_i) + \frac{1}{2n} \|f\|^2_{\mathcal{H}} - \frac{1}{2n} \|\tilde{f}\|^2_{\mathcal{H}} \nonumber\\
        &= \frac{1}{n} \frac{1}{2\lambda_2} \sum_{i \in S_1} \delta^2(X_i) - \frac{1}{n} \sum_{i \in S_1} \tilde{w}_i \delta(X_i) + \frac{1}{n} \sum_{i \in [n]} \delta(X_i) + \frac{1}{2n} \|f\|^2_{\mathcal{H}} - \frac{1}{2n} \|\tilde{f}\|^2_{\mathcal{H}} \nonumber\\
        &~~ + \frac{1}{n} \sum_{i \in S_1} (\tilde{w}_i - w^*_i) \delta(X_i) - \frac{1}{n} \langle \tilde{f}, \delta \rangle_{\mathcal{H}} \\
        &= \frac{1}{2\lambda_2 n} \sum_{i \in S_1} \delta^2(X_i) - \frac{1}{n} \sum_{i \in S_1} w^*_i \delta(X_i) + \frac{1}{n} \sum_{i \in [n]} \delta(X_i) + \frac{1}{2n} \|f - \tilde{f}\|^2_{\mathcal{H}}  \nonumber\\
        &= G(\delta) - H(\delta) + \frac{1}{2n} \|f - \tilde{f}\|^2_{\mathcal{H}},  \nonumber
\end{align}
where
$$ G(\delta) = \frac{1}{2\lambda_2 n} \sum_{i \in S_1} \delta^2(X_i) ~~\text{and}  $$
$$ H(\delta) =  \frac{1}{n} \sum_{i \in S_1} w^*_i \delta(X_i) - \frac{1}{n} \sum_{i \in [n]} \delta(X_i),$$
and the lines from (S4) are derived by subtracting $n^{-1}\left\{-\sum_{i \in S_1} (\tilde{w}_i - w^*_i) \delta(X_i) + \langle \tilde{f}, \delta \rangle_{\mathcal{H}}\right\}$. From Section A.2 of \textcite{hirshberg_augmented_2017}, we can show that $\hat{w} \approx \tilde{w}$ as long as $R(\delta) \leq 0$.

By Lemma S1 and Lemma S2, there exists $d_n = o(n^{-1/4})$ such that for all $\epsilon > 0$ and $u \in \mathcal{H}_1$,
\begin{align}
    \mathbb{E}[u^2(X)] &\geq d^2_n && \text{implies } n^{-1}_1 \sum_{i \in S_1} u^2(X_i) \geq 2^{-1} \mathbb{E}[u^2(X)]\\
    \mathbb{E}[u^2(X)] &\leq d^2_n && \text{implies } |H(u)| \leq d^2_n
\end{align}
with probability at least $1-\epsilon$. Define $b_n = \max\{ 4 \lambda_2 \cdot n / n_1, 2 n \cdot d^2_n \}$, where $n_1 = \sum_{i\in [n]} \mathbbm{1}(A_i = +1)$. Consider the case when $\delta/b_n \in \mathcal{H}_1$ from which $\| \delta/b_n \|_{\mathcal{H}} \leq 1$, hence $\|\delta \|_{\mathcal{H}} \leq b_n$. Suppose $\mathbb{E}[\delta^2(X)] \leq 
b^2_n d^2_n$, so that $\mathbb{E}[\delta^2(X) / b^2_n] \leq d^2_n$. Then by Lemma S2, $|H(\delta/b_n)| \leq d^2_n$, and 
$$|H(\delta)| = b_n \cdot |H(\delta/b_n)| \leq b_n d^2_n.$$
This leads to the following inequality:
$$ R(\delta) \geq G(\delta) - |H(\delta)| \geq G(\delta) - b_n d^2_n. $$
Therefore, $R(\delta) \leq 0$ is only possible when $G(\delta) \leq b_n d^2_n$.

On the other hand, suppose $\mathbb{E}[\delta^2(X)] \geq b^2_n d^2_n$. Let $n_1 = |S_1|$. By Lemma S1,
$$ \frac{1}{n_1} \sum_{i \in S_1} \delta^2(X_i) \geq \frac{1}{2} \mathbb{E}[\delta^2(X)] = \frac{1}{2}\beta^2, $$
with probability at least $1-\epsilon$, where $\beta = \sqrt{\mathbb{E}[\delta^2(X)]}$. Note that $\beta \geq b_n d_n$. By rearranging, we obtain
$$ \frac{1}{2\lambda_2 n} \sum_{i \in S_1} \delta^2(X_i) \geq \frac{n_1}{4\lambda_2 n } \beta^2.$$
Define $\delta_1 = d_n \cdot \delta / \beta$. Then
$$ \| \delta_1 \|_{\mathcal{H}} = \left\|\frac{ d_n \cdot \delta}{\beta} \right\|_{\mathcal{H}} = \frac{\| \delta \|_{\mathcal{H}} d_n}{ \beta} \leq \frac{b_n \cdot d_n}{ \beta} \leq 1. $$
Also, we have
$$ \mathbb{E}[\delta^2_1(X)] = \mathbb{E}\left[ \frac{d^2_n \cdot \delta^2(X)}{\beta^2} \right] =  \frac{d^2_n \cdot \mathbb{E}[\delta^2(X)]}{\beta^2} \leq d^2_n. $$
By Lemma S2, $|H(\delta_1)| \leq d^2_n$. Thus,
$$ |H(\delta)| = \left|H \left( \frac{\beta \delta_1}{d_n} \right)\right| = \frac{\beta |H(\delta_1)| }{d_n} \leq \frac{\beta}{d_n} d^2_n = \beta d_n = \frac{\beta^2 d_n}{\beta} \leq \frac{\beta^2}{b_n}. $$
where the last inequality comes from $1 / b_n \geq d_n / \beta$.
Note that
$$ G(\delta) = \frac{n_1}{2\lambda_2 n} \cdot \frac{1}{n_1}\sum_{i \in S_1} \delta^2(X_i) \geq \frac{n_1}{2\lambda_2 n} \cdot \frac{1}{2} \mathbb{E}[\delta^2(X)] = \frac{n_1}{2\lambda_2 n} \cdot \frac{1}{2} \beta^2 = \frac{n_1}{4\lambda_2 n} \beta^2, $$
hence
$$ R(\delta) \geq G(\delta) - |H(\delta)| \geq \frac{n_1}{4\lambda_2 n} \beta^2 - \frac{1}{b_n} \beta^2 \geq 0 $$
from that $b_n \geq 4 \lambda_2 n / n_1$.

Now let $\| \delta \|_{\mathcal{H}} > b_n$. Define $k = \|\delta\|_{\mathcal{H}}/(b_n)> 1$. Let $\delta_2 = \delta / k$. Then
$$ \| \delta_2 \|_{\mathcal{H}} = \left\| \frac{\delta}{k} \right\|_{\mathcal{H}} = \frac{1}{k} \| \delta \|_{\mathcal{H}} \leq \frac{b_n}{\| \delta \|_{\mathcal{H}}} \| \delta \|_{\mathcal{H}} \leq b_n.$$
Note that $\delta = k \delta_2$. Then
\begin{align}
    R(\delta) - k R(\delta_2) &= R(k\delta_2) - kR(\delta_2) \nonumber\\
        &= k(k-1) \frac{1}{2\lambda_2 n} \sum_{i \in S_1} \delta^2_2(X_i) - (k-k)H(\delta_2) + k(k-1) \frac{1}{2n} \|\delta_2\|^2_{\mathcal{H}} \nonumber\\
        &\geq 0 \nonumber.
\end{align}
This shows that $R(\delta) \geq k R(\delta_2) \geq R(\delta_2)$. Therefore, $R(\delta_2) \geq 0$ implies $R(\delta) \geq 0$. Note that $\| \delta_2 \|_{\mathcal{H}} \leq b_n$. When $\mathbb{E}[\delta^2_2(X)] > b^2_n d^2_n$, $R(\delta_2) \geq G(\delta_2) - |H(\delta_2)| \geq 0$ from the similar derivation as above. Suppose $\mathbb{E}[\delta^2_2(X)] \leq b^2_n d^2_n$. Then, also from the derivation above, $|H(\delta_2)| \leq b_n d^2_n$, and
$$ R(\delta_2) \geq G(\delta_2) - |H(\delta_2)| + \frac{1}{2n} \|\delta_2\|^2_{\mathcal{H}}.$$
From this, we have the following:
$$ R(\delta_2) \geq - |H(\delta_2)| + \frac{1}{2n} \|\delta_2\|^2_{\mathcal{H}} \geq -b_n d^2_n  + \frac{1}{2n }b^2_n. $$
The lower bound is non-negative since $b_n \geq 2n d^2_n$.

Therefore, $R(\delta) \leq 0$ with probability at least $1-\epsilon$ only when $G(\delta) \leq b_n d^2_n$. Since $\hat{f} = \tilde{f} + \hat{\delta}$ minimizes $d(f)$, this inequality holds when $\hat{\delta} = \hat{f} - \tilde{f}$, from which $G(\hat{\delta}) \leq b_n d^2_n$. Then
\begin{align}
    G(\hat{\delta})  &=\frac{1}{2 \lambda_2 n} \sum_{i \in S_1} \hat{\delta}^2(X_i) = \frac{\lambda_2}{2n} \sum_{i \in S_1} \left(\frac{\hat{f}(X_i) - \tilde{f}(X_i)}{\lambda_2} \right)^2 \nonumber\\
        &= \frac{\lambda_2}{2n} \sum_{i \in S_1} \left(\hat{w}^{un}_i - \tilde{w}_i \right)^2 \leq b_n d^2_n. \nonumber
\end{align}
Since $d_n = o(n^{-1/4})$ and $b_n = O(\sqrt{n})$, $b_n d^2_n \rightarrow 0$ as $n \rightarrow \infty$. When $n$ is sufficiently large, there exists a small $\eta > 0$ such that $\eta < b_n d^2_n$. From this,
$$ \text{Pr}\left( \frac{\lambda_2}{2n} \sum_{i \in S_1} \left(\hat{w}^{un}_i - \tilde{w}_i \right)^2 \leq \eta \right) \geq \text{Pr}\left( \frac{\lambda_2}{2n} \sum_{i \in S_1} \left(\hat{w}^{un}_i - \tilde{w}_i \right)^2 \leq b_n d^2_n \right) \geq 1 - \epsilon. $$
Therefore, $\frac{1}{n}\sum_{i \in S_1} (\hat{w}^{un}_i - \tilde{w}_i)^2 \rightarrow 0$ in probability.

From the regularity condition that $\mathcal{H}$ is universal with respect to the supremum norm, we can choose a sequence $\{u_j; j \in \mathbb{N}\} \subset \mathcal{H}$ such that $\|u_j - f^*\|_{\infty} \rightarrow 0$ as $j \rightarrow \infty$, where $\mathbb{N}$ denotes the set of natural numbers. Moreover, we can choose the subsequence $\{u_{jn}\}$ such that $\|u_{jn}\|_{\mathcal{H}}/\sqrt{n} = o(1)$ as $n \rightarrow \infty$. We can find the corresponding weights $v_{jn,i} = -u_{jn}(X_i) / \lambda_2$ from which
\begin{align}
    \max_{i \in S_1}|v_{jn,i} - w^*_i| &= \max_{i \in S_1}\left| \frac{-v_{jn}(X_i)}{\lambda_2} - \frac{-f^*(X_i)}{\lambda_2} \right| \nonumber\\
        &= \lambda^{-1}_2 \max_{i \in S_1}| -v_{jn}(X_i) + f^*(X_i)| \rightarrow 0. \nonumber
\end{align}
Since $\tilde{f}$ is the minimizer of $\tilde{d}(f)$, we have
\begin{align}
    \frac{1}{n}\sum_{i \in S_1} (\tilde{w}_i - w^*_i)^2 &\leq \frac{2}{\lambda_2 n} \tilde{d}(\tilde{f}) \nonumber\\
        &= \frac{1}{n} \sum_{i \in S_1} (\tilde{w}_i - w^*_i)^2  + \frac{1}{\lambda_2 n} \|\tilde{f}\|^2_{\mathcal{H}} \nonumber\\
        &\leq \frac{2}{\lambda_2n} \tilde{d}(u_{jn}) \leq \max_{i \in S_1}(v_{jn,i} - w^*_i)^2  + \frac{1}{\lambda_2} \frac{\|u_{jn}\|^2_{\mathcal{H}}}{\sqrt{n}}. \nonumber
\end{align}
The last line converges to 0 as $n \rightarrow \infty$, so $\frac{1}{n}\sum_{i \in S_1} (\tilde{w}_i - w^*_i)^2 \rightarrow 0$ in probability. By triangular inequality,
$$ \sqrt{\frac{1}{n}\sum_{i \in S_1} (\hat{w}^{un}_i - w^*_i)^2} \leq \sqrt{\frac{1}{n}\sum_{i \in S_1} (\hat{w}^{un}_i - \tilde{w}_i)^2} + \sqrt{\frac{1}{n}\sum_{i \in S_1} (\tilde{w}_i - w^*_i)^2}. $$
Therefore, $n^{-1} \sum_{i \in S_1} (\hat{w}^{un}_i - {w}^*_i)^2 \rightarrow 0$ in probability. This implies $n^{-1} \sum_{i \in S_1} (\hat{w}_i - {w}^*_i)^2 = o_p(1)$.

\begin{customlemma}{S1}
    Suppose $u \in \mathcal{H}_1$ has the second moment $\mathbb{P} u^2 \geq d^2$, where $d \geq d_n = o(n^{-1/4})$. Then
    $$ \text{Pr}\left( \mathbb{P}_n u^2 \geq \frac{1}{2} \mathbb{P} u^2 \right) \geq 1 - 2 \exp (-d_1 n d^2_n / M^2_{\infty}(\mathcal{H}_1) ), $$
    where $M_{\infty}(\mathcal{H}_1) = \sup_{u \in \mathcal{H}_1} \sup_{x \in \mathcal{X}} |f(x)|$, and $d_1$ is a constant that depends on the multiplier 1/2.
\end{customlemma}
\begin{proof}
    The proof follows from Lemma 4 and Section A.5.1 of \textcite{hirshberg_augmented_2017} that uses Corollary 3.3 of \textcite{mendelson_extending_2017} and introduces the uniform lower bound on the ratio of the empirical and the true second moments that gives
    $$ \text{Pr}\left( \mathbb{P}_n u^2 / P u^2 \geq \xi \right) \geq 1 - 2 \exp \{-d_1 n d^2_n / M^2_{\infty}(\mathcal{H}_1)\}, $$
    where $\xi < 1$. We chose $\xi = 1/2$.
\end{proof}

\begin{customlemma}{S2}
    Define an operator $H:\mathcal{H} \mapsto \mathbb{R}$ as
    $$ H(u) = \frac{1}{n}\sum_{i \in S_1}w^*_i u(X_i) - \frac{1}{n} \sum_{i \in [n]} u(X_i). $$
    Suppose $u \in \mathcal{H}_1$ has the second moment $\mathbb{P}u^2 \leq d^2_n$, where $d_n = o(n^{-1/4})$. Then for all $\epsilon > 0$,
    $$ \text{Pr}( |H(u)| \leq d^2_n ) \geq 1- \epsilon. $$
\end{customlemma}
\begin{proof}
    Note that $\mathbb{E}[\mathbbm{1}(A=+1) w^*(X) u(X)] = E[u(X)]$. Therefore,
    \begin{align}
        H(u) &= \frac{1}{n}\sum_{i \in S_1}w^*_i u(X_i) - \frac{1}{n} \sum_{i \in [n]} u(X_i) \nonumber\\
        &= \frac{1}{n}\sum_{i \in S_1}w^*_i u(X_i) - \frac{1}{n} \sum_{i \in [n]} u(X_i) - \mathbb{E}[\mathbbm{1}(A=+1) w^*(X) u(X)]  + \mathbb{E}[u(X)] \nonumber\\
        &= (\mathbb{P}_n - \mathbb{P})\{\mathbbm{1}(A=+1) w^* u\} - (\mathbb{P}_n - \mathbb{P})(u). \nonumber
    \end{align}
    Since $\pi(+1, x) > 0$, $\mathbbm{1}(A=+1)w^*(x)$ is bounded above. Hence the function class $\{ t \cdot w^*(x); t \in \{0,1\} \}$ is uniformly bounded and $P$-Donsker. Also, from the regularity condition, for all $u \in \mathcal{H}_1$,
    $$ |u(x)| = |\langle u, K(\cdot, x) \rangle_{\mathcal{H}}| \leq \|u\|_{\mathcal{H}} \sqrt{K(x,x)} \leq \sqrt{K(x,x)} < \infty. $$
    Therefore, from Corollary 9.32 (v) of \textcite{kosorok_introduction_2008}, $\{\mathbbm{1}(A=+1) w^* u: u \in \mathcal{H}_1\}$ is a $\mathbb{P}$-Donsker class. From Section A.5.1 of \textcite{hirshberg_augmented_2017}, there exists $d_{1n} = o(n^{-1/4})$ such that
    $$ \text{Pr}\Big( (\mathbb{P}_n - \mathbb{P})\{\mathbbm{1}(A=+1) w^* u \leq d^2_{1n} / 2\} \Big) \geq 1 - \epsilon/2. $$
    Similarly, for $d_{2n} = o(n^{-1/4})$, $\text{Pr}\left\{ (\mathbb{P}_n - \mathbb{P})(u) \leq d^2_{2n}/2 \right\} \geq 1 - \epsilon/2$.
    Define $d_n = \max(d_{1n}, d_{2n})$. By combining these two inequalities, we have
    $$ \text{Pr}(|H(u)| \leq d^2_n) \geq 1 - \epsilon. $$
\end{proof}

\section{Convergence of $g_{\hat{w}}(x)$ to $g(x)$}

We first demonstrate that $\sup_{x \in \mathcal{X}}|g_n(x) - g(x)| = o_p(1)$ for compact $\mathcal{X}$. Then we use this result to show that $\sup_{x \in \mathcal{X}}|g_{\hat{W}}(x) - g(x)| = o_p(1)$

 \subsection{Convergence of $g_n(x)$ to $g(x)$}
We show that $g_n(x)$ uniformly converges to $g(x)$ in probability for compact $\mathcal{X}$. Let $\tilde{Y}=\left(\pi^{-1}(A, X) - 1\right) Y$. Then
\begin{align}
    |g_n(x) - g(x)| &= \left| \frac{1}{n} \sum_{i \in [n]} x^\top (n^{-1}\mathbb{X}^\top \mathbb{X})^{-1} X_i \tilde{Y}_i - \mathbb{E}[x^\top \{\mathbb{E}(X X^\top) \}^{-1} X\tilde{Y}] \right| \nonumber\\
        &= \left| x^\top (n^{-1}\mathbb{X}^\top \mathbb{X})^{-1} \left(n^{-1} \sum_{i \in [n]} X_i \tilde{Y}_i \right) - x^\top \{\mathbb{E}(XX^\top)\}^{-1} \mathbb{E}[X \tilde{Y}] \right| \nonumber
\end{align}
By the weak law of large numbers, 
$$ (n^{-1}\mathbb{X}^\top \mathbb{X})^{-1} = ( \mathbb{E}[X X^\top] + R_n )^{-1};~~ n^{-1} \sum_{i \in [n]} X_i \tilde{Y}_i = \mathbb{E}[X \tilde{Y}] + r_n, $$
where $R_n \to 0_{p \times p}$ and $r_n \to 0_p$ as $n \to \infty$ in probability. From this 
$$  (n^{-1}\mathbb{X}^\top \mathbb{X})^{-1} \left(n^{-1} \sum_{i \in [n]} X_i \tilde{Y}_i \right) \to_p \{\mathbb{E}(XX^\top)\}^{-1} \mathbb{E}[X \tilde{Y}]. $$ 
By the Cauchy-Schwarz inequality,
\begin{align}
    \sup_{x \in \mathcal{X}}|g_n(x) - g(x)| &= \sup_{x \in \mathcal{X}} \left| x^\top (n^{-1}\mathbb{X}^\top \mathbb{X})^{-1} \left(n^{-1} \sum_{i \in [n]} X_i \tilde{Y}_i \right) - x^\top \{\mathbb{E}(XX^\top)\}^{-1} \mathbb{E}[X \tilde{Y}] \right| \nonumber\\
        &= \sup_{x \in \mathcal{X}} \left| x^\top \left\{    (n^{-1}\mathbb{X}^\top \mathbb{X})^{-1} \left(n^{-1} \sum_{i \in [n]} X_i \tilde{Y}_i \right) -\{\mathbb{E}(XX^\top)\}^{-1} \mathbb{E}[X \tilde{Y}] \right\} \right| \nonumber\\
        &\leq \sup_{x \in \mathcal{X}} \| x \|_2 \cdot \left\| (n^{-1}\mathbb{X}^\top \mathbb{X})^{-1} \left(n^{-1} \sum_{i \in [n]} X_i \tilde{Y}_i \right) -\{\mathbb{E}(XX^\top)\}^{-1} \mathbb{E}[X \tilde{Y}] \right\|_2 \nonumber\\
        &= \sup_{x \in \mathcal{X}} \|x\| \cdot o_p(1) = o_p(1), \nonumber
\end{align}
since $\sup_{x \in \mathcal{X}} \|x\| < \infty$ follows from compactness of $\mathcal{X}$.

    \subsection{Proof of Corollary \ref{cor:gx_converge}}
For any $x \in \mathcal{X}$, by Cauchy-Schwarz inequality,
\begin{align}
    \sup_{x \in \mathcal{X}}|g_{\hat{w}}(x) - g_n(x)| &= \sup_{x \in \mathcal{X}}\left| \frac{1}{n}\sum_{i \in [n]} x^\top (n^{-1}\mathbb{X}^\top \mathbb{X})^{-1}X_i (\hat{w}_i - 1)Y_i - \frac{1}{n}\sum_{i \in [n]} x^\top (n^{-1}\mathbb{X}^\top \mathbb{X})^{-1}X_i \left\{\frac{1}{\pi(A_i,X_i)} - 1\right\}Y_i \right| \nonumber\\
    &= \sup_{x \in \mathcal{X}} \left| \frac{1}{n}\sum_{i \in [n]} x^\top (n^{-1}\mathbb{X}^\top \mathbb{X})^{-1}X_i \left\{\hat{w}_i - \frac{1}{\pi(A_i,X_i)} \right\} Y_i \right| \nonumber\\
    &\leq \frac{1}{n}\sum_{i \in [n]} \sup_{x \in \mathcal{X}}|x^\top (n^{-1}\mathbb{X}^\top \mathbb{X})^{-1}X_i| \cdot |Y_i| \cdot \left|\hat{w}_i - \frac{1}{\pi(A_i,X_i)} \right| \nonumber\\
    &\leq \frac{1}{n}\sum_{i \in [n]} \sup_{x \in \mathcal{X}} \| x \|_2 \left\| (n^{-1} \mathbb{X}^\top \mathbb{X})^{-1} \right\|_2 \cdot \|X_i\|_2 \cdot |Y_i| \cdot \left|\hat{w}_i - \frac{1}{\pi(A_i,X_i)} \right| \nonumber\\
    &= \sup_{x \in \mathcal{X}} \| x \|_2 \left\| (n^{-1}\mathbb{X}^\top \mathbb{X})^{-1} \right\|_2 \cdot \max_{i \in [n]}\|X_i\|_2 \cdot \sqrt{\frac{1}{n}\sum_{i \in [n]}Y^2_i} \sqrt{\frac{1}{n}\sum_{i \in [n]} \left\{\hat{w}_i - \frac{1}{\pi(A_i,X_i)}\right\}^2}. \nonumber
\end{align}
Since $\mathcal{X}$ is compact  $\sup_{x \in \mathcal{X}} \| x \|_2 < \infty$ and $\max_{i \in [n]}\|X_i\|_2 < \infty$. Also, by weak law of large numbers, $\left\| (n^{-1}\mathbb{X}^\top \mathbb{X})^{-1} \right\|_2 \to_p \left\|\{\mathbb{E}[XX^\top]\}^{-1}\right\|_2$. Thus
$$ \sup_{x \in \mathcal{X}} \| x \|_2 \cdot \left\| (n^{-1}\mathbb{X}^\top \mathbb{X})^{-1} \right\|_2 \cdot \max_{i \in [n]}\|X_i\|_2 = \sup_{x \in \mathcal{X}} \| x \|_2 \cdot \left\|\{\mathbb{E}[XX^\top]\}^{-1} \right\|_2 \cdot \max_{i \in [n]}\|X_i\|_2 + o_p(1). $$
Hence, we can express the left hand side by $O_p(1)$. From Theorem \ref{thm:WeightConv},
\begin{align}
   \sup_{x \in \mathcal{X}} |g_{\hat{w}}(x) - g_n(x)| &\leq O_p(1) \sqrt{ \frac{1}{n}\sum_{i \in [n]} Y^2_i} \sqrt{\frac{1}{n}\sum_{i \in [n]} \left\{\hat{w}_i - \frac{1}{\pi(A_i,X_i)}\right\}^2} \nonumber\\
    &= O_p(1) \mathbb{E}[Y^2] \cdot o_p(1) \nonumber\\
    &= o_p(1). \nonumber
\end{align}

Therefore, $g_{\hat{w}}(x) \to g(x)$ over $x \in \mathcal{X}$ in probability from that $\sup_{x \in \mathcal{X}}|g_{\hat{w}}(x) - g(x)| \leq \sup_{x \in \mathcal{X}}|g_{\hat{w}}(x) - g_n(x)| + \sup_{x \in \mathcal{X}}|g_n(x) - g(x)| = o_p(1)$ by triangular inequality.

\section{Proof of Theorem \ref{thm:CrossCov}}
Let $w^*_i = 1/\pi(A_i,X_i)$. Then
\begin{align}
    \frac{1}{n} \sum_{i \in [n]} \tilde{Z}_i f(X_i) &= \frac{1}{n} \sum_{i \in [n]} \hat{w}_i A_i \{Y_i - g_{\hat{w}}(X_i)\} f(X_i) \nonumber\\
        &= \frac{1}{n} \sum_{i \in [n]} (\hat{w}_i - w^*_i) A_i \{Y_i - g_{\hat{w}}(X_i)\} f(X_i) + \frac{1}{n} \sum_{i \in [n]} w^*_i A_i \{Y_i - g_{\hat{w}}(X_i)\} f(X_i) \nonumber\\
        &= \frac{1}{n} \sum_{i \in [n]} (\hat{w}_i - w^*_i) A_i \{Y_i - g(X_i)\} f(X_i) + \frac{1}{n} \sum_{i \in [n]} (\hat{w}_i - w^*_i) A_i \{g(X_i) - g_{\hat{w}}(X_i)\} f(X_i) \nonumber\\
        &~~ + \frac{1}{n} \sum_{i \in [n]} w^*_i A_i \{g(X_i) - g_{\hat{w}}(X_i)\} f(X_i) + \frac{1}{n} \sum_{i \in [n]} w^*_i A_i \{Y_i - g(X_i)\} f(X_i) \nonumber\\
        &= \frac{1}{n} \sum_{i \in [n]} (\hat{w}_i - w^*_i) A_i \{Y_i - g(X_i)\} f(X_i) + \frac{1}{n} \sum_{i \in [n]} (\hat{w}_i - w^*_i) A_i \{g(X_i) - g_{\hat{w}}(X_i)\} f(X_i) \nonumber\\
        &~~ + \frac{1}{n} \sum_{i \in [n]} w^*_i A_i \{g(X_i) - g_{\hat{w}}(X_i)\} f(X_i) + \mathbb{E}[Z f(X)] + o_p(1),\nonumber
\end{align}
where the last equality comes from the weak law of large numbers. We show that the first three terms are $o_p(1)$. By the Cauchy-Schwarz inequality, the first term is bounded by
\begin{align}
    \sqrt{\frac{1}{n} \sum_{i \in [n]} (\hat{w}_i - w^*_i)^2 } \sqrt{\frac{1}{n}\sum_{i \in [n]} \{Y_i-g(X_i)\}^2 f^2(X_i)} &= o_p(1) \sqrt{\frac{1}{n}\sum_{i \in [n]} \pi^2(A_i,X_i) w^{*2}_i\{Y_i-g(X_i)\}^2 f^2(X_i)} \nonumber\\
        &\leq o_p(1) \sqrt{\frac{1}{n}\sum_{i \in [n]} Z^2_i f^2(X_i)} \nonumber\\
        &= o_p(1) \sqrt{\mathbb{E}[Z^2f^2(X)] + o_p(1)} = o_p(1). \nonumber
\end{align}
The second term is bounded by
\begin{align}
   \sqrt{\frac{1}{n} \sum_{i \in [n]} (\hat{w}_i - w^*_i)^2 } \sqrt{\frac{1}{n}\sum_{i \in [n]}\{g(X_i) - g_{\hat{w}}(X_i)\}^2 f^2(X_i)} &\leq o_p(1) \sup_{x \in \mathcal{X}} |g(x) - g_{\hat{w}}(x)| \sqrt{\frac{1}{n} \sum_{i \in [n]} f^2(X_i)} \nonumber\\
    &= o_p(1) \sqrt{\mathbb{E}[f^2(X)] + o_p(1) } = o_p(1), \nonumber
\end{align}
since $\sup_{x \in \mathcal{X}} |g(x) - g_{\hat{w}}(x)| = o_p(1)$ (Corollary \ref{cor:gx_converge}). The third term is bounded by
\begin{align}
    \sqrt{ \frac{1}{n} \sum_{i \in [n]} w^{*2}_i \{g(X_i) - g_{\hat{w}}(X_i)\}^2 } \cdot \sqrt{\frac{1}{n} \sum_{i \in [n]} f^2(X_i)} &\leq \max_{i \in [n]} \{w^{*}_i\} \sup_{x \in [n]}|g(X_i) - g_{\hat{w}}(X_i)| \cdot \sqrt{\frac{1}{n} \sum_{i \in [n]} f^2(X_i)} \nonumber\\
        &=\max_{i \in [n]} \{w^{*}_i\} \cdot o_p(1) \sqrt{\mathbb{E}[f^2(X)] + o_p(1)} = o_p(1), \nonumber
\end{align}
which follows from $1/\pi(A,X) < \infty$ from \textbf{Assumption 2}. By combining these results, we have $n^{-1}\sum_{i \in [n]} \tilde{Z}_i f(X_i) = \mathbb{E}[Zf(X)] + o_p(1)$.

\section{Proof of Theorem \ref{thm:ConsistentValue}}
%By Theorem~\ref{thm:CrossCov} with the function $f(X)$ replaced by $\mathbbm{1}\{ d(X) = A \}$ and $\mathbbm{1}\{\mathrm{sign} \circ f(X) \neq A\}$, respectively, we have %The proof is omitted for
%$$ 
%\frac{1}{n} \sum_{i \in [n]} \hat{w}_i Y_i \cdot \mathbbm{1}\{ d(X_i) = A_i \} = \mathbb{E}[Y\{d(X)\}] + o_p(1) ~~ \text{and} 
%$$
%$$ 
%\frac{1}{n} \sum_{i \in [n]} \hat{w}_i Y_i \cdot \mathbbm{1}\{ \mathrm{sign} \circ f(X_i) \neq A_i \} = \mathcal{R}(f) + o_p(1).
%$$
Whenever $\mathrm{sign}\{ Y_i - g_{\hat{w}}(X_i) \} \neq \mathrm{sign}\{ Y_i - g(X_i) \}$, we have
$$ |Y_i - g(X_i)| \leq |Y_i - g_{\hat{w}}(X_i) - \{ Y_i - g(X_i) \}| = |g_{\hat{w}}(X_i) - g(X_i)|. $$
From Corollary \ref{cor:gx_converge}, $|g_{\hat{w}}(X_i) - g(X_i) | \leq \epsilon$ with high probability for large $n$. Also, $| \mathrm{sign}\{ Y_i - g_{\hat{w}}(X_i) \} - \mathrm{sign}\{ Y_i - g(X_i) \} | = 2$ whenever $\mathrm{sign}\{ Y_i - g_{\hat{w}}(X_i) \} \neq \mathrm{sign}\{ Y_i - g(X_i) \}$. Thus
\begin{align}
    \sum_{i \in [n]} | \mathrm{sign}\{ Y_i - g_{\hat{w}}(X_i) \} - \mathrm{sign}\{ Y_i - g(X_i) \} | &= 2 \sum_{i \in [n]} \mathbbm{1}\{ \mathrm{sign}( Y_{g_{\hat{w}},i} ) \neq \mathrm{sign}( Y_{g,i} )\} \nonumber\\
        &\leq 2 \sum_{i \in [n]} \mathbbm{1}\{ |Y_i - g(X_i)| \leq | g_{\hat{w}}(X_i) - g(X_i) | \} \nonumber\\
        &\leq 2 \sum_{i \in [n]} \mathbbm{1}\{ |Y_i - g(X_i)| \leq \epsilon \} \nonumber\\
        &\leq 2 C n^{\xi}. \nonumber
\end{align}
Likewise, $\sum_{i \in [n]} | \mathrm{sign}\{ Y_i - g_{\hat{w}}(X_i) \} - \mathrm{sign}\{ Y_i - g(X_i) \} |^2 \leq 4 C n^{\xi}.$ Note that
\begin{align}
   \frac{1}{n} \sum_{i \in [n]} \hat{w}_i |Y_{g_{\hat{w}}, i}| \phi\left\{ A_i \cdot \mathrm{sign}(Y_{g_{\hat{w}}, i}) f(X_i) \right\} &= \frac{1}{n} \sum_{i \in [n]} \left\{ \hat{w}_i - \frac{1}{\pi(A_i, X_i)} \right\} |Y_{g_{\hat{w}},i}| \phi\left\{ A_i \cdot \mathrm{sign}(Y_{g_{\hat{w}},i}) f(X_i) \right\} \nonumber\\
    &~~ + \frac{1}{n} \sum_{i \in [n]}  \frac{|Y_i - g_{\hat{w}}(X_i)|}{\pi(A_i, X_i)} \phi\left\{ A_i \cdot \mathrm{sign}(Y_{g_{\hat{w}},i}) f(X_i) \right\}. \nonumber
\end{align}
The first term can be re-expressed as
$$ \frac{1}{n} \sum_{i \in [n]} \left\{ \hat{w}_i - \frac{1}{\pi(A_i, X_i)} \right\} \pi(A_i,X_i) \frac{|Y_{g_{\hat{w}},i}|}{\pi(A_i,X_i)} \phi\left\{ A_i \cdot \mathrm{sign}(Y_{g_{\hat{w}},i}) f(X_i) \right\}. $$
By Cauchy inequality, the first term is bounded by
$$\sqrt{\frac{1}{n} \sum_{i \in [n]} \left( \hat{w}_i - \frac{1}{\pi(A_i, X_i)} \right)^2} \cdot \sqrt{ \frac{1}{n} \sum_{i \in [n]} \pi^2(A_i,X_i) \left( \frac{|Y_i - g_{\hat{w}}(X_i)|}{\pi(A_i, X_i)} \phi\left\{ A_i \cdot \mathrm{sign}(Y_{g_{\hat{w}}}) f(X_i) \right\} \right)^2 } $$
$$ \leq \sqrt{\frac{1}{n} \sum_{i \in [n]} \left( \hat{w}_i - \frac{1}{\pi(A_i, X_i)} \right)^2} \cdot \sqrt{ \frac{1}{n} \sum_{i \in [n]} \left( \frac{|Y_i - g_{\hat{w}}(X_i)|}{\pi(A_i, X_i)} \phi\left\{ A_i \cdot \mathrm{sign}(Y_{g_{\hat{w}}}) f(X_i) \right\} \right)^2 },$$
where the inequality comes from $\pi(A_i,X_i) < 1$. By Corollary \ref{cor:gx_converge}, 
\begin{align}
    \big| |Y - g_{\hat{w}}(x)| - |Y - g(x)| \big| &\leq |Y - g_{\hat{w}}(x) - \{Y - g(x)\}| \nonumber\\
        &= |g(x) - g_{\hat{w}}(x)| = o_p(1). \nonumber
\end{align} 
Moreover, by \textbf{Assumption 2} and compactness of $\mathcal{X}$, $1/\pi(A,X) \leq M_{\pi} < \infty$ where $1/M_{\pi} = \inf_{a \in \mathcal{A}, x \in \mathcal{X}} \pi(a,x)$. Then  
$$ \frac{|Y-g_{\hat{w}}(x)|}{\pi(A,X)} \leq \frac{|Y-g(x)|}{\pi(A,X)} + \frac{|g(x) - g_{\hat{w}}(x)|}{\pi(A,X)} = \frac{|Y-g(x)|}{\pi(A,X)} + o_p(1).$$ 
Thus
$$ \frac{1}{n} \sum_{i \in [n]} \left( \frac{|Y_{g_{\hat{w}},i}|}{\pi(A_i, X_i)} \phi\left\{ A_i \cdot \mathrm{sign}(Y_{g_{\hat{w},i}}) f(X_i) \right\} \right)^2 = \frac{1}{n} \sum_{i \in [n]} \left( \frac{|Y_{g,i}|}{\pi(A_i, X_i)} \phi\left\{ A_i \cdot \mathrm{sign}(Y_{g_{\hat{w},i}}) f(X_i) \right\} \right)^2 + o_p(1), $$
and
\begin{align}
    \frac{1}{n} \sum_{i \in [n]} \left( \frac{|Y_{g,i}|}{\pi(A_i, X_i)} \phi\left\{ A_i \cdot \mathrm{sign}(Y_{g_{\hat{w},i}}) f(X_i) \right\} \right)^2 &\leq \frac{M_g^2}{n} \sum_{i \in [n]} \phi^2\left\{ A_i \cdot \mathrm{sign}(Y_{g,i}) f(X_i) \right\} \nonumber\\
    &~~ + \frac{M_g^2}{n} \sum_{i \in [n]} \left[ \phi\left\{ A_i \mathrm{sign}(Y_{g_{\hat{w}},i}) f(X_i) \right\} - \phi\left\{ A_i \mathrm{sign}(Y_{g,i}) f(X_i) \right\} \right]^2 \nonumber\\
    &\leq \frac{M_g^2}{n} \sum_{i \in [n]} \phi^2\left\{ A_i \cdot \mathrm{sign}(Y_{g,i}) f(X_i) \right\} \nonumber\\
    &~~ + M_g^2 M_f^2 L^2 \frac{1}{n} \sum_{i \in [n]} \{ \mathrm{sign}(Y_{g_{\hat{w}},i}) - \mathrm{sign}(Y_{g,i}) \}^2 \nonumber\\
    &= \frac{M_g^2}{n} \sum_{i \in [n]} \phi^2\left\{ A_i \cdot \mathrm{sign}(Y_{g,i}) f(X_i) \right\} + 4 M_g^2 M_f^2 L^2 C n^{\xi-1} \nonumber\\
    &\to M_g^2 \mathbb{E}[\phi^2\left\{ A \cdot \mathrm{sign}(Y_{g}) f(X) \right\}] \nonumber
\end{align}
in probability. Therefore, by Theorem \ref{thm:WeightConv} and Cauchy-Schwarz inequality,
$$ \frac{1}{n} \sum_{i \in [n]} \left\{ \hat{w}_i - \frac{1}{\pi(A_i, X_i)} \right\} |Y_i - g_{\hat{w}}(X_i)| \phi\left\{ A_i \cdot \mathrm{sign}(Y_{g_{\hat{w}},i}) f(X_i) \right\} = o_p(1).$$
From the given conditions,
$$ \frac{1}{n} \sum_{i \in [n]} \frac{|Y_{g,i}|}{\pi(A_i, X_i)} \left[ \phi\left\{ A_i \cdot \mathrm{sign}(Y_{g_{\hat{w}},i}) f(X_i) \right\} - \phi\left\{ A_i \cdot \mathrm{sign}(Y_{g,i}) f(X_i) \right\} \right] \leq \frac{M_g M_f}{n} 2LCn^{\xi}, $$  %\frac{M_g M_g L}{n} 2LCn^{\xi},$$
almost everywhere. Hence,
$$ \frac{1}{n} \sum_{i \in [n]}  \frac{|Y_{g,i}|}{\pi(A_i, X_i)} \phi\left\{ A_i \cdot \mathrm{sign}(Y_{g_{\hat{w}},i}) f(X_i) \right\} = \frac{1}{n} \sum_{i \in [n]}  \frac{|Y_{g,i}|}{\pi(A_i, X_i)} \phi\left\{ A_i \cdot \mathrm{sign}(Y_{g,i}) f(X_i) \right\} + o_p(1). $$
By Corollary \ref{cor:gx_converge}, \textbf{Assumption 2}, and the weak law of large numbers
$$ \frac{1}{n} \sum_{i \in [n]}  \frac{|Y_i - g_{\hat{w}}(X_i)|}{\pi(A_i, X_i)} \phi\left\{ A_i \cdot \mathrm{sign}(Y_{g_{\hat{w}},i}) f(X_i) \right\} \to \mathbb{E}\left[ \frac{|Y-g(X)|}{\pi(A,X)} \phi\{ A \cdot \mathrm{sign}(Y_g) f(X) \} \right]. $$
in probability. Therefore,
$$\mathbb{P}_n \hat{w} |Y - g_{\hat{w}}(X)| \phi\left\{ A \cdot \mathrm{sign}(Y_{g_{\hat{w}}}) f(X) \right\} \to \mathcal{R}_{\phi, g} (f)$$
in probability.

\section{Proof of Theorem \ref{thm:MatrixConv}}

The proof relies on assumptions (A6) to (A10) adopted from \textcite{fukumizu_gradient-based_2014}: (A6)  $\mathcal{H}_{\mathcal{X}}$ is separable; (A7) $K_X(\cdot, \cdot)$ is measurable, and $\mathbb{E}[K_{X}(X,X)] < \infty$, $\mathbb{E}[Z^2] < \infty$; (A8) $K_X(x, x')$ is continuously differentiable and $\partial K_{X}(\cdot, x) / \partial x^{(a)}$ is in the range of the covariance operator $C_{XX}$ for $a \in [p]$; (A9) $\mathbb{E}[Z | X = \cdot] \in \mathcal{H}_{\mathcal{X}}$; (A10) The function $v \mapsto \mathbb{E}[Z | B_0^\top X=v]$ is differentiable for any $v \in \mathcal{V}$. %, and the linear functional $g \mapsto \partial \varphi_g(v) / \partial z^{(a)}$ is continuous for any $v \in \mathbb{R}^u$ and $a \in [u]$. 
These are not explicitly used in this proof. We additionally introduce assumptions (A11) and (A12) which are also adopted from \textcite{fukumizu_gradient-based_2014}: (A11) For each $p = p_n$, there are $\xi_p \geq 0$ and $C_p \geq 0$ such that a function $h_{a,x} \in \mathcal{H}_{\mathcal{X}}$ satisfies
$$ \frac{\partial K_X(\cdot, x)}{\partial x^{(a)}} = C^{\xi_p + 1}_{XX} h_{a,x},$$
and $\|h_{a,x}\|_{\mathcal{H}_{\mathcal{X}}} \leq C_p$ for any $a \in [p]$ and $x \in \mathcal{X}$; (A12) Define $\alpha_p = \sqrt{\mathbb{E}[K^2_X(X,X)] - \mathbb{E}[K^2_X(X, X')]}$, where $X'$ is an independent copy of $X$. Then $\alpha_p / \sqrt{n} \rightarrow 0$ as $n \rightarrow \infty$.

By triangular inequality,
$$ \left| \hat{W}_{ij}(x)-M_{ij}(x) \right| \leq \left| \hat{W}_{ij}(x)-\hat{M}_{ij}(x) \right| + \left| \hat{M}_{ij}(x)-M_{ij}(x) \right|. $$
The first term:
$$\left| \hat{W}_{ij}(x)-\hat{M}_{ij}(x) \right|$$
$$ \leq \left| \frac{1}{n} \sum_{i \in [n]} (\hat{C}_{XX}+\epsilon_n I)^{-1} \frac{\partial K_X(X_i,x)}{\partial x^{(m)}} \tilde{Z}_i \right| \cdot \left| \frac{1}{n} \sum_{i \in [n]} (\hat{C}_{XX}+\epsilon_n I)^{-1} \frac{\partial K_X(X_i,x)}{\partial x^{(j)}} (\tilde{Z}_i-Z_i) \right| $$
$$ + \left| \frac{1}{n} \sum_{i \in [n]} (\hat{C}_{XX}+\epsilon_n I)^{-1} \frac{\partial K_X(X_i,x)}{\partial x^{(m)}} (\tilde{Z}_i - Z_i) \right| \cdot \left| \frac{1}{n} \sum_{i \in [n]} (\hat{C}_{XX}+\epsilon_n I)^{-1} \frac{\partial K_X(X_i,x)}{\partial x^{(j)}} Z_i \right|. $$
Define $(\hat{C}_{XX} + \epsilon_n I)^{-1} \partial K_X(\cdot, x)/\partial x^{(a)} = \hat{f}_{a,x}(\cdot)$ and $C^{-1}_{XX} \partial K_X(\cdot, x)/\partial x^{(a)} = f_{a,x}(\cdot)$. Let $w^*_i = 1/\pi(A_i,X_i)$. Let $S_1 = \{ i \in [n]: A_i = +1 \}$. Then
\begin{align}
    \frac{1}{n} \sum_{i \in [n]} \hat{f}_{a,x}(X_i) (\tilde{Z}_i - Z_i) &= \frac{1}{n} \sum_{i \in S_1} \hat{w}_i Y_i \hat{f}_{a,x}(X_i) - \frac{1}{n} \sum_{i \in S_1} w^*_i Y_i \hat{f}_{a,x}(X_i) \nonumber\\ 
        &~~ - \left\{ \frac{1}{n} \sum_{i \notin S_1}^{n} \hat{w}_i Y_i \hat{f}_{a,x}(X_i) - \frac{1}{n} \sum_{i \notin S_1} w^*_i Y_i \hat{f}_{a,x}(X_i) \right\} \nonumber\\
        &~~ - \frac{1}{n} \sum_{i \in S_1} \hat{f}_{a,x}(X_i) \hat{w}_i g_{\hat{w}}(X_i) + \frac{1}{n} \sum_{i \notin S_1} \hat{f}_{a,x}(X_i) \hat{w}_i g_{\hat{w}}(X_i) \nonumber\\
        &= \frac{1}{n} \sum_{i \in [n]} A_i (\hat{w}_i - w^*_i) Y_i \hat{f}_{a,x}(X_i) - \frac{1}{n} \sum_{i \in [n]} A_i \hat{w}_i g_{\hat{w}}(X_i) \hat{f}_{a,x}(X_i). \nonumber
\end{align}
Then the first term in the latter equality becomes 
\begin{align}
    \frac{1}{n} \sum_{i \in [n]} A_i(\hat{w}_i - w^*_i) Y_i \hat{f}_{a,x}(X_i) &\leq \sqrt{\frac{1}{n} \sum_{i \in [n]} (\hat{w}_i - w^*_i)^2} \sqrt{\frac{1}{n} \sum_{i \in [n]} Y^2_i \hat{f}^2_{a,x}(X_i)} \nonumber\\
        &= o_p(1) \left\{\sqrt{E\left[Y^2 f^2_{a,x}(X)\right]} + O_p(1/\sqrt{n}) \right\} \nonumber
\end{align}
by law of large numbers \parencite{fukumizu_statistical_2007}, Theorem~\ref{thm:CrossCov} and the finite second moment assumption of the product of $Y$ and the inverse covariance operator. Likewise, the second term, $n^{-1} \sum_{i \in [n]} A_i \hat{w}_i g_{\hat{w}}(X_i) \hat{f}_{a,x}(X_i)$, is $o_p(1)$ from the proof of Theorem~\ref{thm:CrossCov}. Similarly, $n^{-1} \sum_{i \in [n]} \hat{f}_{a,x}(X_i) Z_i$ is $O_p(1)$ and 
\[
\frac{1}{n} \sum_{i \in [n]} \hat{f}_{a,x}(X_i) \tilde{Z}_i = \frac{1}{n} \sum_{i \in [n]} \hat{f}_{a,x}(X_i) (\tilde{Z}_i - Z_i) + \frac{1}{n} \sum_{i \in [n]} \hat{f}_{a,x}(X_i) Z_i = o_p(1) + O_p(1).
\]

From Theorem 3 and Theorem 4 of \cite{fukumizu_gradient-based_2014}, $\left| \hat{M}_{ij}(x) - M_{ij}(x) \right| =  O_p\left(n^{-\min\{1/3, (2\xi+1)/(4\xi+4)\}}\right)$. By combining these results,
$\left| \hat{W}_{ij}(x)-M_{ij}(x) \right| = o_p(1) + O_p\left(n^{-\min\{1/3, (2\xi+1)/(4\xi+4)\}}\right)$. Since the dimensionality is fixed, this rate of convergence applies to $\|\hat{W}(x) - M(x)\|_F$. 

%$$ f_{a,x}(X_i) = \left\langle C^{-1}_{XX} \frac{\partial K(\cdot,x)}{\partial x^{(a)}}, K(\cdot, X_i) \right\rangle $$
%$$ \hat{f}_{a,x}(X_i) = \left\langle \hat{C}^{-1}_{XX} \frac{\partial K(\cdot,x)}{\partial x^{(a)}}, K(\cdot, X_i) \right\rangle $$
%$$ \lim_{n \rightarrow \infty}\hat{f}_{a,x}(X_i) = f_{a,x}(X_i) $$

Define $\tilde{M}_n$ as $\tilde{W}_n$ so that the $(i,j)$-th entry is $n^{-1} \sum_{k \in [n]} \hat{M}_{ij}(X_k)$. By triangular inequality,
$$ \left\| \frac{1}{n}\sum_{i \in [n]} \hat{W}(X_i) - \mathbb{E}[M(X)] \right\|_F \leq \left\| \frac{1}{n}\sum_{i \in [n]}\hat{W}(X_i) - \tilde{M}_n \right\|_F  + \left\|\tilde{M}_n - \mathbb{E}[M(X)] \right\|_F.$$
From the derivation above, the first term on the right hand side of the inequality is bounded as
\begin{align}
    \left\| \frac{1}{n}\sum_{k \in [n]} \hat{W}(X_k) - \tilde{M}_n \right\|^2_F &= \sum_{i \in [p]} \sum_{j \in [p]} \left\{ 
n^{-1} \sum_{k \in [n]}\left\{\hat{W}_{ij}(X_k) - \hat{M}_{ij}(X_k)\right\} \right\}^2 \nonumber\\
    &\leq \sum_{i \in [p]} \sum_{j \in [p]} \left\{ 
n^{-1} \sum_{k \in [n]}\left|\hat{W}_{ij}(X_k) - \hat{M}_{ij}(X_k)\right| \right\}^2 \nonumber\\
    &= p^2 o_p(1). \nonumber
\end{align}
Therefore, $\left\| \frac{1}{n}\sum_{k \in [n]}\hat{W}(X_k) - \tilde{M}_n \right\|_F = o_p(1)$.
By Theorem 3 of \cite{fukumizu_gradient-based_2014}, the second term on the right hand side is
$$\left\|\tilde{M}_n - \mathbb{E}[M(X)] \right\|_F = O_{p}\left( n^{-\min\left\{ \frac{1}{3}, \frac{2 \xi + 1}{4 \xi + 4}\right\}} \right).$$
We obtain the convergence in probability by combining these results.

\begin{remark}
Theorem~\ref{thm:MatrixConv} can be extended to the setting where the number of covariates $p$ grows with the sample size, generalizing to our setting Theorem 4 of \textcite{fukumizu_gradient-based_2014}, under the assumption that the true propensity scores are known. Specifically, under the additional assumptions (A11) and (A12), we can establish
$$ \left\|\hat{M}(x) - M(x)\right\|_F = O_{p}\left(p \cdot C^2_p \left(\frac{\alpha^2_p}{n}\right)^{\min\left\{ \frac{1}{3}, \frac{2 \xi_p + 1}{4 \xi_p + 4}\right\}} \right) $$
for any $x \in \mathcal{X}$, and
$$ \left\| \tilde{M}_n - \mathbb{E}[M(X)] \right\|_F = O_{p}\left(p\cdot \frac{C^2_p}{\sqrt{n}} + p \cdot C^2_p \left(\frac{\alpha^2_p}{n}\right)^{\min\left\{ \frac{1}{3}, \frac{2 \xi_p + 1}{4 \xi_p + 4}\right\}} \right).$$
as $n \to \infty$, where $\alpha_p$, $C_p$, and $\xi_p$ are quantities that depend on the dimensionality $p$. However, establishing analogous results for KCB weights presents significant technical challenges.
\end{remark}

\section{Proof of Lemma \ref{lem:linTransConv}}
Let $Y_{g_{\hat{w}}} = Y-g_{\hat{w}}(X)$, $Y_{g_{\hat{w}},i} = Y_i-g_{\hat{w}}(X_i)$. Recall that  
$$ Q(\tilde{f}^{\hat{V}}) = \frac{1}{n} \sum_{i \in [n]} \hat{w}_i |Y_i - g_{\hat{w}}(X_i)| \phi\left\{ A_i \cdot \text{sign}(Y_{g_{\hat{w}},i}) \tilde{f}^{\hat{V}}(\hat{B}^\top X_i) \right\} + \lambda_n \alpha^\top \hat{G} \alpha. $$
Since $Q(\tilde{f}^{\hat{V}})$ is minimized when $\alpha = \hat{\alpha} \in \mathbb{R}^n$ and $\alpha_0 = \hat{\alpha}^{\hat{V}}_{0n}$,  $\tilde{f}^{\hat{V}}_n(\hat{B}^\top x) = h^{\hat{V}}_n (\hat{B}^\top x) + \alpha^{\hat{V}}_{0n}$, where $h^{\hat{V}}_n (\hat{B}^\top x) = \sum_{i \in [n]} \hat{\alpha}_i K_u(\hat{B}^\top X_i, \hat{B}^\top x)$. Let $\tilde{f}^{\hat{V}}(\hat{B}^\top x) = h^{\hat{V}}(\hat{B}^\top x) + \alpha_0$, where $h^{\hat{V}}(\hat{B}^\top x) = \sum_i \alpha_i K_u(\hat{B}^\top X_i, \hat{B}^\top x)$.  Then,
\begin{align}
     \mathbb{P}_n \hat{w} |Y-g_{\hat{w}}(X)| \phi\left\{ A \cdot \text{sign}(Y_{g_{\hat{w}}}) \tilde{f}^{\hat{V}}_n(\hat{B}^\top X) \right\} &\leq \mathbb{P}_n \hat{w} |Y-g_{\hat{w}}(X)| \phi\left\{ A \cdot \text{sign}(Y_{g_{\hat{w}}}) \tilde{f}^{\hat{V}}_n(\hat{B}^\top X) \right\} + \lambda_n \|h^{\hat{V}}_n\|^2 \nonumber\\
        &\leq \mathbb{P}_n \hat{w} |Y-g_{\hat{w}}(X)| \phi\left\{ A \cdot \text{sign}(Y_{g_{\hat{w}}}) \tilde{f}^{\hat{V}}(\hat{B}^\top X) \right\} + \lambda_n \|h^{\hat{V}}\|^2, \nonumber
\end{align}
where $\|h^{\hat{V}}_n\|^2 = \hat{\alpha}^\top \hat{G} \hat{\alpha}$ and $\|h^{\hat{V}}\|^2 = \alpha^\top \hat{G} \alpha$. We have $|\tilde{f}^{\hat{V}}(\hat{B}^\top x)| \leq \| \tilde{f}^{\hat{V}} \| \cdot K_u(\hat{B}^\top x, \hat{B}^\top x) < \infty$ from Cauchy-Schwarz inequality for a bounded kernel $K_u$. By replacing $f(X)$ by $\tilde{f}^{\hat{V}}(\hat{B}^\top X)$ in Theorem~\ref{thm:ConsistentValue} and from that $\lambda_n \rightarrow 0$,
$$ \mathbb{P}_n \hat{w} |Y-g_{\hat{w}}(X)| \phi\left\{ A \cdot \text{sign}(Y_{g_{\hat{w}}}) \tilde{f}^{\hat{V}}(\hat{B}^\top X) \right\} + \lambda_n \|h^{\hat{V}}\|^2 - \mathcal{R}_{\phi, g}(\tilde{f}^{\hat{V}}) \to 0 $$
in probability. Hence, with probability 1,
$$\inf_{\tilde{f}^{\hat{V}}} \mathcal{R}_{\phi, g}(\tilde{f}^{\hat{V}}) - \mathbb{P}_n \hat{w} |Y-g_{\hat{w}}(X)| \phi\left\{ A \cdot \text{sign}(Y_{g_{\hat{w}}}) \tilde{f}^{\hat{V}}_n (\hat{B}^\top  X) \right\}$$
$$ \leq \mathcal{R}_{\phi, g}(\tilde{f}^{\hat{V}}_n) - \mathbb{P}_n \hat{w} |Y-g_{\hat{w}}(X)| \phi\left\{ A \cdot \text{sign}(Y_{g_{\hat{w}}}) \tilde{f}^{\hat{V}}_n (\hat{B}^\top  X) \right\}.$$
The left hand side is
$$ \inf_{\tilde{f}^{\hat{V}}} \mathcal{R}_{\phi, g}(\tilde{f}^{\hat{V}}) - \mathbb{P}_n \hat{w} |Y-g_{\hat{w}}(X)| \phi\left\{ A \cdot \text{sign}(Y_{g_{\hat{w}}}) \tilde{f}^{\hat{V}}_n (\hat{B}^\top  X) \right\} + \mathcal{R}_{\phi,g}(\tilde{f}^{\hat{V}}_n) - \mathcal{R}_{\phi,g}(\tilde{f}^{\hat{V}}_n) $$
$$ = \inf_{\tilde{f}^{\hat{V}}} \mathcal{R}_{\phi, g}(\tilde{f}^{\hat{V}}) - \mathcal{R}_{\phi,g}(\tilde{f}^{\hat{V}}_n) + \mathcal{R}_{\phi,g}(\tilde{f}^{\hat{V}}_n) - \mathbb{P}_n \hat{w} |Y-g_{\hat{w}}(X)| \phi\left\{ A \cdot \text{sign}(Y_{g_{\hat{w}}}) \tilde{f}^{\hat{V}}_n (\hat{B}^\top  X) \right\}. $$
Therefore, $\mathcal{R}_{\phi, g}(\tilde{f}_n^{\hat{V}}) - \inf_{\tilde{f}^{\hat{V}}} \mathcal{R}_{\phi, g}(\tilde{f}^{\hat{V}}) \rightarrow 0$ as long as 
$$\mathbb{P}_n \hat{w}|Y_{g_{\hat{w}}}| \phi\left( A \cdot \text{sign}(Y_{g_{\hat{w}}}) \tilde{f}^{\hat{V}}_n(\hat{B}^\top X) \right) + \lambda_n \|h^{\hat{V}}_n\|^2 - \mathcal{R}_{\phi, g}(\tilde{f}^{\hat{V}}_n) = o_p(1).$$
Since $\mathcal{R}_{\phi, g_{\hat{w}}}(\tilde{f}^{\hat{V}}_n) - \mathcal{R}_{\phi, g}(\tilde{f}^{\hat{V}}_n) = o_p(1)$ by continuous mapping theorem, we show that 
$$\mathbb{P}_n \hat{w}|Y_{g_{\hat{w}}}| \phi\left( A \cdot \text{sign}(Y_{g_{\hat{w}}}) \tilde{f}^{\hat{V}}_n(B_0^\top X) \right) + \lambda_n \|h^{\hat{V}}_n\|^2 - \mathcal{R}_{\phi, g_{\hat{w}}}(\tilde{f}^{\hat{V}}_n) = o_p(1).$$
For any $h^{\hat{V}}(\hat{B}^\top x) = \sum_i \alpha_i K_u(\hat{B}^\top X_i, \hat{B}^\top x)$ and $\alpha_0 \in \mathbb{R}$,
$$ \mathbb{P}_n \hat{w}|Y_{g_{\hat{w}}}| \phi\left\{ A \cdot \text{sign}(Y_{g_{\hat{w}}}) \tilde{f}^{\hat{V}}_n(\hat{B}^\top X) \right\} + \lambda_n \|h^{\hat{V}}_n\|^2 \leq \mathbb{P}_n \hat{w}|Y_{g_{\hat{w}}}| \phi\left( A \cdot \text{sign}(Y_{g_{\hat{w}}}) \{ h^{\hat{V}}(\hat{B}^\top X) + \alpha_0 \} \right) + \lambda_n \|h^{\hat{V}}\|^2, $$
hence we can choose $h^{\hat{V}}=0$ $(\alpha_i = 0 ~~ \forall i \in \mathbb{N})$ and $\alpha_0=0$ so that
\begin{align}
   \mathbb{P}_n \hat{w}|Y_{g_{\hat{w}}}| \phi\left( A \cdot \text{sign}(Y_{g_{\hat{w}}}) \tilde{f}^{\hat{V}}_n(\hat{B}^\top X) \right) + \lambda_n \|h^{\hat{V}}_n\|^2 &\leq \phi(0) \mathbb{P}_n \hat{w} |Y_{g_{\hat{w}}}| \nonumber\\
    &= \phi(0) \mathbb{P}_n \hat{w} \pi(A,X) \frac{|Y - g_{\hat{w}}(X)|}{\pi(A,X)} \nonumber\\
    &\leq \phi(0) \mathbb{P}_n \hat{w} \frac{|Y - g_{\hat{w}}(X)|}{\pi(A,X)} \nonumber
    %&\leq \phi(0) M_w M_g n^{2/3}. \nonumber
\end{align}
Define $K_i = X^\top (n^{-1} \mathbb{X}^\top \mathbb{X})^{-1} X_i (\hat{w}_i - 1) \in \mathbb{R}$. Then $|K_i| \leq M_X M_w n^{1/3}$, where $M_X < \infty$ is the bound such that $|X^\top (n^{-1} \mathbb{X}^\top \mathbb{X})^{-1} X_i| \leq M_X$. When $n^{-1} \mathbb{X}^\top \mathbb{X}$ is singular, we can use ridge-regularization and replace it with $n^{-1} \mathbb{X}^\top \mathbb{X} + \lambda I_p$, where $\lambda > 0$. We have $1/\pi(A,X) \leq M_{\pi} < \infty$ as in the proof of Theorem \ref{thm:ConsistentValue} by \textbf{Assumption 2} and the compactness of $\mathcal{X}$. Furthermore, since $g(x)$ is well-defined and $\mathcal{X}$ is compact, we have $|g(X)| \leq M$ almost surely for $M < \infty$. Note that
\begin{align}
    \frac{|Y - g_{\hat{w}}(X)|}{\pi(A,X)} &= \frac{|Y - g(X) + g(X) - g_{\hat{w}}(X)|}{\pi(A,X)} \nonumber\\
        &\leq \frac{|Y - g(X)|}{\pi(A,X)} + \frac{|g(X) - g_{\hat{w}}(X)|}{\pi(A,X)} \nonumber\\
        &\leq M_g + \frac{|n^{-1} \sum_{i \in [n]} g(X_i) - n^{-1} \sum_{i \in [n]} X^\top (n^{-1} \mathbb{X}^\top \mathbb{X})^{-1} X_i (\hat{w}_i - 1) Y_i |}{\pi(A,X)} \nonumber\\
        &= M_g + \frac{|n^{-1} \sum_{i \in [n]} \{g(X_i) - K_i Y_i \}|}{\pi(A,X)} \nonumber\\
        &= M_g + \frac{|n^{-1} \sum_{i \in [n]} \{K_i g(X_i) - K_i Y_i + (1-K_i) g(X_i) \}|}{\pi(A,X)} \nonumber\\
        &\leq M_g + \frac{n^{-1} \sum_{i \in [n]} |K_i| |g(X_i) - Y_i|}{\pi(A,X) } + \frac{n^{-1} \sum_{i \in [n]}|1-K_i| |g(X_i)| }{\pi(A,X)} \nonumber\\
        &\leq M_g + \frac{n^{-1} \sum_{i \in [n]} |K_i| |g(X_i) - Y_i|/\pi(A_i,X_i)}{\pi(A,X) } + \frac{n^{-1} \sum_{i \in [n]}|1-K_i| |g(X_i)|}{\pi(A,X)} \nonumber\\
        &\leq M_g + M_g M_X M_{\pi} M_w n^{1/3} + M M_{\pi} + M_X M_{\pi}  M M_w n^{1/3} \nonumber \\
        &\leq  \{M_g + M_g M_X M_{\pi} M_w + (M_X M_w + 1) M_{\pi}  M \}  n^{1/3}. \nonumber
\end{align}
Define $M_h$ as $ M_h = \sqrt{\phi(0) M_w \{M_g + M_g M_X M_{\pi} M_w + (M_X M_w + 1) M_{\pi}  M\}} $.
Then, $\lambda^*_n \|h^{\hat{V}}_n\|^2 \leq M^2_h$, where $\lambda^*_n = n^{-2/3}\lambda_n$. From this, we have the bound $ \|\sqrt{\lambda^*_n} h^{\hat{V}}_n\| \leq M_h$. Note that we have $|\sqrt{\lambda^*_n} \alpha^{\hat{V}}_{0n}| \leq |\sqrt{\lambda_n} \alpha^{\hat{V}}_{0n}| < M_{\alpha_0}$, $\lambda^*_n \rightarrow 0$, and $n\lambda^*_n \rightarrow \infty$.

Define the RKHS with bounded kernel as $\mathcal{H}_H = \{\sqrt{\lambda^*_n} h: \|\sqrt{\lambda^*_n} h\| \leq M_h\}$. Since the RKHS norm is bounded, $|\sqrt{\lambda^*_n} h^{\hat{V}}_n| \leq \bar{H}$ for an envelope function $\bar{H}$ such that $\mathbb{P}\bar{H}^2 < \infty$. Also, from the proof of Lemma A.9 of \textcite{hable_asymptotic_2012}, $\mathcal{H}_H$ satisfies the uniform entropy bound. Thus $\mathcal{H}_H$ is $\mathbb{P}$-Donsker. Likewise, $\{\sqrt{\lambda^*_n} (h+\alpha_0): \|\sqrt{\lambda^*_n} h\| \leq M_h, |\sqrt{\lambda^*_n}\alpha_0|\leq M_{\alpha_0} \}$ is $\mathbb{P}$-Donsker.
Let $C$ be the Lipschitz constant of $\phi(\cdot)$. Define $\phi_{\lambda^*_n}(c) = \sqrt{\lambda^*_n} \phi(c / \sqrt{\lambda^*_n})$. Note that
\begin{align}
    \phi_{\lambda^*_n}(c_2) - \phi_{\lambda^*_n}(c_1) &= \sqrt{\lambda^*_n}\phi(c_2/\sqrt{\lambda^*_n}) - \sqrt{\lambda^*_n}\phi(c_1/\sqrt{\lambda^*_n}) \nonumber\\
        &= \sqrt{\lambda^*_n} C \cdot (c_2/\sqrt{\lambda^*_n} - c_1/\sqrt{\lambda^*_n})= C(c_2 - c_1). \nonumber
\end{align}
Hence, $\phi_{\lambda^*_n}(\cdot)$ is Lipschitz continuous. Also,
$$ \sqrt{\lambda^*_n}\frac{|Y_{g_{\hat{w}}}|}{\pi(A,X)} \phi\left( A \cdot \text{sign}(Y_{g_{\hat{w}}}) \{ h(\hat{B}^\top X) + \alpha_0\} \right) = \frac{|Y_{g_{\hat{w}}}|}{\pi(A,X)} \phi_{\lambda^*_n}\left( A \sqrt{\lambda^*_n} \cdot \text{sign}(Y_{g_{\hat{w}}}) \{ h(\hat{B}^\top X) + \alpha_0\} \right). $$
Since $|Y - g_{\hat{w}}(X)|/\pi(A,X) \leq M_g$ and $\phi_{\lambda^*_n}(\cdot)$ is Lipschitz, it follows that 
$$\left\{\sqrt{\lambda^*_n} \frac{|Y_{g_{\hat{w}}}|}{\pi(A,X)} \phi\left( A \cdot \text{sign}(Y_{g_{\hat{w}}}) \{ h(\hat{B}^\top X) + \alpha_0\} \right): \|\sqrt{\lambda^*_n}h\| \leq M_h, |\sqrt{\lambda^*_n}\alpha_0| \leq M_{\alpha_0}\right\}$$ 
is $\mathbb{P}$-Donsker by Corollary 9.32-($iv$) of \textcite{kosorok_introduction_2008}. This implies that 
$$ \mathbb{P}_n\frac{|Y_{g_{\hat{w}}}|}{\pi(A,X)} \phi\left( A \cdot \text{sign}(Y_{g_{\hat{w}}})  \tilde{f}^{\hat{V}}_n(\hat{B}^\top X) \right) - \mathcal{R}_{\phi,{g_{\hat{w}}}}(\tilde{f}^{\hat{V}}_n) = O_p(1/\sqrt{n\lambda^*_n}). $$
Since $n\lambda^*_n \rightarrow \infty$ and by Theorem~\ref{thm:ConsistentValue} we have
$$ \mathbb{P}_n \hat{w} |Y_{g_{\hat{w}}}| \phi\left( A \cdot \text{sign}(Y_{g_{\hat{w}}}) \tilde{f}^{\hat{V}}_n(\hat{B}^\top X) \right) - \mathcal{R}_{\phi, {g_{\hat{w}}}}(h^{\hat{V}}_n + \alpha^{\hat{V}}_{0n})$$
$$ = \mathbb{P}_n \hat{w} |Y_{g_{\hat{w}}}| \phi\left( A \cdot \text{sign}(Y_{g_{\hat{w}}}) \tilde{f}^{\hat{V}}_n(\hat{B}^\top X) \right) - \mathbb{P}_n \frac{|Y_{g_{\hat{w}}}|}{\pi(A,X)} \phi\left( A \cdot \text{sign}(Y_{g_{\hat{w}}}) \tilde{f}^{\hat{V}}_n(\hat{B}^\top X) \right)$$
$$ +\mathbb{P}_n \frac{|Y_{g_{\hat{w}}}|}{\pi(A,X)} \phi\left( A \cdot \text{sign}(Y_{g_{\hat{w}}}) \tilde{f}^{\hat{V}}_n(\hat{B}^\top X) \right) - \mathcal{R}_{\phi, {g_{\hat{w}}}}(h^{\hat{V}}_n + \alpha^{\hat{V}}_{0n})$$
$$= o_p(1) + O_p\left(1/\sqrt{n\lambda^*_n}\right) = o_p(1). $$
This implies that $\mathcal{R}_{\phi, g}(\tilde{f}^{\hat{V}}_n) - \inf_{\hat{f}^{\hat{V}}} \mathcal{R}_{\phi, g}(\tilde{f}^{\hat{V}}) \rightarrow 0$ in probability.

\section{Proof of Proposition 1}

We establish two lemmas on the excess risk bound and the expressive power of RKHS from a universal kernel prior to presenting the proof.

    \subsection{Expressive power of an RKHS from a universal kernel}    

Let $\tilde{f}^{\hat{V}*}_{\phi, g}$ be a minimizer of $\mathcal{R}_{\phi,g}(\tilde{f})$ over all continuous functions $\tilde{f}: \hat{\mathcal{V}} \mapsto \mathbb{R}$ and $\mathcal{R}^{\hat{V}*}_{\phi,g} = \mathcal{R}_{\phi,g}(\tilde{f}^{\hat{V}*}_{\phi, g})$. We present a slight modification of Lemma 2.5 of \textcite{zhou_augmented_2017} to establish that the minimum of $\mathcal{R}_{\phi,g}$ over RKHS functions $\tilde{f}^{\hat{V}}:\hat{\mathcal{V}} \mapsto \mathbb{R}$ from a universal kernel is equivalent to $\mathcal{R}^{\hat{V}*}_{\phi,g}$. Note that the difference from Lemma 2.5 of \textcite{zhou_augmented_2017} is that the decision function is defined on $\hat{\mathcal{V}}$ instead of $\mathcal{X}$. However, since $\hat{\mathcal{V}}$ consists of linear transformed variates of $\mathcal{X}$, we can apply Lusin's theorem by using the regular measure $\mu$ on $\mathcal{X}$. Then, we can approximate $\tilde{f}^{\hat{V}*}_{\phi, g}$ from a continuous function $\tilde{f}_1(\hat{B}^\top x)$ such that for all $\epsilon >0$,
$$ \mu\left\{ x \in \mathcal{X}; \tilde{f}_1(\hat{B}^\top x) \neq \tilde{f}^{\hat{V}*}_{\phi, g}(\hat{B}^\top x) \right\} \leq \epsilon. $$
The proof easily follows from the proof of Lemma 2.5 of \textcite{zhou_augmented_2017} by replacing $f(X)$ to $\tilde{f}(\hat{B}^\top X)$.

\begin{customlemma}{S3}
\label{lem:S3}
Suppose that the covariate space $\mathcal{X} \in \mathbb{R}^p$ is a compact metric space and that the RKHS of functions $\tilde{f}^{\hat{V}}: \hat{\mathcal{V}} \mapsto \mathbb{R}$ uses the universal kernel. Suppose that $\phi$ is Lipschitz continuous, and $\tilde{f}^{\hat{V}*}_{\phi,g}:\hat{\mathcal{V}} \mapsto \mathbb{R}$ is measurable and bounded for any $\hat{B}$: $|\tilde{f}^{\hat{V}*}_{\phi,g}| \leq M_{\tilde{f}} < \infty$. Given any distribution $P$ of $(X,A,Y)$ with the bound $|Y_g|/\pi(A,X) \leq M_g < \infty$ almost everywhere with regular marginal distribution on $X$, we have
$$ \inf_{\tilde{f}^{\hat{V}}} \mathcal{R}_{\phi, g}(\tilde{f}^{\hat{V}}) = \mathcal{R}^{\hat{V}*}_{\phi, g}. $$

\end{customlemma}

    \subsection{Excess risk bound}

We present a slight modification of Theorem 2.2 of \textcite{zhou_augmented_2017} to establish the bound of excess risk, $\mathcal{R}(\tilde{f}^{\hat{V}}_n) - \mathcal{R}^*$, by the excess $(\phi, g)$-risk, $\mathcal{R}_{\phi, g}(\tilde{f}^{\hat{V}}_n) - \mathcal{R}^*_{\phi, g}$. 

Prior to introducing the bound, we define $\eta_1(x)$ and $\eta_2(x)$ as follows:
$$ \eta_1(x) = \mathbb{E}[\max\{Y-g(X),0\} | X=x,A=+1] - \mathbb{E}[\max\{ -(Y-g(X)),0\} | X=x,A=-1]; $$
$$ \eta_2(x) = \mathbb{E}[\max\{Y-g(X),0\} | X=x,A=-1] - \mathbb{E}[\max\{ -(Y-g(X)),0\} | X=x,A=+1]. $$
By algebra, we have
$$ \eta_1(x) - \eta_2(X) = \mathbb{E}[Y|X=x, A=+1] -\mathbb{E}[Y|X=x, A=-1], $$
and that
$$ \mathcal{R}_{\phi, g}(f) = \mathbb{E}[\eta_1(X) \phi\{f(X)\} + \eta_2(X) \phi\{-f(X)\}], $$
%$$ \mathcal{R}_{\phi, g}(\tilde{f}^{\hat{V}}) = \mathbb{E}[\eta_1(X) \phi\{\tilde{f}^{\hat{V}} (\hat{B}^\top X)\} + \eta_2(X) \phi\{-\tilde{f}^{\hat{V}} (\hat{B}^\top X)\}]. $$
where $\tilde{f}^V(B_0^\top X)$ and $\tilde{f}^{\hat{V}}(\hat{B}^\top X)$ can be used in place of $f$ and $f(X)$. Motivated by this expression, we can define a generic conditional $(\phi,g)$-risk function as 
$$q_{\eta_1, \eta_2}(\alpha) = \eta_1 \phi(\alpha) + \eta_2 \phi(-\alpha)$$
for $\eta_1 \geq 0$, $\eta_2 \geq 0$, and $\alpha \in \mathbb{R}$. As long as $\phi(\cdot)$ is convex, $q_{\eta_1, \eta_2}(\cdot)$ is convex. Note that 
$$ \mathbb{E}[q_{\eta_1(X), \eta_2(X)} (f)] = \mathcal{R}_{\phi, g}(f). $$

\begin{customlemma}{S4}
\label{lem:S4}
Assume $\phi(\cdot)$ is convex, $\phi'(0)$ exists and $\phi'(0) < 0$. Suppose that for constants $C > 0$ and $s \leq 1$ such that 
$$ |\eta_1 - \eta_2|^s \leq C^s \left\{ q_{\eta_1, \eta_2}(0) - \min_{\alpha \in \mathbb{R}} q_{\eta_1, \eta_2}(\alpha) \right\}. $$
Then, for a decision rule $\tilde{f}^{\hat{V}} \in \mathcal{H}_{\mathcal{\hat{V}}}$,
$$\mathcal{R}(\tilde{f}^{\hat{V}}) - \mathcal{R}^* \leq C \{ \mathcal{R}_{\phi, g}(\tilde{f}^{\hat{V}}) - \mathcal{R}^*_{\phi, g} \}^{1/s}.$$
\end{customlemma}
\begin{proof}
    The proof mostly follows from the proof of Theorem 2.2 of \textcite{zhou_augmented_2017} by replacing $f(X)$ with $\tilde{f}^{\hat{V}}(\hat{B}^\top X)$. For the optimal Bayes decision function $\tilde{f}^*$ such that $d^*(B_0^\top x) = \mathrm{sign} \circ \tilde{f}^*(x)$ and $\mathcal{R}^* = \mathcal{R}(\tilde{f}^*)$,
    $$ \mathbb{E}\left[ \frac{Y}{\pi(A,X)} \mathbbm{1}\{ A \neq \mathrm{sign} \circ \tilde{f}^{\hat{V}}(\hat{B}^\top X) \} |X=x \right] - \mathbb{E}\left[ \frac{Y}{\pi(A,X)} \mathbbm{1}\{ A \neq \mathrm{sign} \circ \tilde{f}^*(B_0^\top X) \} \mid X=x \right] $$
    $$ = \left\{ \mathbb{E}[Y|X=x, A=+1] - \mathbb{E}[Y|X=x, A=-1] \right\} \left[ \mathbbm{1}\{ \mathrm{sign} \circ \tilde{f}^*(B_0^\top x) \} - \mathbbm{1}\{ \mathrm{sign} \circ \tilde{f}^{\hat{V}}(\hat{B}^\top x) \} \right] $$
    $$ \leq |\eta_1(x) - \eta_2(x)| \cdot \mathbbm{1} \left\{ \mathrm{sign} \circ \tilde{f}^{\hat{V}}(\hat{B}^\top x) \big( \eta_1(x) - \eta_2(x) \big) < 0 \right\}. $$
    Taking expectation on the left hand side of the equality leads to $\mathcal{R}(\tilde{f}^{\hat{V}}) - \mathcal{R}^*$. Let $\Delta q_{\eta_1(X), \eta_2(X)}(0) = q_{\eta_1, \eta_2}(0) - \min_{\alpha \in \mathbb{R}} q_{\eta_1, \eta_2}(\alpha)$. By taking the expectation on the right hand side of the inequality, we have
    \begin{align}
        \mathcal{R}(\tilde{f}^{\hat{V}}) - \mathcal{R}^* &\leq \mathbb{E}\left[ |\eta_1(X) - \eta_2(X)| \cdot \mathbbm{1} \left\{ \mathrm{sign} \circ \tilde{f}^{\hat{V}}(\hat{B}^\top X) \big( \eta_1(X) - \eta_2(X) \big) < 0 \right\} \right] \nonumber\\
            &\leq \left( \mathbb{E}\left[ |\eta_1(X) - \eta_2(X)|^s \cdot \mathbbm{1} \left\{ \mathrm{sign} \circ \tilde{f}^{\hat{V}}(\hat{B}^\top X) \big( \eta_1(X) - \eta_2(X) \big) < 0 \right\} \right] \right)^{1/s} \nonumber\\
            &\leq C \cdot \left( \mathbb{E}\left[ \Delta q_{\eta_1(X), \eta_2(X)}(0) \cdot \mathbbm{1} \left\{ \mathrm{sign} \circ \tilde{f}^{\hat{V}}(\hat{B}^\top X) \big( \eta_1(X) - \eta_2(X) \big) < 0 \right\} \right] \right)^{1/s} \nonumber\\
            &\leq C \cdot \left( \mathbb{E}\left[ \Delta q_{\eta_1(X), \eta_2(X)}(\tilde{f}^{\hat{V}}) \cdot \mathbbm{1} \left\{ \mathrm{sign} \circ \tilde{f}^{\hat{V}}(\hat{B}^\top X) \big( \eta_1(X) - \eta_2(X) \big) < 0 \right\} \right] \right)^{1/s} \nonumber\\
            &\leq C \cdot \left( \mathbb{E}\left[ \Delta q_{\eta_1(X), \eta_2(X)}(\tilde{f}^{\hat{V}}) \right] \right)^{1/s} =  C \left( \mathcal{R}_{\phi,g}(\tilde{f}^{\hat{V}}) - \mathcal{R}^*_{\phi, g} \right)^{1/s}, \nonumber
    \end{align}
    where the second inequality follows from Jensen's inequality, and the third inequality follows from the given condition. From the given condition on $\phi(\cdot)$, $\phi(\cdot)$ is Fisher consistent. Since $\mathrm{sign}(\tilde{f}^{\hat{V}}) \cdot (\eta_1-\eta_2) < 0$ implies that 0 lies in between $\tilde{f}^{\hat{V}}$ and $\tilde{f}^*$, $q_{\eta_1, \eta_2}(0) \leq q_{\eta_1, \eta_2}(\tilde{f}^{\hat{V}})$. Thus, the fourth inequality follows from that $\Delta q_{\eta_1(X), \eta_2(X)}(0) \leq \Delta q_{\eta_1(X), \eta_2(X)}(\tilde{f}^{\hat{V}})$.
\end{proof}

In case of hinge loss, we have that $q_{\eta_1, \eta_2}(0) - \min_{\alpha \in \mathbb{R}} q_{\eta_1, \eta_2}(\alpha) = |\eta_1 - \eta_2|$. Thus, when $C=1$ and $s=1$ in Lemma \ref{lem:S4},
$$\mathcal{R}(\tilde{f}^{\hat{V}}_n) - \mathcal{R}^* \leq \mathcal{R}_{\phi, g}(\tilde{f}^{\hat{V}}_n) - \mathcal{R}^*_{\phi, g}.$$

    \subsection{Proof of proposition}

Let $\lambda^*_n = n^{-1/3} \lambda_n$. It follows from the proof of Proposition 2.7 of \textcite{zhou_augmented_2017} that $\sqrt{\lambda^*_n}\alpha^{\hat{V}}_{0n}$ is bounded. From the proof of Lemma \ref{lem:linTransConv}, we have that $\| \sqrt{\lambda^*_n} h^{\hat{V}}_n \| < \infty$. Note that $|\tilde{f}^{\hat{V}*}_{\phi, g}|\leq 1$ for hinge loss from Appendix A of \textcite{zhou_augmented_2017}. Then by Lemma \ref{lem:S3}, $\inf_{\tilde{f}^{\hat{V}}} \mathcal{R}_{\phi, g}(\tilde{f}^{\hat{V}}) = \mathcal{R}^{\hat{V}*}_{\phi, g}$. By combining these results and the triangular inequality, 
\begin{align}
    \mathcal{R}_{\phi,g}(\tilde{f}^{\hat{V}}_n) - \mathcal{R}^*_{\phi, g} &\leq |\mathcal{R}_{\phi,g}(\tilde{f}^{\hat{V}}_n) - \mathcal{R}^{\hat{V}*}_{\phi,g}| + |\mathcal{R}^{\hat{V}*}_{\phi,g} - \mathcal{R}^{*}_{\phi,g}| \nonumber\\
        &\leq |\mathcal{R}_{\phi,g}(\tilde{f}^{\hat{V}}_n) - \inf_{\tilde{f}^{\hat{V}}}\mathcal{R}_{\phi,g}(\tilde{f}^{\hat{V}})| + | \inf_{\tilde{f}^{\hat{V}}}\mathcal{R}_{\phi,g}(\tilde{f}^{\hat{V}}) - \mathcal{R}^{\hat{V}*}_{\phi,g} | + |\mathcal{R}^{\hat{V}*}_{\phi,g} - \mathcal{R}^{*}_{\phi,g}|. \nonumber
\end{align}
From Lemma \ref{lem:linTransConv}, $|\mathcal{R}_{\phi,g}(\tilde{f}^{\hat{V}}_n) - \inf_{\tilde{f}^{\hat{V}}}\mathcal{R}_{\phi,g}(\tilde{f}^{\hat{V}})| = o_p(1)$. Lemma \ref{lem:S3} implies $| \inf_{\tilde{f}^{\hat{V}}}\mathcal{R}_{\phi,g}(\tilde{f}^{\hat{V}}) - \mathcal{R}^{\hat{V}*}_{\phi,g} | = 0$. For a uniformly bounded continuous function $\tilde{f}:\mathbb{R}^u \mapsto \mathbb{R}$, we have $\tilde{f}(\hat{B}^\top x) \to \tilde{f}(B_0^\top x)$ by continuous mapping theorem and Corollary \ref{cor:eigenvectorConv}. Thus, $\inf_{\tilde{f}|_{\hat{\mathcal{V}}}} \mathcal{R}_{\phi, g}(\tilde{f}) \to \inf_{\tilde{f}|_{\mathcal{V}}} \mathcal{R}_{\phi, g}(\tilde{f})$, from which, $|\mathcal{R}^{\hat{V}*}_{\phi,g} - \mathcal{R}^{*}_{\phi,g}| = o_p(1)$. When $\phi(\cdot)$ is the hinge loss, Lemma \ref{lem:S4} implies $\mathcal{R}(\tilde{f}^{\hat{V}}_n) - \mathcal{R}^* \leq  \mathcal{R}_{\phi,g}(\tilde{f}^{\hat{V}}_n) - \mathcal{R}^*_{\phi, g}$. Therefore, we have $\mathcal{R}(\tilde{f}^{\hat{V}}_n) - \mathcal{R}^* = o_p(1)$.

\begin{figure}
\centering
\includegraphics[width = 5.5in]{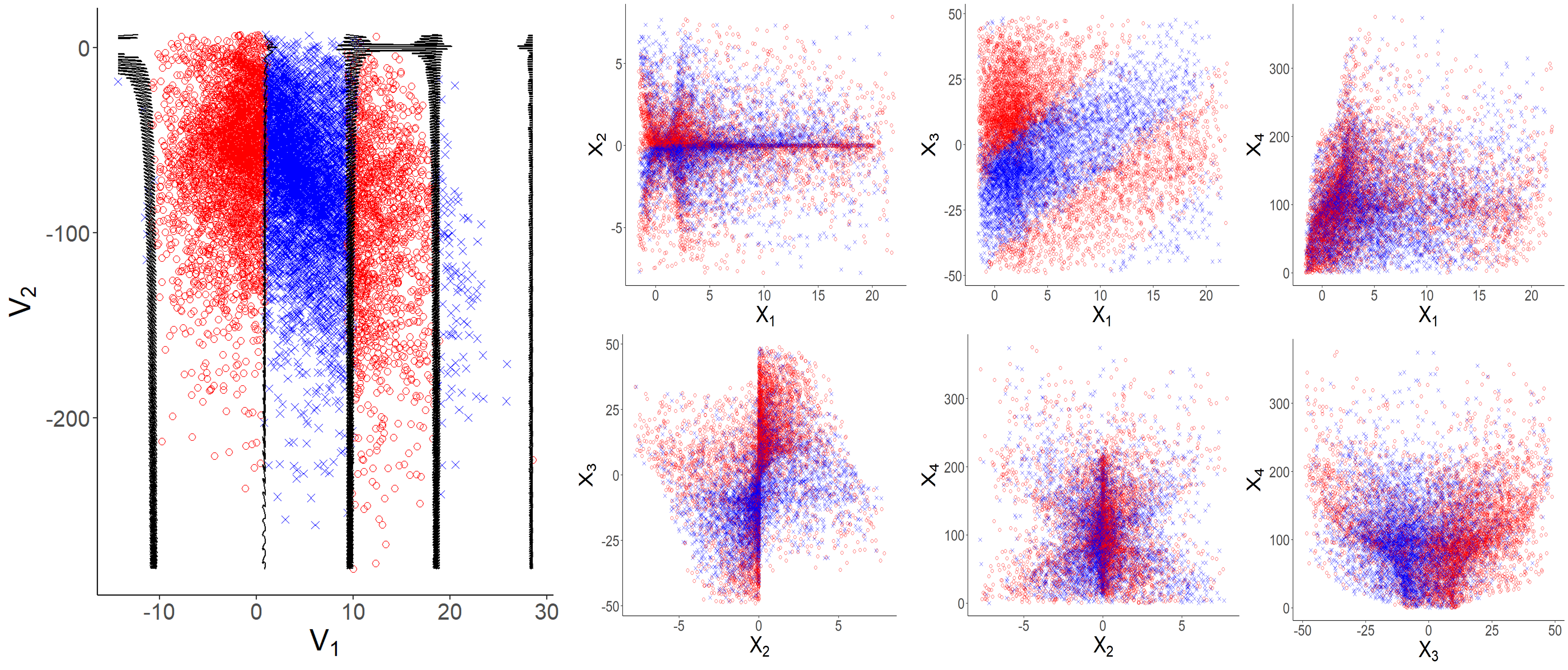}
\caption{The distribution of the optimal treatment regimes according to different levels of covariates in Setting 2. The red circles and blue crosses are data points from the test dataset and represent $d^*(X)=-1$ (or $d^*(V_0)=-1$) and $d^*(X)=+1$ (or $d^*(V_0)=+1$), respectively. The solid lines on the left plot display the decision boundaries in the dimension-reduced space $\left\{v_0 = \left(v^{(1)}_0, v^{(2)}_0 \right); \tilde{f}^*(v_0) = 0\right\}$.}
\label{fig:Setting2}
\end{figure}

\section{Equivalence of the risk and its alternative form}
In Section \ref{sec:AOL}, we claimed that the minimization of $\mathcal{R}(f) = \mathbb{E}\left[ Y/\pi(A, X) \mathbbm{1}\{\mathrm{sign} \circ f(X) \neq A\} \right]$ is equivalent to the minimization of  $\mathbb{E}[ |Y - g(X)|/\pi(A,X) \mathbbm{1}\{A \cdot \text{sign}(Y - g(X)) \neq \mathrm{sign} \circ f(X)\} ]$.

Let $\tilde{m}(a,x) = \mathbb{E}[ x^\top (E[X X^\top])^{-1}X \pi(-a, X)/\pi(-a, x) \mathbb{E}[Y \mid X, A= a]]$ for $a \in \{-1, +1\}$. Then $g(x)$ can be re-expressed as
$$ g(x) = \pi(-1,x) \tilde{m}(+1, x) + \pi(+1,x) \tilde{m}(-1, x). $$
Note that $d(x) = \mathrm{sign} \circ f(x)$. Define 
$$\tilde{h}(x,d) = \tilde{m}(+1,x) \mathbbm{1}\{d(x) = +1\} + \tilde{m}(-1,x) \mathbbm{1}\{d(x) = -1\}.$$ 
First, we show that
$$ \mathbb{E}\left[ \frac{Y-\tilde{h}(X,d)}{\pi(A,X)} \mathbbm{1}\{d(X) = A\} + \tilde{h}(X,d) \right] =  \mathbb{E}\left[ \frac{Y}{\pi(A,X)} \mathbbm{1}\{d(X) = A\} \right], $$
where the left hand side is the doubly robust form of the value function introduced in \textcite{zhang_robust_2012}. Note that
\begin{align}
    \mathbb{E}\left[ \tilde{h}(X,d) - \frac{\tilde{h}(X,d) \mathbbm{1}\{d(X)=A\}}{\pi(A,X)} \right] &=  \mathbb{E}\left[ \tilde{h}(X,d) - \mathbb{E}\left[ \frac{\tilde{h}(X,d) \mathbbm{1}\{d(X)=A\}}{\pi(A,X)} \mid X \right] \right] \nonumber\\
        &= \mathbb{E}[ \tilde{h}(X,d) ] - \mathbb{E}\left[ \mathbb{E}\left[ \frac{\tilde{h}(X,d) \mathbbm{1}\{d(X)=+1\}}{\pi(+1,X)} \mid X,A=+1 \right] \pi(+1,X)\right] \nonumber\\ 
        &~~ - \mathbb{E}\left[ \mathbb{E}\left[ \frac{\tilde{h}(X,d) \mathbbm{1}\{d(X)=-1\}}{\pi(-1,X)} \mid X,A=-1 \right] \pi(-1,X)\right] \nonumber\\
        &= \mathbb{E}[ \tilde{h}(X,d) ] - \mathbb{E}[\tilde{h}(X,d) \mathbbm{1}\{d(X) +1\}] - \mathbb{E}[\tilde{h}(X,d) \mathbbm{1}\{d(X) -1\}] \nonumber\\
        &= \mathbb{E}[ \tilde{h}(X,d) ] - \mathbb{E}[\tilde{h}(X,d)] = 0. \nonumber
\end{align}
Therefore,
\begin{align}
     \mathbb{E}\left[ \frac{Y-\tilde{h}(X,d)}{\pi(A,X)} \mathbbm{1}\{d(X) = A\} + \tilde{h}(X,d) \right] &= \mathbb{E}\left[ \frac{Y}{\pi(A,X)} \mathbbm{1}\{d(X) = A\} \right] + \mathbb{E}\left[ \tilde{h}(X,d) - \frac{\tilde{h}(X,d) \mathbbm{1}\{d(X)=A\}}{\pi(A,X)} \right] \nonumber\\
     &=\mathbb{E}\left[ \frac{Y}{\pi(A,X)} \mathbbm{1}\{d(X) = A\} \right]. \nonumber
\end{align}

The integrand on the left hand side can be rearranged as follows
\begin{align}
    \frac{Y-\tilde{h}(X,d)}{\pi(A,X)} \mathbbm{1}\{d(X) = A\} + \tilde{h}(X,d) &= \frac{Y-g(X) + g(X)- \tilde{h}(X,d)}{\pi(A,X)} \mathbbm{1}\{d(X) = A\} + \tilde{h}(X,d) \nonumber\\
        &= \frac{Y-g(X)}{\pi(A,X)} \mathbbm{1}\{d(X) = A\} \nonumber\\
        &~~ + \frac{g(X)- \tilde{h}(X,d)}{\pi(A,X)} \mathbbm{1}\{d(X) = A\} + \tilde{h}(X,d). \nonumber
\end{align}
The last two terms become
\begin{align}
    \frac{g(X)- \tilde{h}(X,d)}{\pi(A,X)} \mathbbm{1}\{d(X) = A\} + \tilde{h}(X,d) &= \frac{[\pi(-1,X) - \mathbbm{1}\{d(X)=+1\}]\tilde{m}(+1,X)}{\pi(A,X)}\mathbbm{1}\{d(X) = A\} \nonumber\\
        &~~ + \frac{[\pi(+1,X) - \mathbbm{1}\{d(X)=-1\}]\tilde{m}(-1,X) }{\pi(A,X)} \mathbbm{1}\{d(X) = A\} \nonumber\\
        &~~ + \tilde{m}(+1,X) \mathbbm{1}\{d(X)=+1\} + \tilde{m}(-1,X) \mathbbm{1}\{d(X)=-1\}. \nonumber
\end{align}
When $A=+1$, 
\begin{align}
    \frac{g(X)- \tilde{h}(X,d)}{\pi(+1,X)} \mathbbm{1}\{d(X) = +1\} + \tilde{h}(X,d) &= \frac{[\pi(-1,X) - \mathbbm{1}\{d(X)=+1\}]\tilde{m}(+1,X)}{\pi(+1,X)}\mathbbm{1}\{d(X) = +1\} \nonumber\\
        &~~+ \frac{[\pi(+1,X) - \mathbbm{1}\{d(X)=-1\}]\tilde{m}(-1,X) }{\pi(+1,X)} \mathbbm{1}\{d(X) = +1\} \nonumber\\
        &~~+ \tilde{m}(+1,X) \mathbbm{1}\{d(X)=+1\} + \tilde{m}(-1,X) \mathbbm{1}\{d(X)=-1\} \nonumber\\
        &= \frac{\tilde{m}(+1,X) \mathbbm{1}\{d(X) +1\}}{\pi(+1,X)} - \tilde{m}(+1,X) \mathbbm{1}\{d(X) +1\} \nonumber\\
        &~~ - \frac{\tilde{m}(+1,X) \mathbbm{1}\{d(X) +1\}}{\pi(+1,X)} + \tilde{m}(-1,X) \mathbbm{1}\{d(X) = +1\} \nonumber\\
        &~~ -\frac{\tilde{m}(-1,X) \mathbbm{1}\{d(X) -1\} \mathbbm{1}\{d(X) +1\}}{\pi(+1,X)} \nonumber\\
        &~~ + \tilde{m}(+1,X) \mathbbm{1}\{d(X)=+1\} + \tilde{m}(-1,X) \mathbbm{1}\{d(X)=-1\} \nonumber\\
        &= \tilde{m}(-1,X) \mathbbm{1}\{d(X) = +1\} + \tilde{m}(-1,X) \mathbbm{1}\{d(X) = -1\} \nonumber\\
        &= \tilde{m}(-1,X). \nonumber
\end{align}
Likewise, when $A = -1$,
$$ \frac{g(X)- \tilde{h}(X,d)}{\pi(-1,X)} \mathbbm{1}\{d(X) = -1\} + \tilde{h}(X,d) = \tilde{m}(+1,X). $$
From these two results, 
$$ \frac{g(X)- \tilde{h}(X,d)}{\pi(A,X)} \mathbbm{1}\{d(X) = A\} + \tilde{h}(X,d) = \tilde{m}(-A,X). $$
Therefore,
$$ \frac{Y-\tilde{h}(X,d)}{\pi(A,X)} \mathbbm{1}\{d(X) = A\} + \tilde{h}(X,d) = \frac{Y-g(X)}{\pi(A,X)} \mathbbm{1}\{d(X) = A\} + \tilde{m}(-A,X). $$
From this, maximizing the value function $\mathbb{E}[Y\{d(X)\}]$ is equivalent to the maximization of
$$ \mathbb{E}\left[ \frac{Y-g(X)}{\pi(A,X)} \mathbbm{1}\{d(X) = A\} + \tilde{m}(-A,X) \right]. $$
By using the finding in Section 2.2 of \textcite{liu_augmented_2018},
\begin{align}
   \mathbb{E}\left[ \frac{Y-g(X)}{\pi(A,X)} \mathbbm{1}\{d(X) = A\} + \tilde{m}(-A,X) \right] &= \mathbb{E}\left[ \frac{|Y - g(X)|}{\pi(A,X)} \mathbbm{1}\{A \cdot \text{sign}(Y - g(X)) = d(X)\} \right] \nonumber\\
    &~~ + \mathbb{E}[ \tilde{m}(-A,X)] + \mathbb{E}\left[ \frac{\max\{-Y+g(X),0\}}{\pi(A,X)} \right]. \nonumber
\end{align}
Note that
\begin{align}
    \mathbb{E}[\tilde{m}(-A,X)] &= \int \mathbb{E}\left[ x^\top (E[X X^\top])^{-1}X \frac{\pi(a,X)}{\pi(a,x)} \mathbb{E}[Y|X, A=-a] \right] dP_{A,X}(a,x) \nonumber\\
        &= \int \int \mathbb{E}\left[ x^\top (E[X X^\top])^{-1}X \frac{\pi(a,X)}{\pi(a,x)} \mathbb{E}[Y|X, A=-a] \right] dP_{A|X}(a|x) dP_X(x) \nonumber\\
        &= \int \mathbb{E}\left[ x^\top (E[X X^\top])^{-1}X \frac{\pi(+1,X)}{\pi(+1,x)} \mathbb{E}[Y|X, A=-1] \right] \pi(+1,x) dP_X(x)  \nonumber\\
        &~~ + \int \mathbb{E}\left[ x^\top (E[X X^\top])^{-1}X \frac{\pi(-1,X)}{\pi(-1,x)} \mathbb{E}[Y|X, A=+1] \right] \pi(-1,x) dP_X(x) \nonumber\\
        &= \int \mathbb{E}[x^\top (E[X X^\top])^{-1}X\{ \pi(+1,X) \mathbb{E}[Y|X,A=-1] + \pi(-1,X) \mathbb{E}[Y|X,A=+1] \}]  dP_X(x) \nonumber\\
        &= \int \mathbb{E}\left[x^\top (E[X X^\top])^{-1}X \mathbb{E}\left[\frac{\pi(-A,X)}{\pi(A,X)}Y \mid X \right]  \right]  dP_X(x) \nonumber\\
        &= \int \mathbb{E}\left[x^\top (E[X X^\top])^{-1}X \mathbb{E}\left(\frac{\pi(-A,X)}{\pi(A,X)}Y\right)  \right]  dP_X(x) \nonumber\\
        &= \mathbb{E}\left[X^\top \right] \left\{\mathbb{E}[XX^\top] \right\}^{-1} \mathbb{E}\left[\frac{\pi(-A,X)}{\pi(A,X)} XY\right]. \nonumber
\end{align}

Note that only the first term contains the decision rule. Therefore, the decision rule that maximizes the value function is equivalent to the decision rule that maximizes the first term. Equivalently, the minimizing $d(x)$ of $\mathcal{R}(f)$  is the minimizer of 
$$ \mathbb{E}\left[ \frac{|Y - g(X)|}{\pi(A,X)} \mathbbm{1}\{A \cdot \text{sign}(Y - g(X)) \neq d(X)\} \right] $$

\section{Simulation details}

    \subsection{Performance evaluation}
The accuracy of the fitted decision rules was evaluated using the test data as 
$$n^{-1}_{test} \sum\limits_{i=1}^{n_{test}}\mathbbm{1} \left\{\hat{d}\left(X^{test}_i\right) = d^*\left(X^{test}_i\right)\right\},$$ 
while the value function was evaluated using the unbiased estimator of \eqref{eq:value} from \textcite{qian_performance_2011}:
\begin{equation}
\label{eq:Test}
    \frac{ n^{-1}_{test} \sum\limits_{i=1}^{n_{test}} \mathbbm{1}\left\{A^{test}_i = \hat{d}\left(X^{test}_i\right)\right\} Y^{test}_i / \pi \left(A^{test}_i, X^{test}_i\right)}{n^{-1}_{test} \sum\limits_{i=1}^{n_{test}} \mathbbm{1}\left\{A^{test}_i = \hat{d}\left(X^{test}_i\right)\right\} / \pi \left(A^{test}_i, X^{test}_i\right) },
\end{equation}
where the superscript \textit{test} indicates that the observation comes from the test set. Note that $\hat{d}\left(\hat{V}_i\right)$ is evaluated in place of $\hat{d}(X_i)$ in for the proposed DOL method. 

    \subsection{Estimation of ITR}

The KCB weights were obtained by using the R package \texttt{ATE.ncb} (\url{github.com/raymondkww/ATE.ncb}). Default options were used. In gKDR, the dimension $u$ of the central mean subspace was selected using the same two-fold cross-validation procedure as for $\lambda_n$. %The regularization parameter $\epsilon_n$ can be tuned similarly; however, 
We fixed $\epsilon_n = 10^{-5}$. RKHS with Gaussian kernels were used to estimate ITRs and to implement gKDR. Median heuristic bandwidths were selected as recommended in \textcite{gretton_kernel_2012}. The R function \texttt{ipop} from the \texttt{kernlab} package \parencite{karatzoglou_kernlab_2004} was used to implement the weighted SVM formulation implementing AOL. The SI model was fitted using \texttt{simml} and \texttt{pred.simml} from the \texttt{simml} package \parencite{park_hyung_simml_2021}. The function \texttt{ql} from the R package \texttt{DTRlearn2} \parencite{chen_yuan_dtrlearn2_2020} was used to fit the $\ell_1$-PLS models. R-learner was implemented by using \texttt{rkern} from R package \texttt{rlearner} (\url{https://github.com/xnie/rlearner}). The median heuristic bandwidths were used for the kernel ridge regression model and the regularization parameters were tuned by the five-fold cross-validation. The decision rule was obtained by the sign of the estimated heterogeneous treatment effect.

    \subsection{Simulation settings}

\subsection{Setting 1} %% R3
The observed covariates $X \in \mathbb{R}^{50}$ are defined as $X^{(1)} = \exp(Z^{(1)}/2)$; $X^{(2)} = Z^{(2)} / \{1 + \exp(Z^{(1)})\}$;$X^{(3)} = (Z^{(1)} \cdot Z^{(3)} / 25 + 0.6)^3$; $X^{(4)} = (Z^{(2)} + Z^{(4)} + 20)^2$; $X^{(i)} = Z^{(i)}$ for $i \in \{5,\cdots,50\}$. 
The basis matrix $B_0$ consists of two columns $(0.6/a, 0.5/a, -0.2/a,\mathbf{0}^\top _{47})^\top $ and $(0, 0.2/b, 0.5/b, 0, -0.5/b, \mathbf{0}^\top _{46})^\top $, where $a = \sqrt{0.6^2 + 0.5^2 + (-0.2)^2}$ and $ b = \sqrt{0.2^2 + 0.5^2 + (-0.5)^2}$. Therefore, the dimension-reduced covariates are $V = B^\top _0 X \in \mathbb{R}^2$. The outcome values were generated from the following main effect and treatment-covariate interaction term:
$$ 
\mu(X) = 5 + 6Z^{(1)} + 8Z^{(2)} + 3Z^{(3)} + 5Z^{(4)} + 5Z^{(5)}, 
$$
$$ 
\tilde{f}^V(V) = 5 \cdot \sin\left( \frac{\pi}{V^{(1)}+1} \cdot \frac{1}{\sqrt{-V^{(2)}}} \right) + 2.5 \cdot \sin\left(\pi V^{(1)}\right) \cdot 
\log\left(-V^{(2)}\right),
$$
from which $Y \sim N\{\mu(X) + A \cdot \tilde{f}^V(V), 1\}$. The propensity score for the non-randomized study is 
$$Pr(A = +1 | X) = \frac{\exp \left(Z^{(1)} - Z^{(2)} - Z^{(4)} - Z^{(6)} - Z^{(8)} - Z^{(10)}\right)}{ 1 + \exp\left(Z^{(1)} - Z^{(2)} - Z^{(4)} - Z^{(6)} - Z^{(8)} - Z^{(10)}\right)}.$$

\subsection{Setting 2} %52ss
We set $X \in \mathbb{R}^{50}$ by defining $X^{(1)} = \exp(Z^{(1)}+1) + Z^{(2)}$; $X^{(2)} = Z^{(2)2} \cdot Z^{(3)}$; $X^{(3)} = \sin(2 Z^{(3)}) \cdot (Z^{(4)}+5)^2$; $X^{(4)}=\left( Z^{(2)3} + Z^{(4)} + 10 \right)\cdot \left( Z^{(2)} + Z^{(4)3} + 10 \right)$; $X^{(i)} = Z^{(i)}$ for $i \in [n]  \backslash \{1,2,3,4\}$. The dimension-reduced covariate $V = B^\top _0 X \in \mathbb{R}^2$ is defined as in Setting 1. The outcome is generated as $N\{\mu(X) + A\cdot \tilde{f}^V(V) , 1 \}$, where
$$\mu(X) = 10 + 7Z^{(1)}+13Z^{(2)}+15Z^{(3)}+15Z^{(4)} + 10Z^{(5)}+ 7Z^{(6)}+13Z^{(7)}+15Z^{(8)}+15Z^{(9)} + 10Z^{(10)}, $$
$$\tilde{f}^V(V) = 6 \sin\left(V^{(1)}/3 \right) \cdot \log\left( 
\left|V^{(2)}\right| + 1 \right) + 4 \cos\left( V^{(2)}\sqrt{\left|V^{(1)}\right|} \right) + 5 \tan^{-1}\left\{2\left(V^{(1)}-1\right) \log\left(\left|V^{(2)}_0\right|+1\right)\right\}.$$
For the non-randomized study setting, we define the propensity score as
$$
\Pr(A = +1 | X) =\frac{\exp \left( -Z^{(3)} + 2Z^{(4)} - Z^{(5)} - 0.5Z^{(6)} \right)}{1+\exp \left( -Z^{(3)} + 2Z^{(4)} - Z^{(5)} - 0.5Z^{(6)} \right)}.
$$
Figure \ref{fig:Setting2} displays the distribution of the Bayes rule at different levels of covariates for Setting 2.

\subsection{Setting 3}

In Setting 3, we consider a higher dimension for the central mean subspace. We set $X \in \mathbb{R}^{50}$ by defining $X^{(2)} = \left(Z^{(2)}-0.2Z^{(4)}+6\right)^3$; $X^{(4)} = \exp\left(0.5Z^{(4)}\right)$; $X^{(6)} = Z^{(6)} \cdot \left\{1 + \exp\left(Z^{(4)}\right) \right\}^{-1} $; $X^{(8)} = \left(Z^{(6)} + Z^{(8)}+20\right)^2$; $X^{(i)} = Z^{(i)}$ for $i \in [n]  \backslash \{2,4,6,8\}$. The dimension-reduced covariate $V = B^\top _0 X \in \mathbb{R}^4$ is defined from a basis matrix that consists of $\left(1/\sqrt{2},-1/\sqrt{2},\mathbf{0}^\top _{48}\right)^\top $, $\left(1/\sqrt{2}, 1/\sqrt{2},\mathbf{0}^\top _{48}\right)^\top $, $\left(0,0,1/\sqrt{2},-1/\sqrt{2},\mathbf{0}^\top _{46}\right)^\top $, and  $\left(0,0,1/\sqrt{2}, 1/\sqrt{2},\mathbf{0}^\top _{46}\right)^\top $. The outcome is generated as $N\{\mu(X) + A\cdot \tilde{f}^V(V) , 1 \}$, where
$$ 
\mu(X) = 6Z^{(1)} + 6Z^{(2)} + 10Z^{(3)} + 10Z^{(4)} + 12Z^{(5)} + 12Z^{(6)} + 8Z^{(7)} + 8Z^{(8)} + 6Z^{(9)} + 6Z^{(10)}, 
$$
$$
\tilde{f}^V(V) = \left(V^{(3)}+10\right)\cos\left\{2\pi \log\left(V^{(1)} -2
\right) \right\} + 3\sqrt{-V^{(2)}}\cdot\sin\left(0.5\pi V^{(1)}\right)/\left(\pi \sqrt{V^{(1)}}\right)
$$
$$+ \tan^{-1}\left\{2\log\left(-V^{(2)}\right)\right\} \sqrt{\left|V^{(4)}-4 \right|}-3.$$
For the non-randomized study setting, we define the propensity score as
$$\Pr(A = +1 | X) =\frac{\exp \left( 0.6Z^{(1)} + 1.2Z^{(2)} + 1.2Z^{(3)} - 0.8Z^{(4)} - Z^{(5)} - Z^{(6)} \right)}{1+\exp \left( 0.6Z^{(1)} + 1.2Z^{(2)} + 1.2Z^{(3)} - 0.8Z^{(4)} - Z^{(5)} - Z^{(6)} \right)}.$$

\section{Additional simulation results}

    \subsection{Results for $n = 500$}

\begin{figure}[H]
\centering
\includegraphics[width = 5.5in]{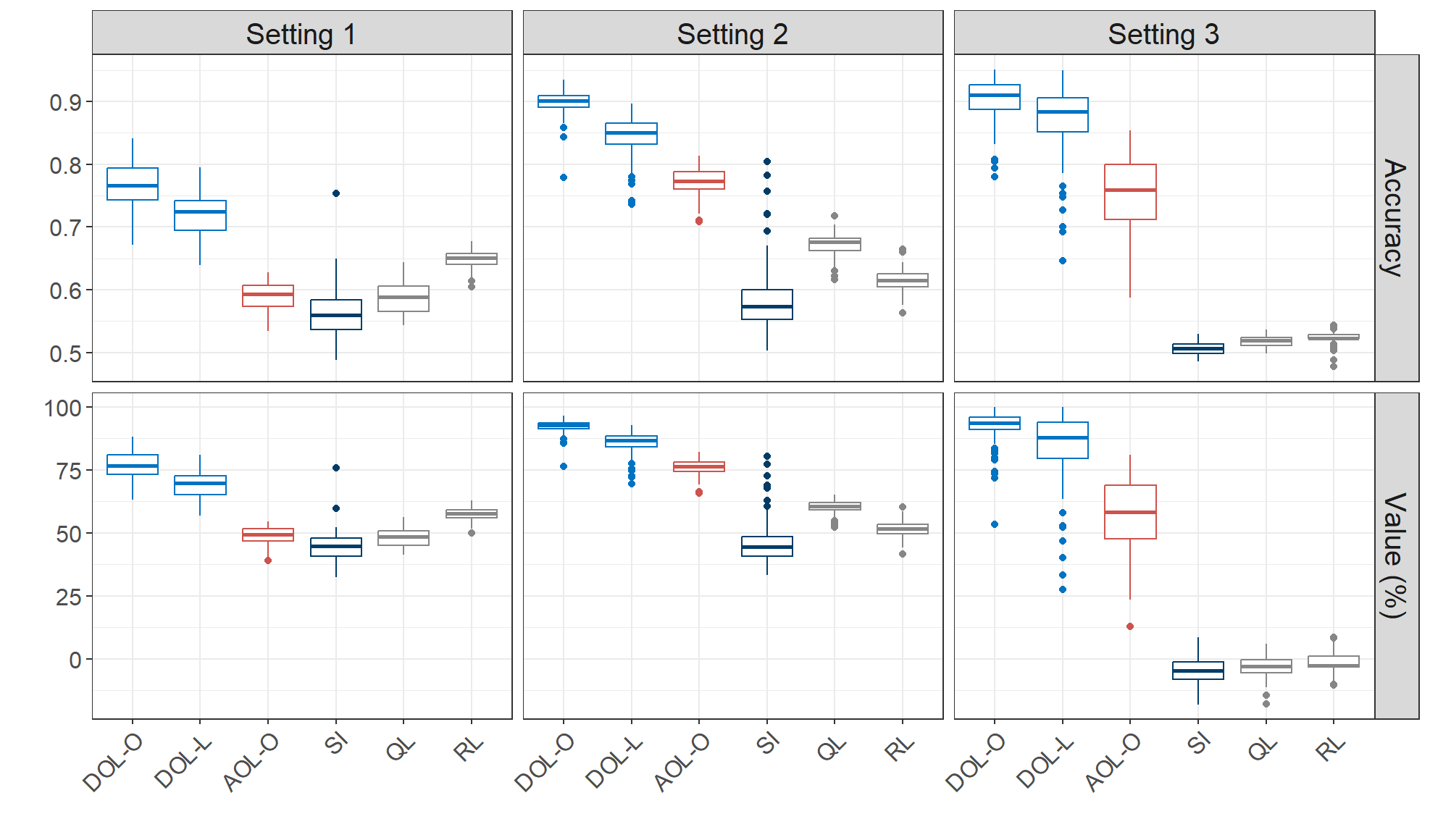}
\caption{Simulation results for $n = 500$ from the randomized scenarios across the three different settings. Accuracy represents the proportion of correctly predicted optimal treatments. Value (\%) represents the value function recovered by the estimated decision rule as a percentage of the Bayes optimal value function. The performance of the proposed method is shown under DOL-O and DOL-L: DOL-O uses the oracle $g(x)$, while DOL uses $g_{\hat{w}}(x)$ estimated via linear regression.
}
\label{fig:Random500}
\end{figure}

\begin{figure}
\centering
\includegraphics[width = 5.5in]{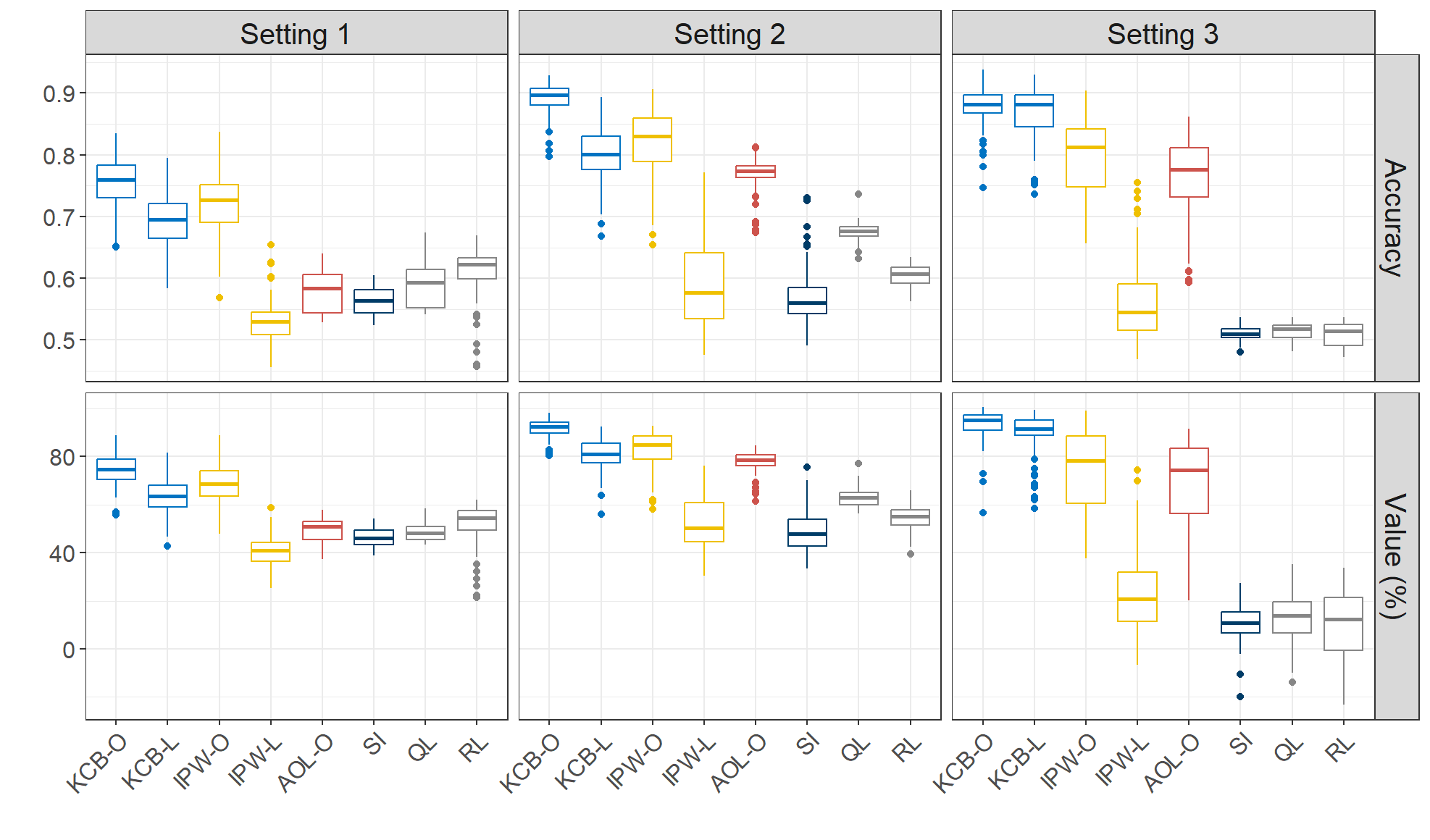}
\caption{Simulation results for $n = 500$ from the non-randomized scenarios across the three different settings. Accuracy represents the proportion of correctly predicted optimal treatments. Value (\%) represents the value function recovered by the estimated decision rule as a percentage of the Bayes optimal value function. The performance of the proposed method is shown under KCB-O and KCB-L: KCB-O uses the oracle $g(x)$, while KCB-L uses the fitted values of $g_{\hat{w}}(x)$ obtained via linear regression. IPW-O uses the oracle $g(x)$, and IPW-L uses $g_{\hat{w}}(x)$ from linear regression, with propensity scores estimated via logistic regression.}
\label{fig:NonRandom500}
\end{figure}

The simulation results for the smaller sample size ($n=500$) in the randomized and non-randomized scenarios are displayed in Figure \ref{fig:Random500} and Figure \ref{fig:NonRandom500}, respectively. With the smaller sample size, we can observe that the performance of the proposed DOL is more pronounced than the results with $n=1000$ when compared to AOL in all settings.

    \subsection{Projection error}
\begin{figure}
\centering
\includegraphics[width = 0.9\linewidth]{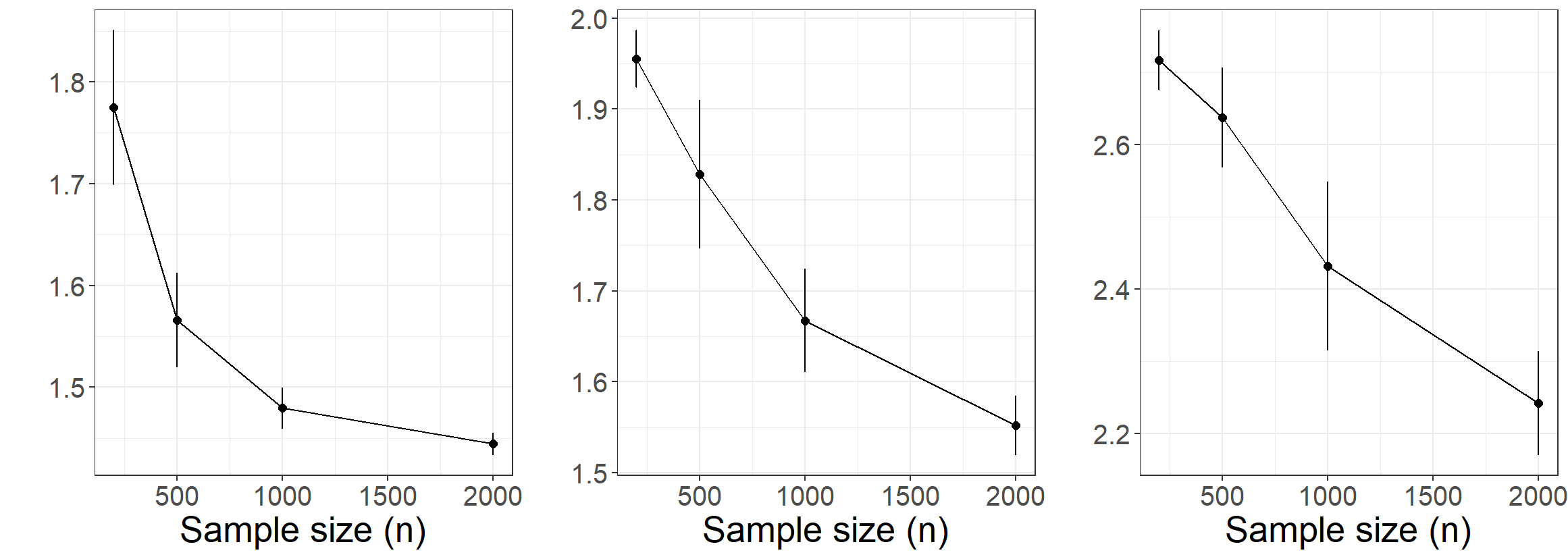}
\caption{Simulation results that display the change in the error in the projection from $\hat{B}$ along different sample sizes represented by $\|\hat{B}\hat{B}^\top - B_0 B_0^\top\|_F$ ($y$-axis). From left to right, the figures correspond to the non-randomized treatment assignment scenarios of Settings 1, 2, and 3. The means and standard deviations are depicted at each sample size.}
\label{fig:FrobErr}
\end{figure}

Here we introduce additional simulation results that show decreasing trends along the increasing sample size in the difference between the estimated projection from $\hat{B}$ and the true projection from $B_0$. To do so, we iterated the process of obtaining $\hat{B}$ and computing the Frobenius distance from $B_0$ by $\|\hat{B}\hat{B}^\top - B_0 B_0^\top\|$ at different sample sizes: 200, 500, 1,000, and 2,000. The simulation was performed on the three non-randomized settings used for comparing different approaches, where the $\hat{B}$ obtained from the pseudo outcomes that use KCB weights. The dimensions for $\hat{B}$ were fixed at the true values and the Tikhonov penalty was fixed at $10^{-5}$ to focus on the association with the sample size. The results are displayed in Figure \ref{fig:FrobErr}. We can notice the decline in the projection errors as the sample size increases.

\section{Real world data}

In this section, we layout the details on the 35 baseline variables: age (18-91), gender, weight, and indicator of service unit (medical ICU or surgical ICU); \textit{severity at admission} measurements from simplified acute physiology score, the sequential organ failure assessment, and Elixhauser comorbidity score; \textit{comorbidity indicators}, which are congestive heart failure, atrial fibrillation, chronic renal disease, liver disease, chronic obstructive pulmonary disease (COPD), coronary artery disease (CAD), stroke, respiratory failure, and malignant tumor; \textit{vital signs} measured from mean arterial pressure, heart rate, temperature ($^\circ$F); \textit{interventions} that use mechanical ventilation, inotropic and vasopressor agents, and/or sedative drugs within 24 hours of ICU admission; \textit{laboratory results} that measured white blood cell, hemoglobin, sodium, potassium, bicarbonate, chloride, creatinine, pH, partial pressure of carbon dioxide (PCO2), platelet count, partial pressure of oxygen (PO2), lactate, and blood urea nitrogen.

    \subsection{Estimation of ITR and evaluation of modified value function}

The gKDR was implemented by selecting the regularization parameter $\epsilon_n$ out of 8 candidate values ($10^{-9} \text{-} 10^{-2}$) and the dimension of central mean subspace ($u$) out of 9 candidate dimensions ($1\text{-}9$) from five-fold cross-validation.

AOL was implemented by using the IPW estimated from logistic regression. In DOL, $g_{\hat{w}}(x)$ was estimated by linear regression. Likewise, $g_(x)$ was estimated by linear regression with the estimated propensity scores. The tuning parameter $\lambda_n$ was chosen from five-fold cross-validation.

The modified value function is evaluated by using the validation set as
$$
\frac{1}{n_{test}} \sum\limits_{i=1}^{n_{test}} \frac{Y^{test}_{i}}{\hat{\pi}(A^{test}_{i}, X^{test}_{i})} \left[ \mathbbm{1}\left\{A^{test}_{i} = \hat{d}\left(X^{test}_{i}\right)\right\} - \mathbbm{1}\left\{A^{test}_{i} =\hat{d}\left(X^{test}_{i}\right)\right\}\right]. 
$$
The true propensity scores are not available to compute the modified value function with the validation set. Thus, the propensity scores were estimated from gradient boosted models \parencite{mccaffrey_propensity_2004} as done in \textcite{feng_transthoracic_2018} by using \texttt{ps} from R package \texttt{twang} \parencite{matthew_cefalu_twang_2024}, and were used in place of $\hat{\pi}(A^{test}_{i}, X^{test}_{i})$.

\end{document}